\documentclass{article}

\usepackage[final]{neurips_2025}

\usepackage[utf8]{inputenc} 
\usepackage[T1]{fontenc}    
\usepackage{hyperref}       
\usepackage{url}            
\usepackage{booktabs}       
\usepackage{amsfonts}       
\usepackage{nicefrac}       
\usepackage{microtype}      
\usepackage{xcolor}         
\usepackage[subrefformat=parens,labelformat=parens]{subfig}
\usepackage{graphicx}
\usepackage{amsmath}
\usepackage{amssymb}
\usepackage{enumitem}
\usepackage{amsthm}
\usepackage{algorithm}
\usepackage[noend]{algpseudocode}
\usepackage{placeins}
\usepackage{xspace}
\usepackage{bbm}

\title{Optimistic Query Routing in Clustering-based\\Approximate Maximum Inner Product Search}

\author{
  Sebastian Bruch \\
  Northeastern University \\
  Boston, MA, USA \\
  \texttt{s.bruch@northeastern.edu}
  \And
  Aditya Krishnan \\
  Microsoft \\
  New York, NY, USA \\
  \texttt{adkrishnan@microsoft.com}
  \And
  Franco Maria Nardini \\
  ISTI-CNR \\
  Pisa, Italy \\
  \texttt{francomaria.nardini@isti.cnr.it}
}

\newtheorem{problem}{Problem}
\newtheorem{lemma}{Lemma}
\newtheorem{fact}{Fact}

\newtheorem{definition}{Definition}

\DeclareMathOperator*{\argmax}{arg\,max \quad}

\DeclareMathOperator*{\R}{\mathbb{R}}

\newcommand{\Dc}{\mathcal{D}}
\newcommand{\Nc}{\mathcal{N}}

\newcommand{\mean}{\textsc{Mean}\xspace}
\newcommand{\normalizedmean}{\textsc{NormalizedMean}\xspace}
\newcommand{\scann}{\textsc{Scann}\xspace}
\newcommand{\subpartition}{\textsc{SubPartition}\xspace}
\newcommand{\optimist}{\textsc{Optimist}\xspace}

\newcommand{\textimage}{\textsc{Text2Image}\xspace}
\newcommand{\music}{\textsc{Music}\xspace}
\newcommand{\deep}{\textsc{DeepImage}\xspace}
\newcommand{\glove}{\textsc{GloVe}\xspace}
\newcommand{\msmarco}{\textsc{MsMarco-MiniLM}\xspace}
\newcommand{\nq}{\textsc{Nq-Ada2}\xspace}

\newcommand{\bigO}{\mathcal{O}}

\begin{document}

\maketitle

\begin{abstract}
Clustering-based nearest neighbor search algorithms partition points into shards to form an index, and search only a subset of shards to process a query. Even though search efficacy is heavily influenced by the algorithm that identifies the shards to probe, it has received little attention in the literature. We study routing in clustering-based maximum inner product search, which includes cosine similarity search. We unpack existing routers and notice the surprising role of optimism. We then take a page from the sequential decision making literature and formalize that insight following the principle of ``optimism in the face of uncertainty.'' In particular, we present a framework that incorporates the moments of the distribution of inner products within each shard to estimate the maximum inner product. We then develop a practical instance of our algorithm that uses only the first two moments to reach the same accuracy as state-of-the-art routers by probing up to $50\%$ fewer points on benchmark datasets without compromising efficiency. Our algorithm is also space-efficient: we design a sketch of the second moment whose size is independent of the number of points and requires $\mathcal{O}(1)$ vectors per shard.
\end{abstract}

\section{Introduction}
\label{section:introduction}

A fundamental operation in many applications of machine learning is nearest neighbor search~\citep{Bruch_2024}: Given $m$ points denoted $\mathcal{X} \subset \R^d$, it finds the $k$ closest points to a query $q \in \R^d$, where closeness is by vector similarity or distance. We focus on inner product, giving the problem of Maximum Inner Product Search (MIPS), which includes Cosine Similarity search as a special case:
\begin{equation}
    \mathcal{S} = \argmax_{u \in \mathcal{X}}^{(k)} \langle q, u \rangle.
    \label{equation:mips}
\end{equation}
As $m$ or $d$ grows, an exact solution is often difficult to obtain within a reasonable budget. The problem is thus relaxed to its \emph{approximate} variant, where we tolerate error to gain speed. The effectiveness of Approximate Nearest Neighbor (ANN) search is characterized by \emph{recall} or \emph{accuracy}, defined as the fraction of true nearest neighbors recalled: $\lvert \tilde{\mathcal{S}} \cap \mathcal{S} \rvert/k$, $\tilde{\mathcal{S}}$ being the set returned by the ANN algorithm.

\medskip
\noindent \textbf{Clustering-based ANN search}: ANN algorithms come in various flavors, from trees~\citep{kdtree,dasgupta2015rptrees}, LSH~\citep{lsh}, to graphs~\citep{hnsw2020,diskann}. Refer to~\citep{Bruch_2024} for a review of this subject. The method relevant to us is the clustering-based approach, also known as Inverted File, which has received much attention in the literature~\citep{auvolat2015clustering,invertedMultiIndex,chierichetti2007clusterPruning,bruch2023bridging,douze2024faiss} and is widely adopted in industrial applications.\footnote{See, for example, \url{https://turbopuffer.com/blog/turbopuffer}, \url{https://www.pinecone.io/blog/serverless-architecture/\#Pinecone-serverless-architecture}, and \url{https://research.google/blog/announcing-scann-efficient-vector-similarity-search/}.}

In this paradigm,  $\mathcal{X}$ is partitioned into $C$ \emph{shards} using a clustering function $\zeta: \R^d \rightarrow [C]$---typically KMeans with $C = \mathcal{O}(\sqrt{m})$. Accompanying this \emph{index} is a \emph{router} $\tau: \mathbb{R}^d \rightarrow [C]^\ell$, which returns $\ell$ shards likely to contain $q$'s nearest neighbors. If $\mu_i$ is the \emph{mean} of the $i$-th shard, a common router is:
\begin{equation}
    \label{equation:mean-router}
    \tau(q) = \argmax^{(\ell)}_{i \in [C]} \langle q, \mu_i \rangle.
\end{equation}

Processing a query $q$ involves two \emph{independent} subroutines. The first, ``routing,'' returns $\ell$ shards using $\tau(q)$, and the subsequent step, ``scoring,'' computes inner products between $q$ and the points in the $\ell$ shards. While the literature has focused on scoring~\citep{pq,opq,multiscaleQuantization,Andre_2021,locallyOptimizedPQ,pqWithGPU,norouzi2013ckmeans}, routing has received little attention. This work turns squarely to the routing step.

We emphasize that, routing is \emph{independent} of the scoring algorithm. For example, we may route using Equation~(\ref{equation:mean-router}) or our novel router to be introduced later. This choice has no algorithmic bearing on scoring whatsoever. In other words, scoring can be accomplished by a linear scan of the chosen shards, computing inner products with Product Quantization (PQ)~\citep{pq}, or searching each shard using a graph ANN algorithm~\cite{diskann,hnsw2020}. All these choices are independent of the type of router utilized in the earlier stage.

\medskip
\noindent \textbf{The importance of routing}: The historical focus on scoring makes sense. Even though routing narrows down the search space, selected shards may contain a large number of points. Scoring must thus be efficient and effective.

We argue that the routing step is important in its own right. First, a more accurate\footnote{We quantify a router's accuracy by the ANN recall in a setup where scoring is by a linear scan over the identified shards, so that the only source of error is the routing procedure.} router leads to fewer data points being scored. For example, if shards are balanced in size, access to an oracle router (i.e., one that identifies the shard with the true nearest neighbor) means that the scoring stage must only compute inner products for $m/C$ points to find the top-$1$ point.

The second reason pertains to scale. As size and dimensionality grows, it is often infeasible to keep the entire index in memory, compression techniques such as PQ notwithstanding. Much of the index must instead rest on secondary storage---cheap but high-latency storage such as disk or blob storage---and accessed when necessary. That line of reasoning has led to the emergence of disk-based graph indexes~\citep{diskann,singh2021freshdiskann,jaiswal2022ooddiskann}.

Translating the same rationale to the clustering-based paradigm implies that shards rest outside of memory, and when a router identifies a subset of shards, scoring must first fetch those shards from storage. This paradigm has gained traction in real-world database systems.\footnote{c.f., documents linked in the footnote above.} This is the context in which we present our research: Storage-backed, clustering-based ANN search systems.

In this framework, a more accurate router lowers the volume of data transferred between storage and memory with implications for storage bandwidth and in-memory cache space utilization, and retrieval throughput. We have included an illustrative example in Appendix~\ref{appendix:illustrative-example} to support this statement.

\medskip
\noindent \textbf{Existing routers and the role of optimism}: We review the relevant literature on routers in Appendix~\ref{appendix:related-work}. Among existing routers, Equation~(\ref{equation:mean-router}), which we call \textbf{\mean}, is the most prominent. It summarizes each shard by its mean point ($\mu_i$'s). This is not an unreasonable choice as the mean is the minimizer of the sum of squared deviations from the mean, and is, in fact, natural if $\zeta$ is KMeans.

\begin{figure*}[t]
\begin{center}
\centerline{
    \subfloat[\label{figure:motivation:mean_vs_normalized_mean}]{
        \includegraphics[width=0.3\linewidth]{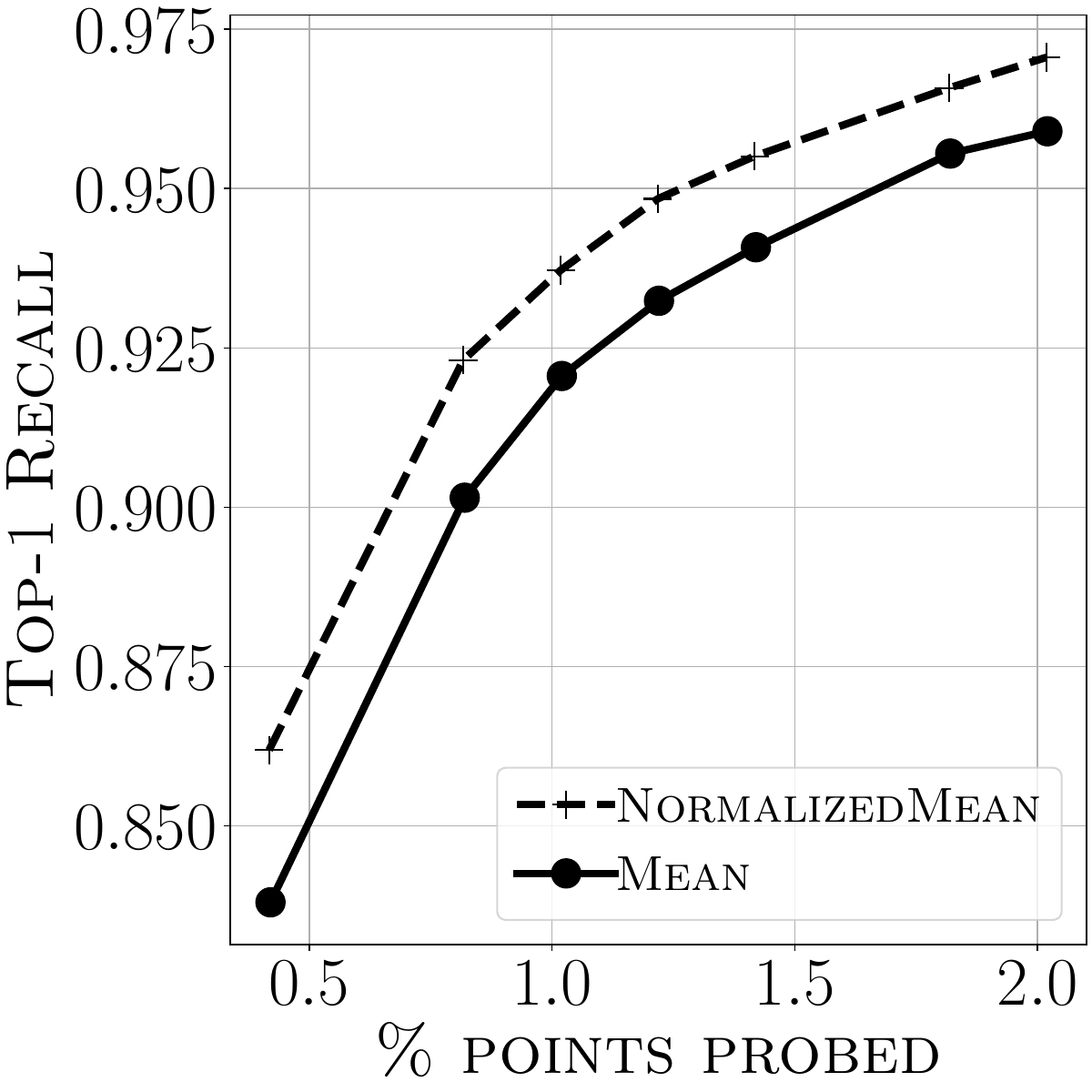}
    }
    \subfloat[\label{figure:motivation:score_distribution}]{
        \includegraphics[width=0.3\linewidth]{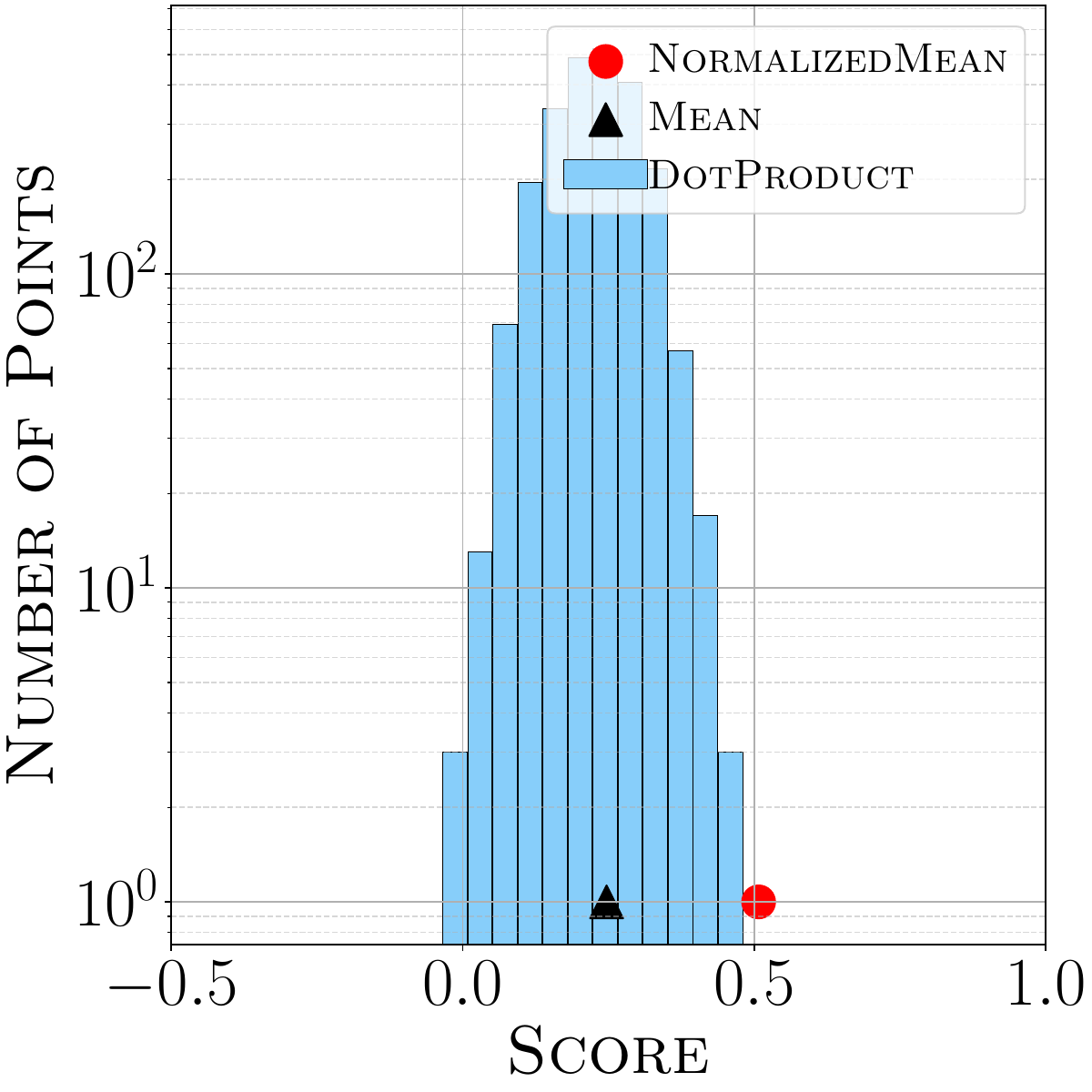}
    }
    \subfloat[\label{figure:motivation:score_distribution_overzealous}]{
        \includegraphics[width=0.3\linewidth]{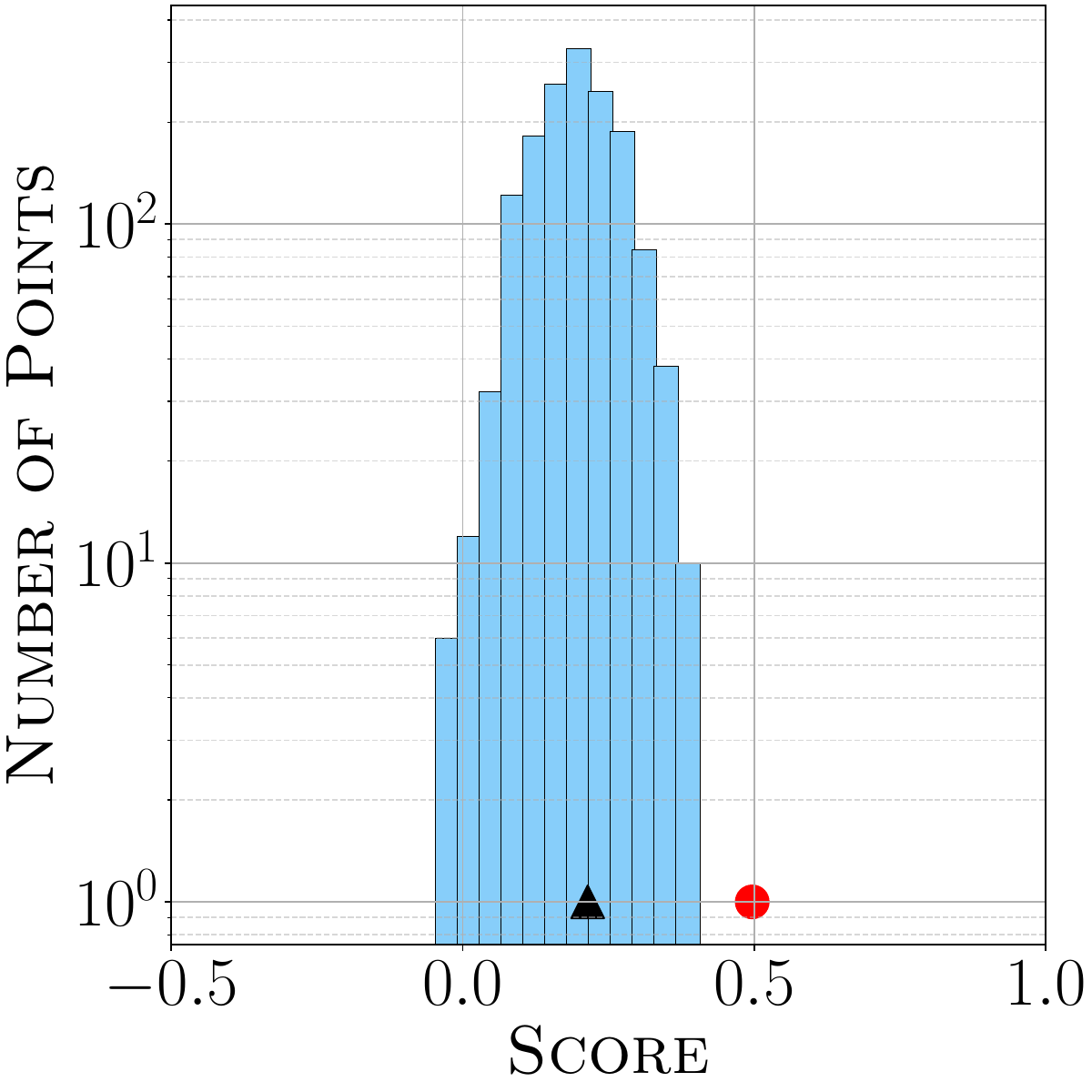}
    }
}
\caption{(a) Top-$1$ recall vs. percentage of points probed on \textimage where points have varying norms; (b) and (c) Distribution of inner products between a shard and a query on \glove. Overlaid are scores computed by \mean and \normalizedmean. \vspace{-1em}}
\label{figure:motivation}
\end{center}
\end{figure*}

Another common router, which we call \textbf{\normalizedmean}, belongs to the same family, but its shard representatives are the $L_2$-normalized means, rather than the unnormalized mean vectors:
\begin{equation}
    \label{equation:normalized-mean-router}
    \tau(q) = \argmax^{(\ell)}_{i \in [C]} \langle q, \frac{\mu_i}{\lVert \mu_i \rVert_2} \rangle.
\end{equation}
This formulation is inherited from \emph{spherical} KMeans~\citep{sphericalKMeans}, which is identical to standard KMeans except that cluster centroids are normalized after every iteration. Because we can assume that $\lVert q \rVert_2 = 1$ without loss of generality, Equation~(\ref{equation:normalized-mean-router}) routes by the angle between $q$ and $\mu_i$'s.

\normalizedmean seems appropriate for search over a sphere: When norms are constant, only angles matter in Equation~(\ref{equation:mips}), making it reasonable to rank shards by the angle between means and $q$. Intriguingly, in many circumstances that deviate from that situation, \normalizedmean performs more accurately than \mean, as evidenced in Figure~\subref*{figure:motivation:mean_vs_normalized_mean}. Let us unpack this phenomenon.

Consider $q$ and a single shard $\mathcal{P}$ with mean $\mu$. Figure~\subref*{figure:motivation:score_distribution} visualizes the distribution of inner products of $q$ and all points in $\mathcal{P}$. Overlaid is the inner product of $q$ and $\mu$, as well as their normalized mean. We observe that $\langle q, \mu / \lVert \mu \rVert_2 \rangle$ lands in the right tail of the distribution.

That is not surprising when $\lVert u \rVert_2=1$ for all $u \in \mathcal{P}$. Clearly $\lVert \mu \rVert_2 \leq 1$, so that $\langle q, \mu / \lVert \mu \rVert_2 \rangle = \langle q, \mu \rangle / \lVert \mu \rVert_2 \geq \langle q, \mu \rangle$; \normalizedmean amplifies \mean by a factor of $1/\lVert \mu \rVert_2$. What is interesting, however, is that the magnitude of this ``boost'' correlates with the variance of $\mathcal{P}$: The more concentrated $\mathcal{P}$ is around the direction of $\mu$, the closer $\mu$ is to the surface of the sphere, making $\lVert \mu \rVert_2$ larger, so that \normalizedmean applies a smaller boost to $\langle q, \mu \rangle$. The opposite is true when points are spread out: shards with a higher variance receive a larger lift by \normalizedmean.

There are a few caveats. First, it is hard to explain the behavior on sets with varying norms. Second, as we observe in Figure~\subref*{figure:motivation:score_distribution_overzealous}, \normalizedmean's estimate can be overly aggressive. Nonetheless, the insight that a router's score for a shard can be influenced by the shard's variance is worth exploring.

\medskip
\noindent \textbf{Contributions and outline}: From the exercise above, we learned that \normalizedmean paints an \emph{optimistic} picture of what the maximum inner product of $q$ and $\mathcal{P}$ could be. \mean, however, gives a \emph{conservative} estimate. We investigate the ramifications of that insight in this work.

In particular, the research question we study is whether a more principled approach to designing optimistic routers leads to more accurate routing on vector sets with variable norms. This question is not unlike those asked in the online learning literature~\citep{lattimore2020bandit}, so it is not surprising that our answer draws from the ``principle of optimism in the face of uncertainty.''

We apply the Optimism Principle to routing in clustering-based MIPS. Our method, presented in Section~\ref{section:method}, uses the concentration of inner products between a query $q$ and points in a shard. In particular, we estimate a score, $\theta_i$, for the $i$-th shard such that, with some confidence, the maximum inner product of $q$ with points in that shard is at most $\theta_i$. Shards are then ranked by this score. When $\theta_i = \mu_i$, we recover the \mean router, and when $\theta_i > \mu_i$ routing is optimistic with a controllable degree of optimism. Importantly, our routing function is independent of how shards are formed.

We outline a general framework that can incorporate as much information as is available about the data distribution to estimate $\theta_i$'s. We then present a concrete, assumption-free instance of our algorithm that uses the first and second moments of the empirical inner product distribution only. Furthermore, we make the resulting algorithm space-efficient by designing a sketch~\citep{woodruff2014sketching} of the second moment. The end-result is a practical algorithm that is straightforward to implement.

We test our proposal in Section~\ref{section:experiments} on a variety of ANN benchmark datasets. As our experiments show, our optimistic router achieves the same recall as state-of-the-art routers but with up to a $50\%$ reduction in the total volume of points evaluated per query. We conclude this work in Section~\ref{section:conclusion}.

\section{Routing by the Optimism Principle: our proposal}
\label{section:method}

As we observed, \normalizedmean is an optimistic estimator, though its behavior is unpredictable. Our goal is to design an optimistic estimator that is statistically principled, thus well-behaved. Throughout this section, all discussions are in the context of a fixed unit-vector $q$. We denote the $i$-th shard by $\mathcal{P}_i$, and write $S_i$ for the set of inner products of $q$ and points in $\mathcal{P}_i$: $S_i = \{ \langle q, u \rangle : u \in \mathcal{P}_i \}$.

\subsection{Formalizing the notion of optimism}
We wish to find the \emph{smallest} threshold $\theta_i \geq \langle q, \mu_i \rangle$ for the $i$-th shard such that the probability that a sample from $S_i$ falls to the left of $\theta_i$ is at least $(1 + \delta)/2$, for some arbitrary $\delta \in [0, 1]$. Formally, we aim to compute a solution $\theta_i$ to the the following optimization problem: \begin{problem}[Optimistic estimator of the maximum inner product in $S_i$]
\begin{equation*}
\label{problem:optimization-problem}
\inf \{\theta_i : \theta_i \geq \langle q, \mu_i \rangle \} \quad \text{ such that } \Pr_{s \sim S_i}[s \leq \theta_i] \geq \frac{1 + \delta}{2}.
\end{equation*}
\end{problem}
The optimal $\theta_i$ is a probabilistic upper-bound on the maximum attainable value in $S_i$; that is, with some confidence, we can assert that no value in $S_i$ is greater than $\theta_i$. We then route $q$ by sorting $\mathcal{P}_i$'s by their $\theta_i$ in descending order and select the top shards. Our router, \optimist, is defined as follow:
\begin{equation}
    \label{equation:optimistic-router}
    \tau(q) = \argmax^{(\ell)}_{i \in [C]} \theta_i.
\end{equation}

Suppose that we have the solution to Problem~\ref{problem:optimization-problem}. As $\delta \rightarrow 0$, the optimal solution approaches $\theta_i = \max (\langle q, \mu_i \rangle, \langle q, M_i \rangle )$, where $M_i$ is the median of $\mathcal{P}_i$. That is a \emph{conservative} estimate of the maximum inner product. As $\delta \rightarrow 1$, $\theta_i$ becomes larger, rendering Equation~(\ref{equation:optimistic-router}) optimistic. At the extreme, the optimal $\theta_i$ is the maximum inner product itself---the most optimistic we can get.

Clearly $\delta$ controls the degree of optimism. When the distribution is fully known, $\delta = 1$ is an appropriate choice: If we know the distribution of inner products, we can expect to be fully confident about the maximum inner product. But when very little about the distribution is known, then $\delta = 0$ is a sensible choice. In effect, the value of $\delta$ reflects our knowledge of the data distribution.

What is left to address is the solution to Problem~\ref{problem:optimization-problem}, which is the topic of the remainder of this section. We first describe a general approach that uses as much information as available about the data distribution. We then present a practical approach that is the foundation of the rest of this work.

\subsection{General solution}
\label{section:general-solution}

Let $\Dc$ be an unknown distribution over $\R^d$ from which $\mathcal{P}_i$ is sampled. It is clear that, if we are able to accurately approximate the quantiles of $S_i$, we can obtain an estimate $\theta_i$ satisfying Problem~\ref{problem:optimization-problem}.

We motivate our approach by considering a special case where $\Dc$ is $\Nc(\mu_i, \Sigma_i)$, a Gaussian with mean $\mu_i \in \R^d$ and covariance $\Sigma_i \in \R^{d \times d}$. In this case, $S_i$ follows a univariate Gaussian distribution with mean $\langle q, \mu_i \rangle$ and variance $q^\top \Sigma_i q$, so that the solution to Problem~\ref{problem:optimization-problem} is simply:
\begin{equation}
\label{equation:gaussian-theta}
\theta_i = \langle q, \mu_i \rangle + \sqrt{q^\top \Sigma_i q} \cdot \Phi_{\Nc(0, 1)}^{-1}\left(\frac{1+\delta}{2}\right),
\end{equation}
which follows by writing the cumulative distribution function (CDF) of $S_i$ in terms of the CDF of a unit Gaussian, denoted $\Phi_{\Nc(0, 1)}$. Notice that, we first modeled the moments of $S_i$ using the moments of $\Dc$, then approximated the CDF (equivalently, the quantile function) of the distribution of $S_i$ from its moments---in this special case, the approximation with the first two moments is, in fact, exact.

Our general solution follows that same logic. In the first step, we can obtain the first $r$ moments of the distribution of $S_i$, denoted $m_j(S_i)$ for $j \in [r]$, from the moments of $\Dc$, denoted $m_j(\Dc)$. In a subsequent step, we use $m_j(S_i)$'s to approximate the CDF of $S_i$.

It is easy to see that the $j$-th moment of $S_i$ can be written as follows:
\begin{equation*}
    m_j(S_i) = q^{\otimes j} \odot m_j(\Dc) \approx \frac{1}{n} \cdot q^{\otimes j} \odot \sum_{u_1, \dots, u_n \sim \Dc} x^{\otimes j},
\end{equation*}
where $^{\otimes j}$ is the $j$-fold tensor product, and $\odot$ tensor inner product.

In a second step, we find a distribution $\tilde{S}_i$ such that $m_j(\tilde{S}_i) \approx m_j(S_i)$ for all $j \in [r]$. This can be done using the ``method of moments''~\cite{pearson1936method} which offers guarantees~\citep{kong2017spectrum,braverman2022sublinear} in terms of a distributional distance such as the Wasserstein-1 distance.

While using higher-order moments leads to a better approximation, computing $m_j(\cdot)$ for $j > 2$ can be prohibitive considering the dimensionality of real datasets, as the space requirement to store $m_j(\Dc)$ grows as $d^j$. We leave an exploration of an efficient version of this two-step approach as future work.

\subsection{Practical solution via concentration inequalities and sketching}\label{sec:practical-ucb}

Noting that Problem~\ref{problem:optimization-problem} is captured by the concept of concentration of measure, we resort to results from that literature to find acceptable estimates of $\theta_i$'s. In particular, we obtain a solution via an application of the one-sided Chebyshev's inequality, resulting in the following lemma.

\begin{lemma}
    \label{lemma:optimal-threshold}
    Denote by $\mu_i$ and $\Sigma_i$ the mean and covariance of the distribution of $\mathcal{P}_i$.
    An upper-bound on the solution to Problem~\ref{problem:optimization-problem} for $\delta \in (0, 1)$ is:
    \begin{equation}
    \label{equation:cdf-inverse}
    \theta_i = \langle q, \mu_i \rangle + \sqrt{\frac{1+\delta}{1 - \delta} \; q^\top \Sigma_i q}.
    \end{equation}
\end{lemma}
\begin{proof}
    The result follows immediately by applying the one-sided Chebyshev's inequality to the distribution, $S_i$,
    of inner products between $q$ and points in $\mathcal{P}_i$:
    \begin{align*}
    \Pr_{s \sim S_i}[s - \langle q, \mu_i \rangle \leq \epsilon] &\geq 1 - \frac{q^\top \Sigma_i q}{q^\top \Sigma_i q + \epsilon^2} = \frac{1 + \delta}{2}
    \implies \epsilon^2 = \frac{1 + \delta}{1 - \delta} \; q^\top \Sigma_i q.
    \end{align*}
    Rearranging the terms to match the expression of Problem~\ref{problem:optimization-problem} gives $\theta_i = \langle q, \mu_i \rangle + \epsilon$, as desired.
\end{proof}

We emphasize that Lemma~\ref{lemma:optimal-threshold} gives a loose upper-bound on the optimal value of $\theta_i$. Because we make no assumptions about the data distribution, the gap between the optimal and estimated $\theta_i$ cannot be expressed in closed form. Had we assumed $S_i$ is sub-Gaussian, it would be easy to bound the gap between $\theta_i$ obtained from Lemma~\ref{lemma:optimal-threshold} and the one obtained from sub-Gaussian concentration bounds. Nonetheless, we find that in practice even this sub-optimal solution proves effective.

There is still one challenge with Equation~\eqref{equation:cdf-inverse}: $\Sigma_i$ can be too large to store as the matrix grows as $d^2$ for each partition. Contrast that with the cost of \mean and \normalizedmean which only need one $d$-dimensional vector per partition. To remedy this, we design a compact approximation of $\Sigma_i$ that can be plugged into Equation~(\ref{equation:cdf-inverse}) to replace $\Sigma_i$. We describe that next to complete our algorithm.

\subsubsection{Approximating the covariance matrix}
Since the procedure we describe is independently applied to each partition, we drop the subscript in $\Sigma_i$ and focus on a single partition. We seek a matrix $\Sigma^\ast \in \R^{d \times d}$ that approximates the positive semi-definite (PSD) matrix $\Sigma$, by minimizing the following standard approximation error:
\begin{equation}
    \label{equation:approximation-error}
    \text{err}(\Sigma, \Sigma^\ast) \triangleq \sup_{\substack{\|v\|_2 = 1}} | v^\top \Sigma v - v^\top \Sigma^\ast v |.
\end{equation}
As we seek a \emph{compact} $\Sigma^\ast$, we constrain the solution space to rank-$t$ matrices: $\text{rank}(\Sigma^\ast) = t$, $t \ll d$.

A standard solution is a result of the Eckhart-Young-Mirsky Theorem: Let $\Sigma = V\Lambda V^\top$ be the eigendecomposition of $\Sigma$. Then, $\Sigma^\ast = [V\sqrt{\Lambda}]_t[V\sqrt{\Lambda}]_t^\top$, where $[ \cdot ]_t$ selects the first $t$ columns of its argument, optimally approximates $\Sigma$ under the stated constraints. We denote this solution by $\Sigma^\textsc{lr}_t$.

We could stop here and use $\Sigma^\textsc{lr}_t$ in lieu of $\Sigma$ in Equation~(\ref{equation:cdf-inverse}). However, while that would be the optimal choice in \emph{the general case}, for real-world datasets, we show that we can find a better approximation.

\begin{definition}[Masked Sketch of Rank $t$]
Rewrite $\Sigma$ as follows, with $D$ its diagonal and $R=\Sigma - D$:
\begin{equation*}
    \Sigma = D + R = D^{\frac{1}{2}} (I + \underbrace{D^{-\frac{1}{2}} R D^{-\frac{1}{2}}}_{R_\circ}) D^{\frac{1}{2}}.
\end{equation*}
Let $Q\Lambda Q^\top$ be $R_\circ$'s eigendecomposition. We call the following the Masked Sketch of Rank $t$ of $\Sigma$:
\begin{align}\label{eqn:sketch}
\Sigma^\textsc{ms}_t = D + D^{\frac{1}{2}} [Q]_t[\Lambda]_t[Q]_t^\top D^{\frac{1}{2}}.
\end{align}
\end{definition}

\begin{algorithm}[t]
\footnotesize
\caption{Indexing and scoring a single partition with \optimist}
\label{alg:full}
\begin{algorithmic}[1]
\Require Partition $\mathcal{P}$ and target rank $t \ll d$.
\Procedure{BuildRouterForPartition}{$\mathcal{P}, t$}
\State $\mu \leftarrow \frac{1}{\lvert \mathcal{P} \rvert}\sum_{u \in \mathcal{P}} u$
\State $\Sigma \leftarrow \frac{1}{\lvert \mathcal{P} \rvert} \sum_{u \in \mathcal{P}} (u - \mu)(u - \mu)^\top$
\State $R_\circ = D^{-1/2} (\Sigma - D) D^{-1/2}$ \Comment{$D$ is the diagonal of $\Sigma$}
\State Find eigendecomposition $R_\circ = Q\Lambda Q^\top$
\State Sort columns of $Q$, $\Lambda$ in non-increasing order
\State $\Lambda_t \leftarrow [\Lambda]_t$, $Q_t \leftarrow [Q]_t$
\State \Return $\{ \mu, D, \Lambda_t, Q_t \}$
\EndProcedure
\Require Query $q \in \R^d$ and optimism parameter $\delta > 0$.
\Procedure{ScorePartitionForQuery}{$q, \delta$; $\{ \mu, D, \Lambda_t, Q_t \}$}
\State $\tilde{q} \leftarrow q \circ \text{diag}(D^{1/2})$ \Comment{Element-wise product}
\State \Return $\langle q, \mu \rangle + \sqrt{\frac{1+\delta}{1 - \delta} \cdot (\|\tilde{q}\|_2^2 + \tilde{q}^\top Q_t\Lambda_t Q_t^\top \tilde{q}) }$
\EndProcedure
\end{algorithmic}
\end{algorithm}

We next establish that, under certain assumptions, $\text{err}(\Sigma, \Sigma^\textsc{ms}_t)$ is lower than $\text{err}(\Sigma, \Sigma^\textsc{lr}_t)$.
 
\begin{lemma}\label{lemma:sketch-guarantee}
Let $\Sigma \in \R^{d \times d}$ be a PSD matrix with diagonal $D$ for which $\min_{i \in [d]} {D_{ii}}/{ \max_{i \in [d]} D_{ii}} \geq 1 - \epsilon$ for some $\epsilon > 0$. For every $1 \leq t \leq d-1$ such that the $(t+1)$-th eigenvalue of $D^{-1/2}\Sigma D^{-1/2}$ is greater than $1$, we have that: $\text{err}(\Sigma, \Sigma^\textsc{ms}_t) \leq \text{err}(\Sigma, \Sigma^\textsc{lr}_t) / (1 - \epsilon)$.
\end{lemma}
We give the proof in Appendix~\ref{appendix:lemma-sketch-error} and show that for datasets in this work the assumptions hold.

\subsubsection{The final algorithm}
Using our solution from Lemma \ref{lemma:optimal-threshold} and our sketch defined in \eqref{eqn:sketch}, we describe our full algorithm in Algorithm~\ref{alg:full} for building our router and scoring a partition. Notice that, since we only store $\mu, D, [\Lambda]_t$ and $[Q]_t$, the router requires $t + 2$ vectors\footnote{Since $[\Lambda]_t$ can be ``absorbed'' into $[Q]_t$ with some care taken for the signs of the eigenvalues.} in $\R^d$ per partition. In our experiments, we choose $t \leq 10$ for all datasets except one and show that much of the performance gains from using the whole covariance matrix can be preserved even by choosing a small value of $t$ independent of $d$.

\subsubsection{Time complexity analysis}
The time complexity of Algorithm~\ref{alg:full} is dominated by eigendecomposition which takes $\bigO{(nd^2 + d^3)}$ time. However, because we only need the top $t$ eigenvectors and values to build the router, we may use highly-efficient algorithms for computing low-rank approximations of PSD matrices such as the randomized Block Krylov Method (see~\citep[Section 6.3]{tropp2018analysis} and~\citep{musco2015randomized}).

The time complexity of these algorithms is proportional to $\bigO{(nd \log(d) \times t)}$, making them much faster than na\"ive eigendecomposition. Since $t$ is usually a small constant in our setting, Algorithm~\ref{alg:full} can be implemented in near linear time complexity in the size of the data.
\section{Experimental evaluation}
\label{section:experiments}

We put our arguments to the test and evaluate \optimist.

\subsection{Setup}
\label{section:experiments:setup}

\begin{table}
\parbox{.55\linewidth}{
    \footnotesize
    \caption{Dataset statistics (size $m$ and dimensions $d$), along with the number of partitions ($C$) and \optimist's default rank configuration ($t$) in our main experiments.}
    \label{table:router-configuration}
    \begin{center}
    \begin{sc}
    \begin{tabular}{lcc|cc}
    \toprule
    Dataset & $m$ & $d$ & $C$ & $t$ \\
    \midrule
    \textimage & $10$M & $200$ & $3{,}000$ & $4$ \\
    \music & $1$M & $100$ & $1{,}024$ & $2$ \\
    \deep & $10$M & $96$ & $3{,}000$ & $2$ \\
    \glove & $1.2$M & $200$ & $1{,}024$ & $4$ \\
    \msmarco & $8.8$M & $384$ & $3{,}000$ & $8$ \\
    \nq & $2.7$M & $1{,}536$ & $1{,}600$ & $30$ \\
    \bottomrule
    \end{tabular}
    \end{sc}
    \end{center}
}
\hfill
\parbox{.4\linewidth}{
    \footnotesize
    \caption{Relative savings during search to achieve a fixed recall. A saving of $x\%$ means that \optimist searches $x\%$ fewer points than \normalizedmean.}
    \label{table:savings}
    \begin{center}
    \begin{sc}
    \begin{tabular}{l|cc}
    \toprule
    Recall & $90\%$ & $95\%$ \\
    \midrule
    \textimage & $23\%$ & $22\%$ \\
    \music & $38\%$ & $54\%$ \\
    \glove & $11\%$ & $5.5\%$ \\
    \msmarco & $22\%$ & $7.7\%$ \\
    \nq & $18\%$ & $20\%$ \\
    \bottomrule
    \end{tabular}
    \end{sc}
    \end{center}
}
\end{table}

\noindent \textbf{Datasets}: Table~\ref{table:router-configuration} summarizes in the leftmost block the main properties of the datasets used in our experiments. A complete description can be found in Appendix~\ref{appendix:datasets}.

\noindent \textbf{Clustering}: For our main results, we partition the datasets with spherical KMeans~\citep{sphericalKMeans}. We include in Appendices~\ref{appendix:experiments:standard-kmeans} and~\ref{appendix:experiments:gmm} results from similar experiments but where the clustering algorithm is standard KMeans and Gaussian Mixture Model (GMM). We cluster each dataset into $C=\sqrt{m}$ shards, where $m$ is the number of data points in the dataset.

\bigskip
\noindent \textbf{Evaluation}: The independence of routing from scoring simplifies the evaluation protocol: By additivity of latency, efficiency gains from a router translate directly into efficiency gains in the end-to-end search (i.e., routing followed by scoring). We can therefore fix the scoring algorithm and examine routers in isolation to compare their accuracy, latency, and other characteristics.

Once a dataset has been partitioned, we fix the partitioning and evaluate all routers on it to facilitate a fair comparison. We evaluate each router $\tau$ as follows. For each test query, we identify the set of shards to probe using $\tau$. We then perform an exact search over the selected shards, obtain the top-$k$ points, and compute recall with respect to the ground-truth top-$k$ set. Because the only source of error is the router's inaccuracy, the measured recall gauges the effectiveness of $\tau$.

We report recall as a function of the number of \emph{data points} probed, rather than the number of \emph{shards probed}. In this way, a comparison of the efficacy of different routers is unaffected by any imbalance in shard sizes, so that a router cannot trivially outperform another by simply prioritizing larger shards.

\bigskip
\noindent \textbf{Routers}: We evaluate the following routers in our experiments:
\begin{itemize}[leftmargin=*]
    \item \mean and \normalizedmean: Defined in Equations~(\ref{equation:mean-router}) and~(\ref{equation:normalized-mean-router}).
    \item \scann$(T)$: Similar to \mean and \normalizedmean, but where routing is determined by inner product between a query and the \scann centroids (c.f., Theorem~$4{.}2$ in~\citep{scann}). \scann has a single hyperparameter $T$, which we set to $0.5$ after tuning.
    \item \subpartition$(t)$: \optimist stores $t + 2$ vectors per partition, where $t$ is the rank in Algorithm~\ref{alg:full}. Following the \textsc{kRt} method of~\cite{gottesburen2024unleashinggraphpartitioninglargescale}, we introduce another baseline where we partition each shard into $t + 2$ sub-shards and use the their centroids as the shard's representatives. Routing is based on the maximum inner product of the query with a shard's representatives.
    \item \optimist$(t, \delta)$: $t$ and $\delta$ are parameters of Algorithm~\ref{alg:full}. By default, we set $\delta=0.8$ and $t$ to values in Table~\ref{table:router-configuration}, but study their effect in Appendix~\ref{appendix:optimist-hyperparameters}. If unspecified, it should be understood that default parameters are used. We write \optimist$(t=d, \cdot)$ to indicate the use of the full covariance matrix.
\end{itemize}

\textbf{Code}: We have implemented all baseline and proposed routers in the Rust programming language. We have open-sourced\footnote{Available at \url{https://github.com/Artificial-Memory-Lab/optimist-router}} our code along with experimental configuration to facilitate reproducibility.

\begin{figure}[t]
\begin{center}
\centerline{
    \subfloat[\glove]{
        \includegraphics[width=0.3\linewidth]{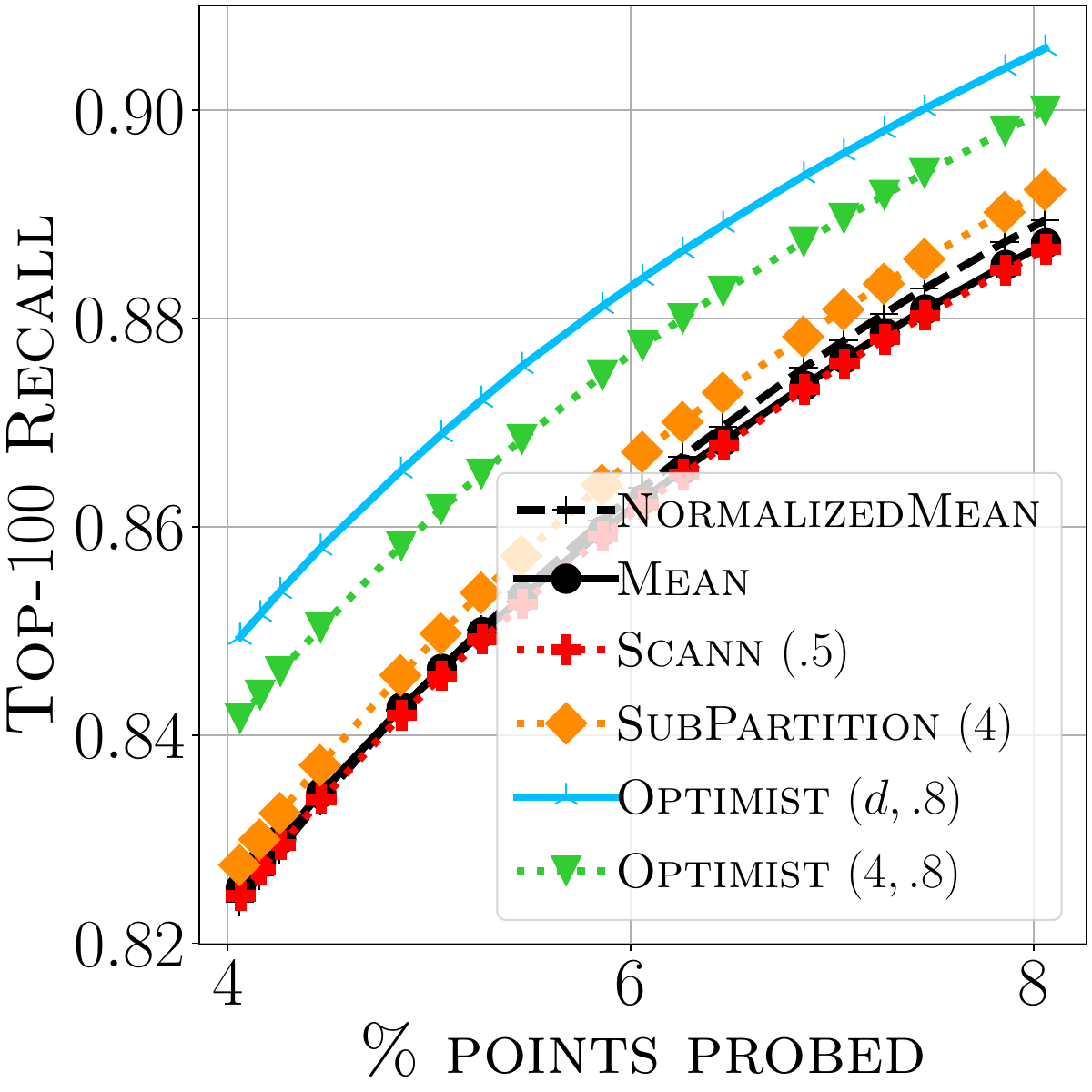}
    }
    \subfloat[\msmarco]{
        \includegraphics[width=0.3\linewidth]{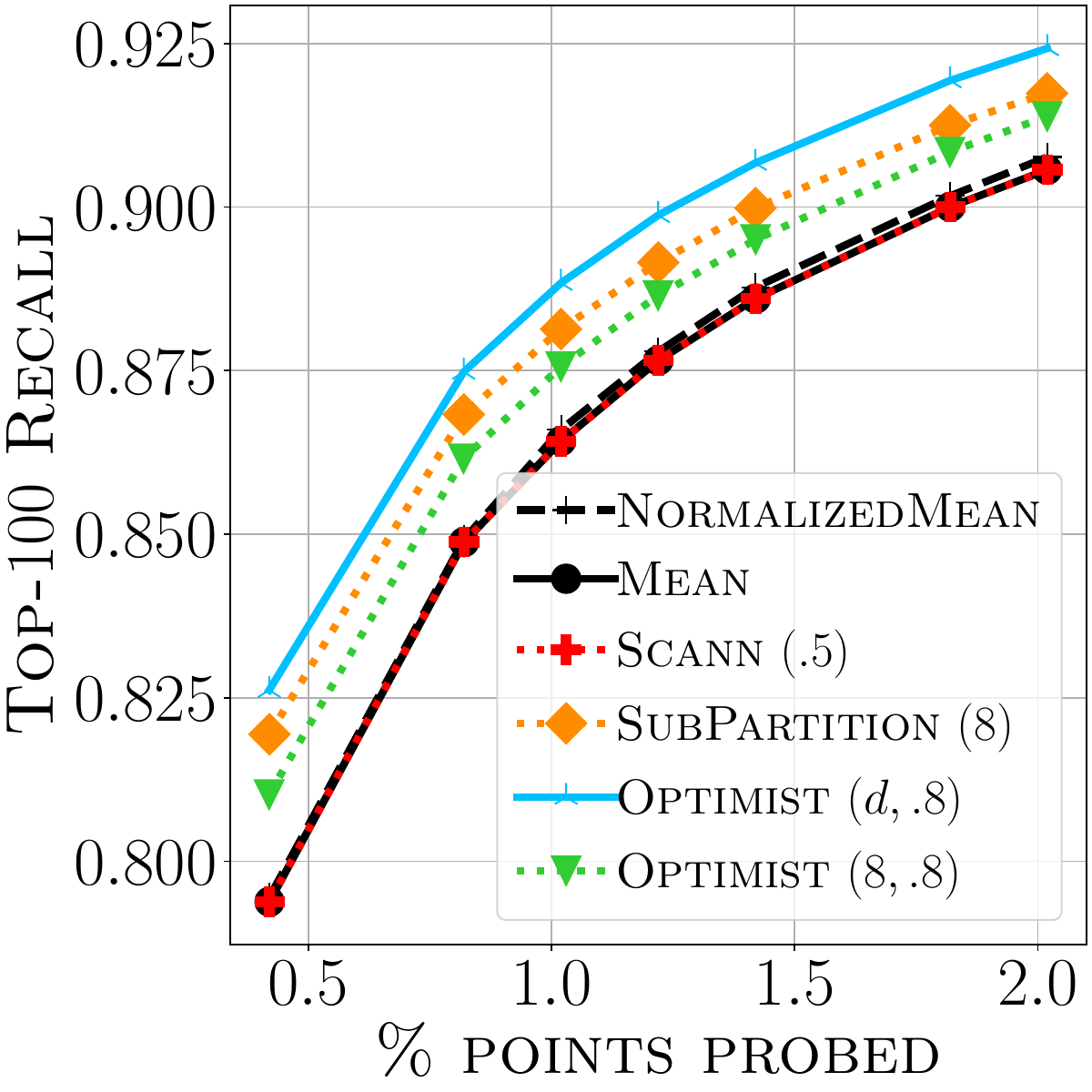}
    }
    \subfloat[\music]{
        \includegraphics[width=0.3\linewidth]{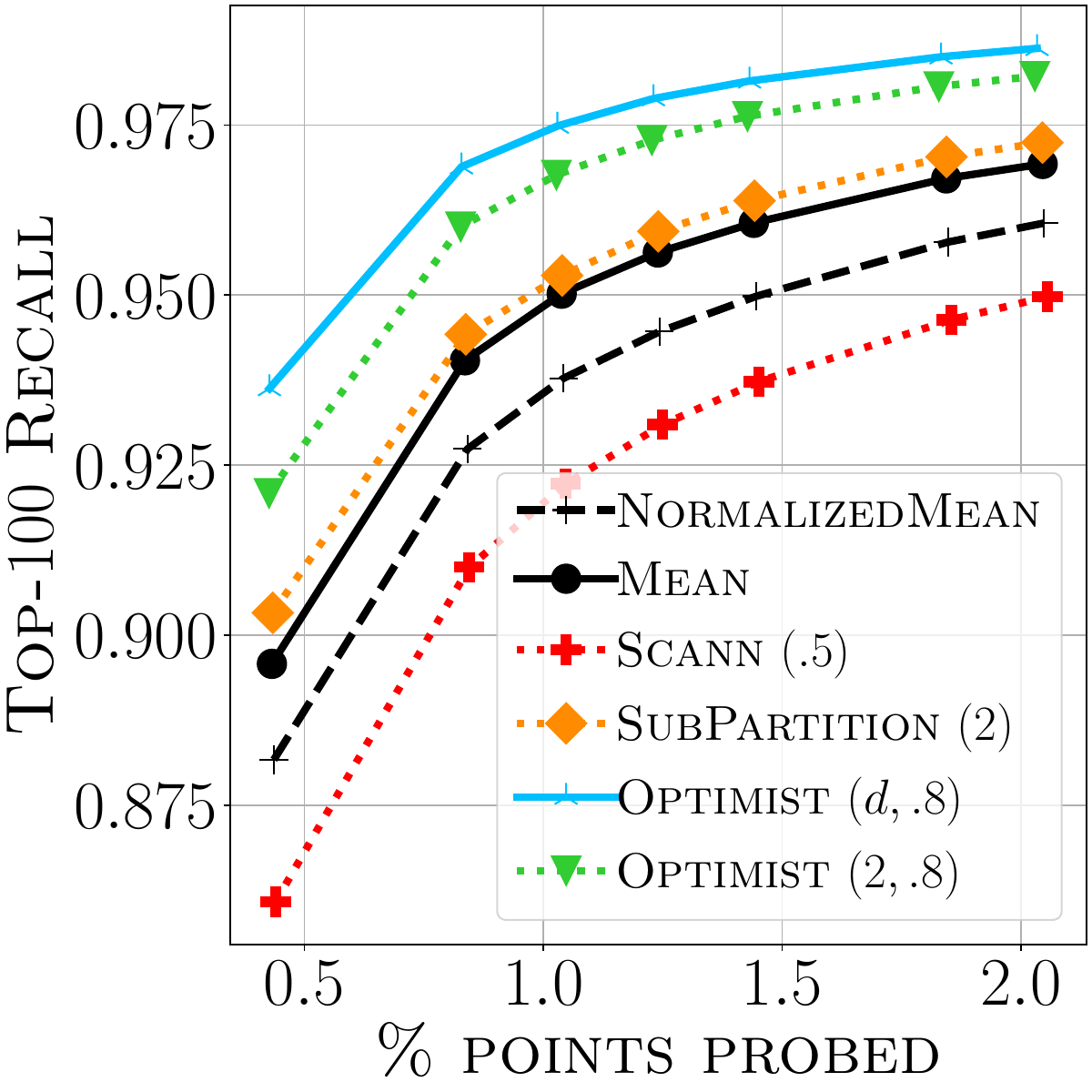}
    }
}
\caption{Top-$100$ recall vs. volume of probed points. Partitioning is by spherical KMeans. \scann has parameter $T$, \subpartition $t$, and \optimist rank $t$ and degree of optimism $\delta$.}
\label{figure:spherical-kmeans:top-100}
\end{center}
\vspace{-0.5cm}
\end{figure}

\subsection{Main results}
\label{section:experiments:results}

\noindent \textbf{Effectiveness}: Figure~\ref{figure:spherical-kmeans:top-100} plots recall versus the volume of points examined for select datasets using spherical KMeans. See Appendix~\ref{appendix:experiments:spherical-kmeans} for full results; Appendix~\ref{appendix:experiments:standard-kmeans} for standard KMeans; and Appendix~\ref{appendix:experiments:gmm} for GMM.

Among baselines, \normalizedmean generally outperforms \mean and \scann, except on \music. \optimist with the full covariance ($t=d$) generally does at least as well as \normalizedmean, but often outperforms it significantly. Interestingly, the gains from \optimist widen as $k$ increases. Finally, while \optimist with a rank-$t$ Masked Sketch shows some degradation, it still yields a higher recall than baselines for larger $k$. \subpartition$(t)$ becomes a strong competitor when $k$ is small.

\optimist shines when data points have varying norms. On \music, at $95\%$ top-$100$ recall, \optimist needs to probe $54\%$ fewer points than \normalizedmean; on average \optimist probes $6{,}666$ points to reach $95\%$ top-$100$ recall whereas \normalizedmean examines $14{,}463$ points.

Table~\ref{table:savings} presents savings on all datasets. On \deep, \optimist scans $9\%$ \emph{more} points than \normalizedmean to reach $90\%$ recall ($55{,}000$ vs $60{,}000$), and $10\%$ more to reach $95\%$ ($95{,}000$ vs $105{,}000$). We suspect this is an artifact of the dataset's construction: each point is represented by the top $96$ principal components of its original features, leading to unusual partition statistics.

\begin{figure*}[t]
\begin{center}
\centerline{
    \subfloat[SSD]{
        \includegraphics[width=0.35\linewidth]{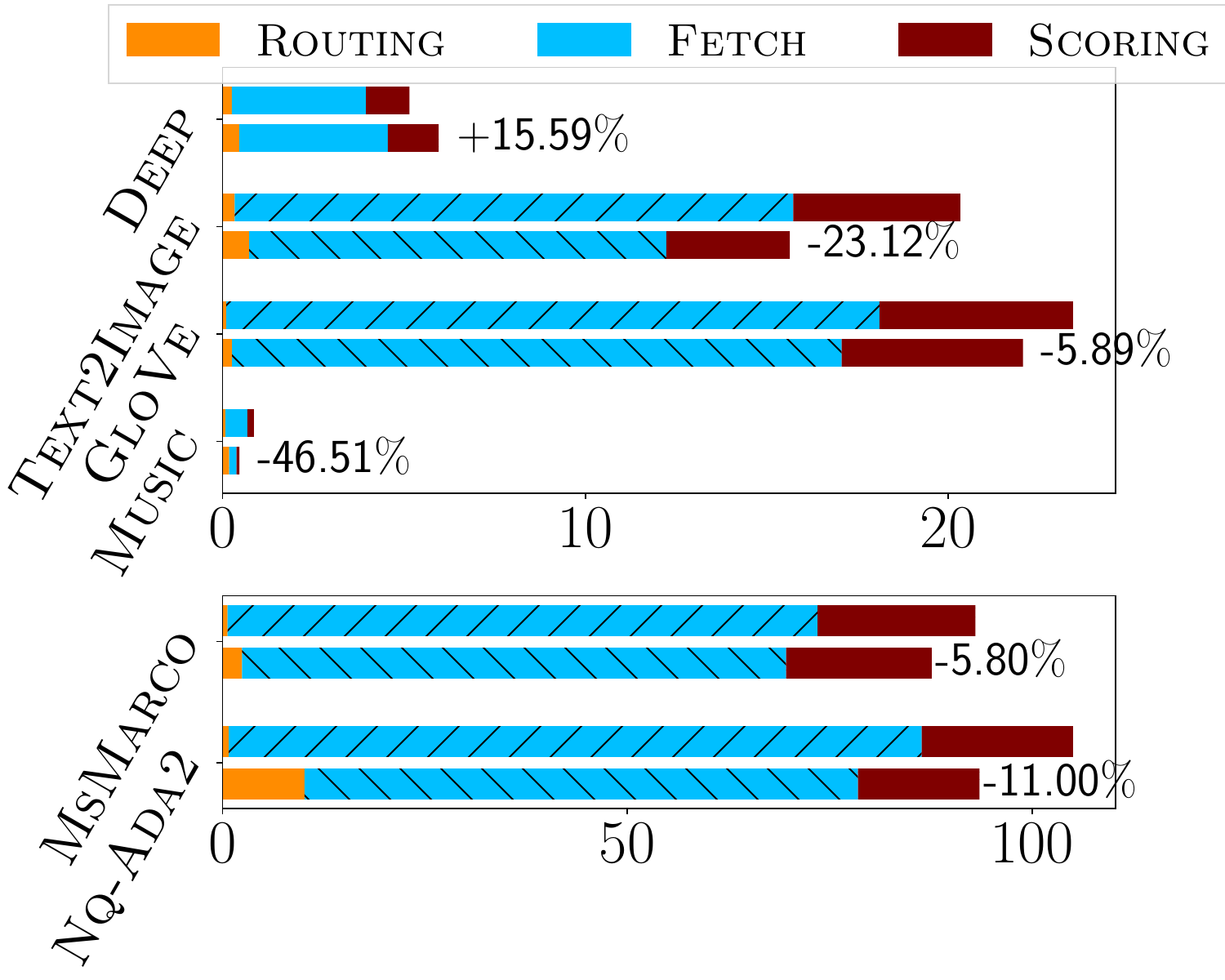}
    }
    \subfloat[Blob Storage]{
        \includegraphics[width=0.35\linewidth]{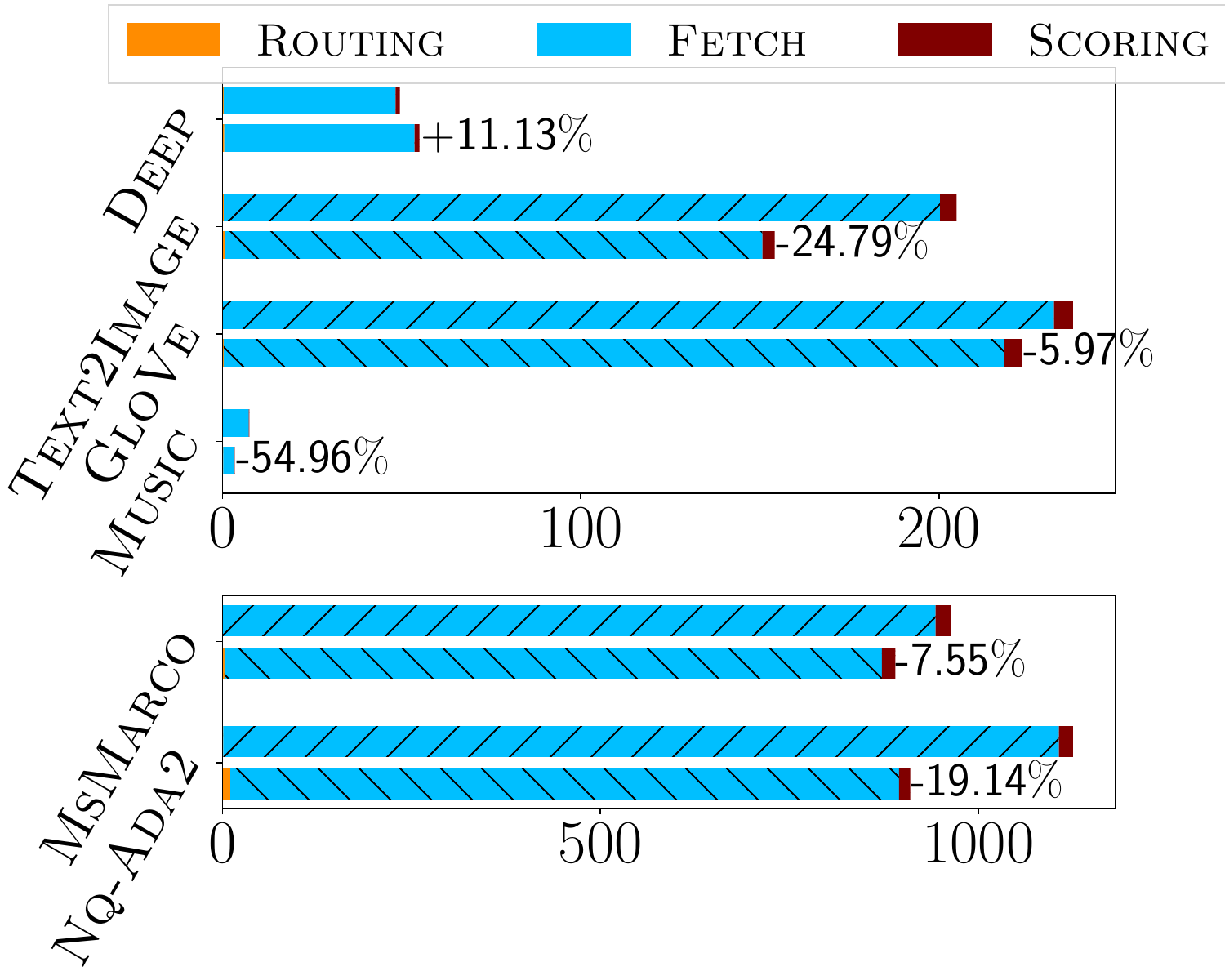}
    }
}
\caption{Mean latency (ms) to reach $95\%$ recall when PQ-compressed shards are on SSD and blob storage. For each dataset, we plot the latency breakdown for \normalizedmean (top bar) and \optimist (bottom), and report relative gains (negative value indicates gain by \optimist).}
\label{figure:latency}
\end{center}
\end{figure*}

\textbf{Latency}: We now turn to latency which includes routing, fetching the chosen shards from storage, and scoring using PQ with $4$-dimensional codebooks. Figure~\ref{figure:latency} reports latencies for \optimist and \normalizedmean. \optimist's gains are more pronounced when shards are stored on blob storage, but even on SSD the gains are substantial. Full results are in Appendix~\ref{appendix:latency}.

We run experiments on AWS c5.xlarge ($4$ vCPUs, $8$GB memory). Baseline bandwidth of the SSD attached to this machine is $1{,}150$Mbps.\footnote{See \url{https://docs.aws.amazon.com/AWSEC2/latest/UserGuide/ebs-optimized.html}} Finally, using $4$ threads, transferring $4$MB of data from blob storage (hosted on Amazon S3) to main memory has P50 latency of $45$ milliseconds (ms).\footnote{See \url{https://github.com/dvassallo/s3-benchmark} for an independent benchmark.}

Let us next present in Table~\ref{table:differences} a unified view of Figure~\ref{figure:latency}, which reports a breakdown of latency at $95\%$ recall, and Table~\ref{table:savings}, which shows the relative savings (in terms of the number of data points probed) between \optimist and \normalizedmean. It is clear that, with a few exceptions, the overall latency savings are roughly equal to the reduction in the number of points probed. As such, change in the number of points probed is a reasonable proxy to change in overall latency.

\begin{table}[t]
    \caption{Difference between \optimist and \normalizedmean in terms of number of points probed and latency. Negative percentages means \optimist leads to gains.}
    \label{table:differences}
    \begin{center}
    \begin{sc}
        \begin{tabular}{lccc}
        \toprule
        Dataset & \# points probed & Latency (Blob) & Latency (SSD) \\
        \midrule
        \textimage & -22\% & -24.8\% & -23\% \\
        \music & -54\% & -55\% & -46.5\% \\
        \deep & +10\% & +11\% & +15.6\% \\
        \glove & -5.5\% & -6\% & -6\% \\
        \msmarco & -7.7\% & -7.6\% & -5.8\% \\
        \nq & -20\%	& -19.1\% & -11\% \\
        \bottomrule
        \end{tabular}
    \end{sc}
    \end{center}
\end{table}

Hardware plays an outsize role in latency improvements. For example, a more limited storage bandwidth would boost \optimist's standing. If bandwidth is abundant and number of threads limited, however, \optimist's advantage would become less significant. Other factors that affect latency comparisons include: target recall level; type of compression; and algorithm used for scoring.

\noindent \textbf{Router size}: On most datasets, the router size difference between \normalizedmean ($1$ vector) and \optimist ($t$ vectors) is negligible. On \glove, for example, it is $0.8$MB versus $4.7$MB ($t=4$) in total. Statistics on other datasets are as follows: \msmarco, $4.4$MB vs. $44$MB ($t=8$); \music, $0.4$MB vs. $1.6$MB ($t=2$); \textimage, $2.3$MB vs. $13.7$MB ($t=4$); \deep, $1.1$MB vs. $4.4$MB ($t=2$). The difference is larger on \nq: $9.4$MB vs $300$MB ($t=30$).

\subsection{Maximum inner product prediction}

\begin{figure*}[ht]
\begin{center}
\centerline{
    \subfloat[\glove]{
        \includegraphics[width=0.3\linewidth]{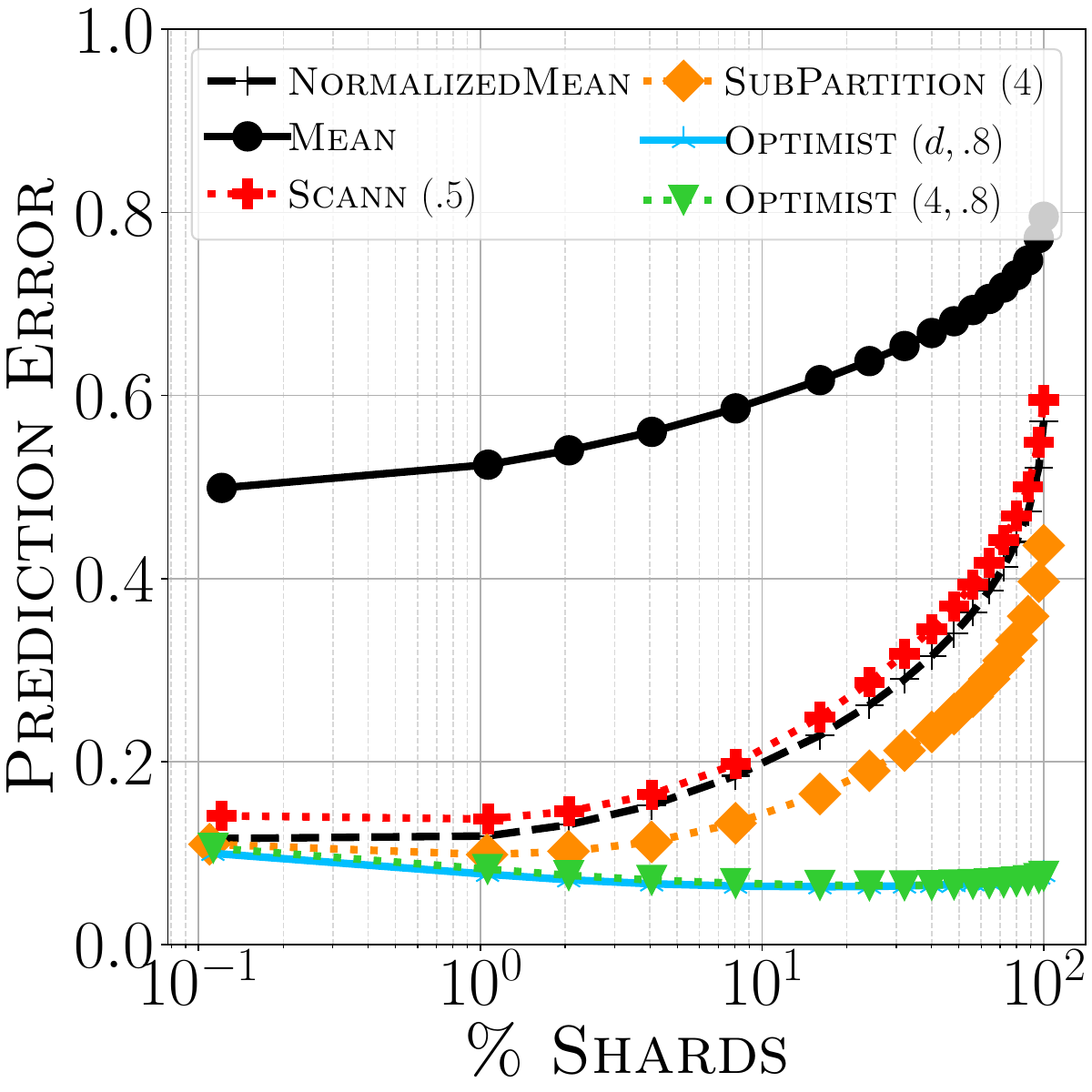}
    }
    \subfloat[\deep]{
        \includegraphics[width=0.3\linewidth]{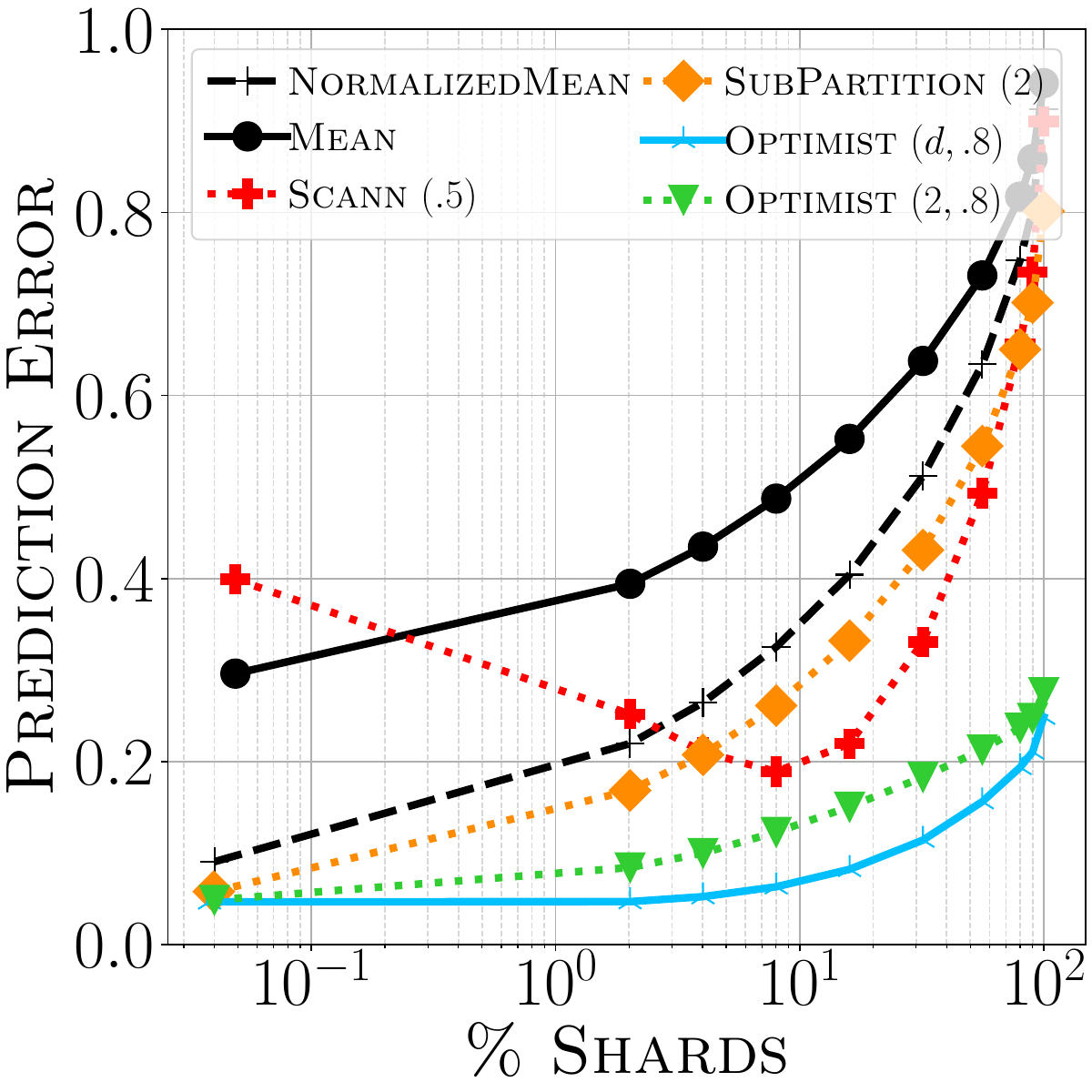}
    }
    \subfloat[\textimage]{
        \includegraphics[width=0.3\linewidth]{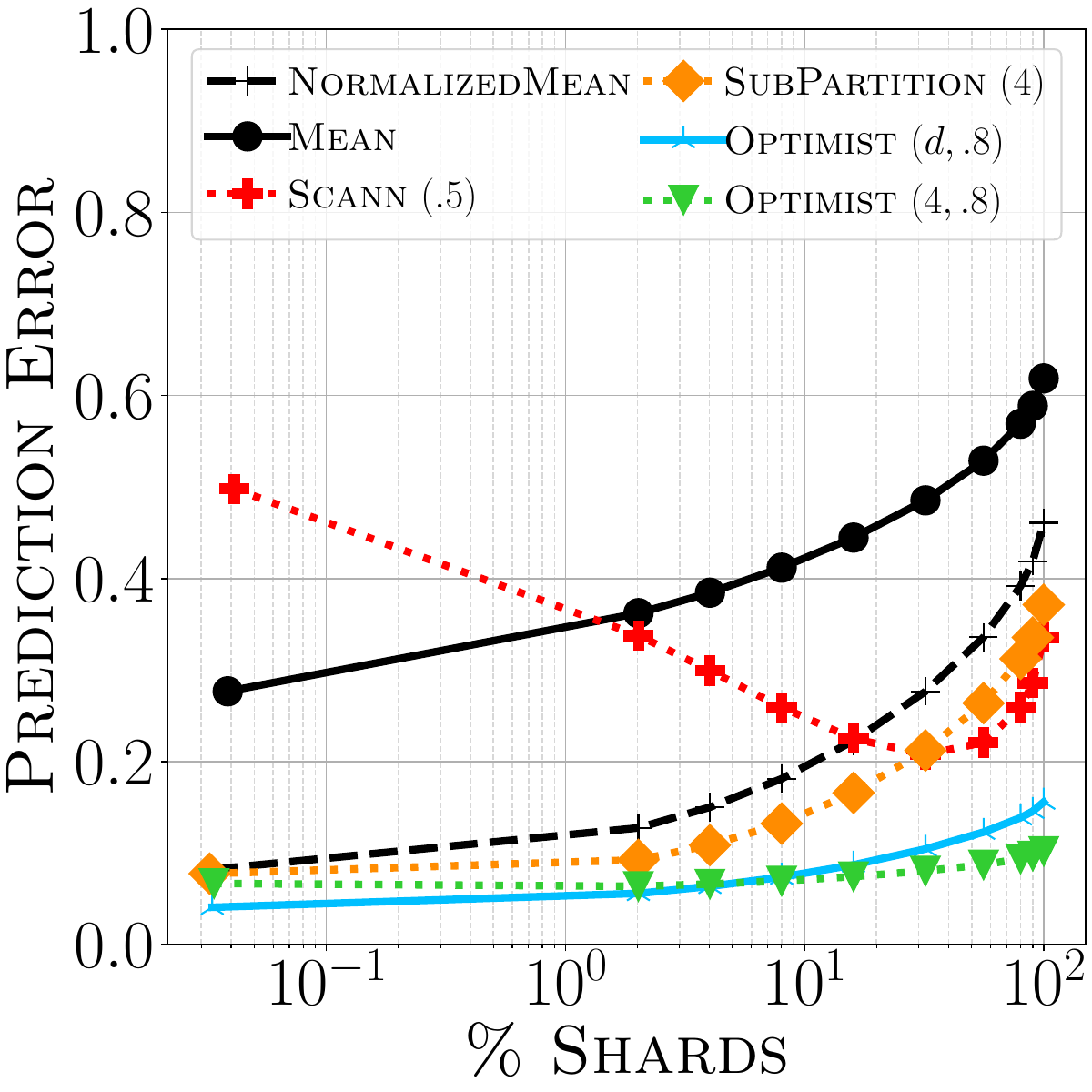}
    }
}
\caption{Mean prediction error $\mathcal{E}_\ell(\tau, \cdot)$ of Equation~(\ref{equation:general-prediction-error}) versus $\ell$ (percent shards).}
\label{figure:prediction-error-breakdown}
\end{center}
\end{figure*}

We claimed that \optimist is statistically principled. It should thus give more accurate estimates of the maximum inner product for all query-partition pairs. We examine that claim in this section. 

Fix a dataset with partitions $\mathcal{P}_i$, router $\tau$, and query $q$. Write $\tau_i$ for the score computed by $\tau$ for $\mathcal{P}_i$ and $q$. $\tau_i$'s induce an ordering $\pi$ among $\mathcal{P}_i$'s, so that $\tau_{\pi_i} \geq \tau_{\pi_{i+1}}$. We quantify the inner product prediction error as follows, for $\ell \in [C]$:
\begin{equation}
\label{equation:general-prediction-error}
    \mathcal{E}_\ell(\tau, q) = \frac{1}{\ell}
        \sum_{i=1}^{\ell} \left\vert \frac{\tau_{\pi_i}}{\max_{u \in \mathcal{P}_{\pi_i}} \langle q, u \rangle}  - 1 \right\vert.
\end{equation}
This error is $0$ when $\tau_i$'s perfectly match the maximum inner product of $q$ and $\mathcal{P}_i$'s. $\ell$ allows us to factor in the \emph{rank} of partitions: we can measure the error only for the top $\ell$ shards according to $\tau$. In this way, if we decide that it is not imperative for a router to accurately predict the maximum inner product in low-ranking shards, we can reflect that choice in our calculation.

We measure Equation~(\ref{equation:general-prediction-error}) on all datasets partitioned by spherical KMeans, and all routers considered in this work. Figure~\ref{figure:prediction-error-breakdown} reports the results for select datasets---see Appendix~\ref{appendix:max-ip-prediction}---where for each choice of $\ell$, we plot $\mathbb{E}_q [\mathcal{E}_\ell(\tau, q)]$ for all queries. For most routers error grows as $\ell \rightarrow C$. $\optimist(t=d, \cdot)$ degrades less severely. Remarkably, \music excepted, when $t \ll d$, the same pattern persists.

\section{Concluding remarks}
\label{section:conclusion}

We studied clustering-based maximum inner product search where points are clustered into shards during indexing. Search involves two independent subroutines: a \emph{router} computes a score for each shard with respect to the query and returns the top $\ell$ shards to probe; then a \emph{scorer} computes (approximate) scores for points in those $\ell$ shards. We considered a storage-backed system where shards do not rest in memory, but are stored on external storage and must be fetched into memory when a router identifies them. This is an important paradigm for nearest neighbor search at scale.

Within this framework, we focused on routing. We motivated our work by an unusual routing behavior: \normalizedmean computes a score that over-estimates the maximum inner product between a query and points in a shard. Interestingly, the extent of over-estimation correlates with the variance of the shard: The more spread-out the points are, the more optimistic the router becomes.

We took the insight that variance should play a role in routing and developed a principled optimistic algorithm, called \optimist, whose score for a shard is a more accurate estimate of the maximum inner product---as confirmed in Figure~\ref{figure:prediction-error-breakdown}, resulting in smaller $\ell$'s to achieve a fixed recall. As Figure~\ref{figure:latency} shows \optimist is particularly attractive when shards rest on some external, high-latency storage.

\optimist has implications beyond latency. Transferring a smaller volume of data to memory means that each query needs a smaller in-memory cache. As a result, cache can be shared among more queries, leading to a higher throughput in systems where memory availability is the main bottleneck.

Our research takes a first step in exploring unsupervised routing for clustering-based MIPS, and identifies a trade-off space that has not been explored before. We have more to do, however. We leave to future work an exploration of a more compact sketch of the covariance; and, an efficient realization of our general solution outlined in Section~\ref{section:general-solution}.

\newpage
\textbf{Acknowledgements}. 
This work was partially supported by the Horizon Europe RIA ``EFRA - Extreme Food Risk Analytics'' (grant agreement n. 101093026) and the PNRR - M4C2 - Investimento 1.3, Partenariato Esteso PE00000013 - ``FAIR - Future Artificial Intelligence Research'' - Spoke 1 ``Human-centered AI''. Horizon Europe and the PNRR programs are funded by the European Commission under the NextGeneration EU program. The views and opinions expressed are solely those of the authors and do not necessarily reflect those of the European Union, nor can the European Union be held responsible for them.

\bibliographystyle{plainnat}
\bibliography{paper}

\newpage
\appendix
\newpage
\section*{NeurIPS Paper Checklist}

\begin{enumerate}

\item {\bf Claims}
    \item[] Question: Do the main claims made in the abstract and introduction accurately reflect the paper's contributions and scope?
    \item[] Answer: \answerYes{} 
    \item[] Justification: We are explicit in our abstract and introduction about the prior research that motivated this work and describe precisely what our novel contributions are.

\item {\bf Limitations}
    \item[] Question: Does the paper discuss the limitations of the work performed by the authors?
    \item[] Answer: \answerYes{} 
    \item[] Justification: We noted in our discussion the limitations of our work. First, we emphasized the sub-optimality of the solution obtained from Lemma~\ref{lemma:optimal-threshold}, but justified its efficacy through empirical experiments. Second, we pointed out that our router, in isolation, is computationally more expensive and requires a larger amount of storage, but explained and demonstrated why this increase in costs can be appropriate depending on the end-to-end setup. Finally, our method performs better when data is stored in a high-latency storage medium, rather than in-memory; we have made a note of that in our discusison.

\item {\bf Theory assumptions and proofs}
    \item[] Question: For each theoretical result, does the paper provide the full set of assumptions and a complete (and correct) proof?
    \item[] Answer: \answerYes{} 
    \item[] Justification: We state all assumptions and complete proofs for all claims made in this work. For some results, we move the proof to the appendix due to space constraints.

    \item {\bf Experimental result reproducibility}
    \item[] Question: Does the paper fully disclose all the information needed to reproduce the main experimental results of the paper to the extent that it affects the main claims and/or conclusions of the paper (regardless of whether the code and data are provided or not)?
    \item[] Answer: \answerYes{} 
    \item[] Justification: We have included all the details of our experiments (including all hyperparameters) as well as a complete description of datasets to facilitate reproducibility. Furthermore, when the work is ready for publication, we plan to release our code and configuration files to open-source. 

\item {\bf Open access to data and code}
    \item[] Question: Does the paper provide open access to the data and code, with sufficient instructions to faithfully reproduce the main experimental results, as described in supplemental material?
    \item[] Answer: \answerYes{} 
    \item[] Justification: We have created a double-blind review-compliant Git repository to host all the required code to reproduce every experiment reported in this work (including supplementary figures). In the repository, we have also included links to processed datasets. These links point to data that is stored in a Google Storage Bucket, and made available in such a way that neither reveals the identity of the authors, nor does it require viewers to log into Google services. The repository can be found at~\url{https://github.com/Artificial-Memory-Lab/optimist-router}.

\item {\bf Experimental setting/details}
    \item[] Question: Does the paper specify all the training and test details (e.g., data splits, hyperparameters, how they were chosen, type of optimizer, etc.) necessary to understand the results?
    \item[] Answer: \answerYes{} 
    \item[] Justification: We give all details including hyperparameters. As for train-test splits, the benchmark datasets used in our work come with a test query set that is separate from the data points.

\item {\bf Experiment statistical significance}
    \item[] Question: Does the paper report error bars suitably and correctly defined or other appropriate information about the statistical significance of the experiments?
    \item[] Answer: \answerYes{} 
    \item[] Justification: While we do not explicitly mention the results of statistical significance tests to avoid clutter, we note here that wherever \optimist performs better than the baselines, the differences are statistically significant with a $p$-value that is often less than $0.001$.

\item {\bf Experiments compute resources}
    \item[] Question: For each experiment, does the paper provide sufficient information on the computer resources (type of compute workers, memory, time of execution) needed to reproduce the experiments?
    \item[] Answer: \answerYes{} 
    \item[] Justification: Most of our experiments are not compute-heavy and are run on a commercial laptop with standard configuration. For experiments that study the latency of different methods, we state the specifications of the hardware used.
    
\item {\bf Code of ethics}
    \item[] Question: Does the research conducted in the paper conform, in every respect, with the NeurIPS Code of Ethics \url{https://neurips.cc/public/EthicsGuidelines}?
    \item[] Answer: \answerYes{} 
    \item[] Justification: We acknowledge that we have reviewed the NeurIPS code of ethics and that our research conforms, in every respect, with the code.

\item {\bf Broader impacts}
    \item[] Question: Does the paper discuss both potential positive societal impacts and negative societal impacts of the work performed?
    \item[] Answer: \answerNA{} 
    \item[] Justification: Our work does not have any societal impact that we could identify.
    
\item {\bf Safeguards}
    \item[] Question: Does the paper describe safeguards that have been put in place for responsible release of data or models that have a high risk for misuse (e.g., pretrained language models, image generators, or scraped datasets)?
    \item[] Answer: \answerNA{} 
    \item[] Justification: Our work does not involve the development of models or the release of new data.

\item {\bf Licenses for existing assets}
    \item[] Question: Are the creators or original owners of assets (e.g., code, data, models), used in the paper, properly credited and are the license and terms of use explicitly mentioned and properly respected?
    \item[] Answer: \answerYes{} 
    \item[] Justification: We have included this information in our full description of datasets used in our experiments.

\item {\bf New assets}
    \item[] Question: Are new assets introduced in the paper well documented and is the documentation provided alongside the assets?
    \item[] Answer: \answerNA{} 
    \item[] Justification: Our research does not result in new assets.

\item {\bf Crowdsourcing and research with human subjects}
    \item[] Question: For crowdsourcing experiments and research with human subjects, does the paper include the full text of instructions given to participants and screenshots, if applicable, as well as details about compensation (if any)? 
    \item[] Answer: \answerNA{} 
    \item[] Justification: Our research does not involve crowdsourcing or human subjects.

\item {\bf Institutional review board (IRB) approvals or equivalent for research with human subjects}
    \item[] Question: Does the paper describe potential risks incurred by study participants, whether such risks were disclosed to the subjects, and whether Institutional Review Board (IRB) approvals (or an equivalent approval/review based on the requirements of your country or institution) were obtained?
    \item[] Answer: \answerNA{} 
    \item[] Justification: Our research does not involve human subjects and as such does not require IRB review.

\item {\bf Declaration of LLM usage}
    \item[] Question: Does the paper describe the usage of LLMs if it is an important, original, or non-standard component of the core methods in this research? Note that if the LLM is used only for writing, editing, or formatting purposes and does not impact the core methodology, scientific rigorousness, or originality of the research, declaration is not required.
    \item[] Answer: \answerNA{} 
    \item[] Justification: We do not use LLMs in any shape or form (including for writing, editing, or formatting) in our work.

\end{enumerate}

\newpage
\section{An illustrative example}
\label{appendix:illustrative-example}
Interestingly, depending on operational factors such as data transfer rates and memory utilization, it would be acceptable for routing to be computationally more expensive as long as it identifies shards more accurately. Let us consider an example to support this statement and motivate our research.

Take Natural Questions~\citep{kwiatkowski-etal-2019-natural} embedded with \textsc{Ada-002}\footnote{\url{https://openai.com/index/new-and-improved-embedding-model/}} denoted \nq, containing $2.7$ million $1{,}536$-dimensional points. Applying KMeans and compressing with the typical $4$-bit PQ codebooks give shards with about $1{,}650$ points each and $1$MB in size. Consider next a commercial machine: AWS c5.xlarge ($4$ vCPUs, $8$GB of memory). Using $4$ threads, moving $4$MB of data from storage (hosted on Amazon's S3) to memory has a mean latency of $45$ milliseconds (ms).\footnote{See \url{https://github.com/dvassallo/s3-benchmark} for an independent and comprehensive benchmark.}

Given this setup, moving $16$ \nq shards from blob storage to the machine's memory takes $180$ms (P50 latency). On this same machine, routing using Equation~(\ref{equation:mean-router}) takes $0.76$ms. As such, so long as a more complex router can achieve the same accuracy as Equation~(\ref{equation:mean-router}) by fetching $16$ fewer shards, but by introducing a latency of no greater than $179$ms, the overall efficiency of ANN search improves. As we show later, our router on the same dataset with the same configuration takes only about $10.1$ms to identify shards, leading to a saving of $170.6$ms per query in the given example.

As this example illustrates, storage-backed, clustering-based ANN search offers a trade-off between not just accuracy and speed, but also other efficiency factors such as I/O bandwidth, and memory and cache utilization. This opens the door to more nuanced research. Our work is a step in that direction.

\section{Related work}
\label{appendix:related-work}

As we argued in Section~\ref{section:introduction}, routing accuracy in clustering-based approximate nearest neighbor search is increasingly relevant. Surprisingly, it has received little attention in the literature and most practical implementations of routing do not go beyond the na\"ive form of Equation~\eqref{equation:mean-router}. In this section, we briefly review the relevant methods that explore this particular topic.

The literature on routing functions can be split into two categories: supervised and unsupervised methods. In the supervised regime, a routing function is learned using a training query distribution. This is best demonstrated by the work of~\cite{vecchiato2024learningtorank}, which formulates the problem of routing as a learning-to-rank task: Given a query the function learns to rank shards, and subsequently uses the learned routing function during search for a test distribution.

That work is similar in spirit to the supervised methods put forth by~\cite{learning-to-index},~\cite{Dong2020Learning}, and~\cite{learned-index-metric-anns}. The difference is that, this set of methods attempt to learn to \emph{partition} the data and form representatives based on the learned functions.

Supervised methods have one caveat: While the methods differ in their approach to the problem, they require a training distribution. Our method, on the other hand, is completely unsupervised and does not have this limitation. Because of this fundamental difference, we do not believe an empirical or theoretical comparison between supervised methods and our method is warranted. In fact, in many instances, the benchmark datasets do not come with a training distribution or have very limited number of training queries.

There are, however, other unsupervised methods in the literature that we have included in our experiments as baselines per Section~\ref{section:experiments:setup} and that we review next in more detail. The first two are the \mean and \normalizedmean routers, which are the \emph{de facto} routing functions in all open-source approximate nearest neighbor search software. These follow the simple form of Equations~\eqref{equation:mean-router} and~\eqref{equation:normalized-mean-router}.

While \scann is proposed as a quantization method for maximum inner product search~\citep{scann}, it implicitly introduces a novel (supervised and unsupervised) routing function---we focus on the unsupervised function. \scann is based on the idea that, points in a shard should not count equally towards the quantization error. That is unlike Product Quantization~\cite{pq}. Instead, every point is weighted based on how likely it is to maximize inner product with an arbitrary query.

What is relevant to this work is that the \scann quantization method produces a shard representative that is a single vector, but that is not the centroid or normalized centroid (c.f., Theorem~$4{.}2$ in~\citep{scann}). We use this representative to route queries to shards.

Another relevant work is the routing functions proposed by~\cite{gottesburen2024unleashinggraphpartitioninglargescale}. They introduce two routing protocols: \textsc{kRt} and \textsc{hRt}. The latter has strong theoretical guarantees, but both methods perform equally well, with \textsc{kRt} having a slight advantage. As such, our review focuses on \textsc{kRt}.

\textsc{kRt}, which we call \subpartition in Section~\ref{section:experiments} for clarity, is based on the idea of partitioning each shard into multiple sub-shards and extracting a representative from each sub-shard. This is particularly useful for shards that are far larger than a conventional configuration of clustering-based ANN search would produce. At query time, shards are ranked by a statistic based on the inner products between the query and the shard representatives. In our instantiation of this routing function, we set that statistic to be the mean of the sub-shards.

\section{Proof of Lemma~\ref{lemma:sketch-guarantee} and justification of its assumptions}
\label{appendix:lemma-sketch-error}

\newcommand{\err}{\text{err}}
Let $\Sigma = D + R$ be the decomposition of the PSD covariance matrix $\Sigma$ into its diagonal $D$ and residual $R = \Sigma - D$. Let $Q\Lambda Q^\top$ be the orthogonal eigendecomposition of $R_\circ = D^{-1/2}RD^{-1/2}$. Recall that we define $\Sigma_t^\textsc{ms}$, for some $1 \leq t \leq d$, as 
$$\Sigma_t^\textsc{ms} := D + D^{{1}/{2}}[Q]_t[\Lambda]_t[Q]_t^\top D^{{1}/{2}}$$

We start by proving Lemma \ref{lemma:sketch-guarantee}, then justify the assumptions of the lemma.

\subsection{Proof of Lemma \ref{lemma:sketch-guarantee}}
We state a few technical results that will simplify the proof of the lemma.

\begin{fact}\label{fact:rayleigh-symmetric}
For any symmetric matrix $M \in \R^{d \times d}$ with eigendecomposition $USU^\top$, we have that for any $v \in \R^{d}$, 
$$v^\top M v = \sum_{i = 1}^d S_i \cdot \langle v, U_i \rangle^2.$$
\end{fact}

\begin{fact}\label{fact:psd-eigenvalues}
    Assuming the eigenvalues $\Lambda$ of $R_\circ$ are sorted in non-increasing order, we have that $I + R_\circ = Q (I + \Lambda) Q^\top$. In words, the eigenvectors of $I + R_\circ$ are the same as $R_\circ$, and the $i$-th eigenvalue is $\Lambda_i + 1$.
    As a corollary, since $I + R_\circ$ is PSD, we have that $\Lambda_i \geq -1$ for all $i \in [d]$.
\end{fact}
\begin{proof}
 This follows easily after noticing that $I=QQ^\top$ because $Q$ is a $d \times d$ matrix with orthonormal columns (and rows).
\end{proof}

\begin{fact}\label{fact:diagonal-error}
For a diagonal matrix $S \in \R^{d \times d}$ with bounded positive entries, i.e. $0 < l \leq S_{ii} \leq u$ for all $i \in [d]$, and arbitrary vector $v \in \R^d$, we have that $l\|v\|_2\leq \|Sv\|_2 \leq u \|v\|_2$.
\end{fact}

\begin{proof}[Proof of Lemma~\ref{lemma:sketch-guarantee}]
First, note that, by the definition of $\text{err}(\cdot, \cdot)$ from
Equation~(\ref{equation:approximation-error}), we can write:
\begin{align*}
    \err(D^{-\frac{1}{2}}\Sigma D^{-\frac{1}{2}}, D^{-\frac{1}{2}}\Sigma_t^\textsc{ms} D^{-\frac{1}{2}}) &= \sup_{v \in \R^d} \frac{|v^\top D^{-1/2} (\Sigma - \Sigma\Sigma_t^\textsc{ms}) D^{-1/2}v|}{\|v\|_2^2}.
\end{align*}
Since $D$ has strictly positive entries on the diagonal, we can do a change of variables, setting $u = D^{1/2}v$. Denoting $\max_{i \in [d]} D_{ii}$ by $\|D\|_\infty$, this gives us:
\begin{align*}
    \err(D^{-\frac{1}{2}} &\Sigma D^{-\frac{1}{2}}, D^{-\frac{1}{2}}\Sigma_t^\textsc{ms} D^{-\frac{1}{2}}) = \sup_{v \in \R^d} \frac{|v^\top (\Sigma - \Sigma_t^\textsc{ms}) v|}{\|D^{1/2}v\|_2^2}\\
    &\geq \frac{1}{\|D\|_\infty} \cdot \sup_{v \in \R^d} \frac{|v^\top (\Sigma - \Sigma_t^\textsc{ms}) v|}{\|v\|_2^2}
    = \frac{\err(\Sigma, \Sigma_t^\textsc{ms})}{\|D\|_\infty}
\end{align*}
where the inequality follows by Fact \ref{fact:diagonal-error}. 

Next, consider the following:
\begin{align}
    \err(D^{-\frac{1}{2}} &\Sigma D^{-\frac{1}{2}}, D^{-\frac{1}{2}}\Sigma_t^\textsc{ms} D^{-\frac{1}{2}}) = \sup_{\lVert v \rVert_2 = 1} \big\lvert v^\top D^{-\frac{1}{2}} (\Sigma - \Sigma_t^\textsc{ms}) D^{-\frac{1}{2}} v \big\rvert \nonumber \\
    &= \sup_{\lVert v \rVert_2 = 1} \big\lvert v^\top ( R_\circ - [Q]_t [\Lambda]_t [Q]_t^\top )v \big\rvert \\
    &= \sup_{\lVert v \rVert_2 = 1} \big\lvert \sum_{i = t + 1}^d \Lambda_i \cdot \langle Q_i, v \rangle^2 \big\rvert \\
    &= \max_{l \in [t+1, d]} \lvert \Lambda_l \rvert.
\end{align}
where the second equality follows by the definition of $\Sigma_t^\textsc{ms}$ and the third by Fact \ref{fact:rayleigh-symmetric}.

Putting our arguments thus far together, we have established that:
\begin{equation}
    \label{equation:lemma-2-proof:step-1}
    \frac{\err(\Sigma, \Sigma_t^\textsc{ms})}{\|D\|_\infty} \leq \max_{l \in [t+1, d]} \lvert \Lambda_l \rvert.
\end{equation}

Using this result, and noting that,
$1 + \Lambda_{t+1} \geq 1$ by assumption, so that $\Lambda_{t+1} \geq 0$;
and $\Lambda_l \geq -1$ for all $l \in [d]$ by Fact~\ref{fact:psd-eigenvalues},
we can derive the following:
\begin{align}
\frac{\err(\Sigma, \Sigma_t^\textsc{ms})}{\|D\|_\infty} &\leq \max_{l \in [t+1, d]} |\Lambda_l| \leq 1 + \Lambda_{t+1} \leq \max_{l \in [t + 1, d]} |1 + \Lambda_{l}| \\
&= \err(\underbrace{I + R_\circ}_{D^{-1/2}\Sigma D^{-1/2}}, [Q]_t [I + \Lambda]_t[Q]_t^\top). \label{eqn:symm-lemma-bound}
\end{align}

All that is left to do is to bound the right-hand side of Equation~(\ref{eqn:symm-lemma-bound}). We do so as follows:
\begin{align}
    \err(D^{-1/2}&\Sigma D^{-1/2}, [Q]_t [I + \Lambda]_t[Q]_t^\top) \\
    &\leq \err(D^{-1/2}\Sigma D^{-1/2}, D^{-1/2}\Sigma_t^\textsc{lr} D^{-1/2}) \\
    &= \sup_{\lVert v \rVert_2=1} \big\lvert v^\top D^{-1/2}(\Sigma - \Sigma_t^\textsc{lr})D^{-1/2} v \big\rvert \\
    &= \sup_{\lVert v \rVert_2=1} \big\lvert \sum_{i, j} v_i v_j (\Sigma - \Sigma_t^\textsc{lr})_{ij} / \sqrt{D_{ii} D_{jj}} \big\rvert \\
    &\leq \sup_{\lVert v \rVert_2=1} \big\lvert \sum_{i, j} v_i v_j (\Sigma - \Sigma_t^\textsc{lr})_{ij} \frac{1}{(1-\epsilon) \lVert D \rVert_\infty} \big\rvert \\
    &\leq \frac{1}{(1-\epsilon) \lVert D \rVert_\infty} \sup_{\lVert v \rVert_2=1} \big\lvert v^\top (\Sigma - \Sigma_t^\textsc{lr}) v \big\rvert \\
    &= \frac{1}{(1-\epsilon) \lVert D \rVert_\infty} \text{err}(\Sigma, \Sigma_t^\textsc{lr}),
\end{align}
where the first inequality is because the left-hand side is the optimal error by the
Eckhart-Young-Mirsky Theorem, and the second inequality is due to the fact that
$D_{ii} \geq (1 - \epsilon)\|D\|_\infty$.

Putting everything together, we have shown that:
\begin{equation*}
    \frac{\err(\Sigma, \Sigma_t^\textsc{ms})}{\|D\|_\infty} \leq \frac{1}{(1-\epsilon) \lVert D \rVert_\infty} \text{err}(\Sigma, \Sigma_t^\textsc{lr}),
\end{equation*}
as desired.
\end{proof}

\begin{figure*}[t]
\begin{center}
\centerline{
    \subfloat[\deep]{
        \includegraphics[width=0.30\linewidth]{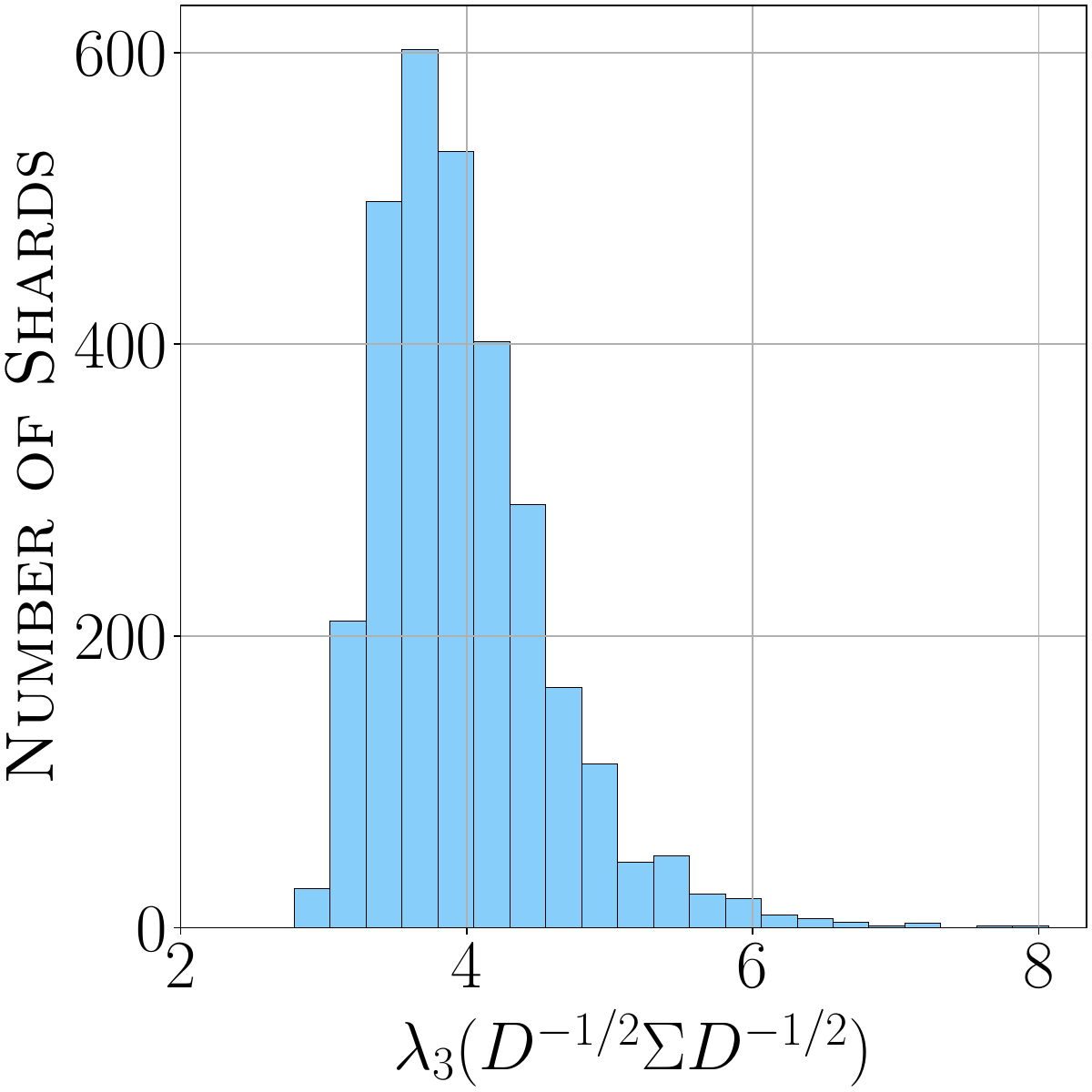}
    }
    \subfloat[\glove]{
        \includegraphics[width=0.30\linewidth]{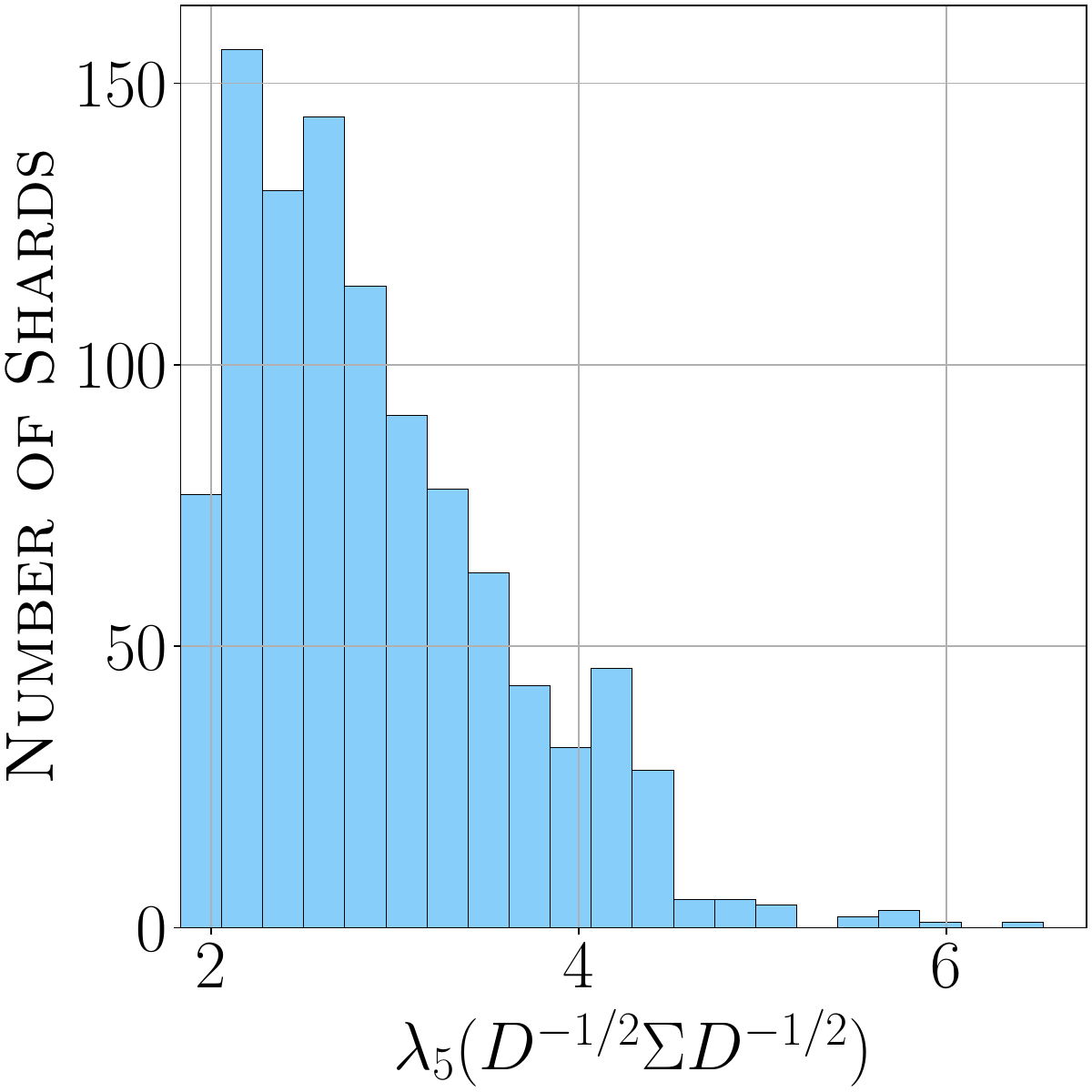}
    }
    \subfloat[\msmarco]{
        \includegraphics[width=0.30\linewidth]{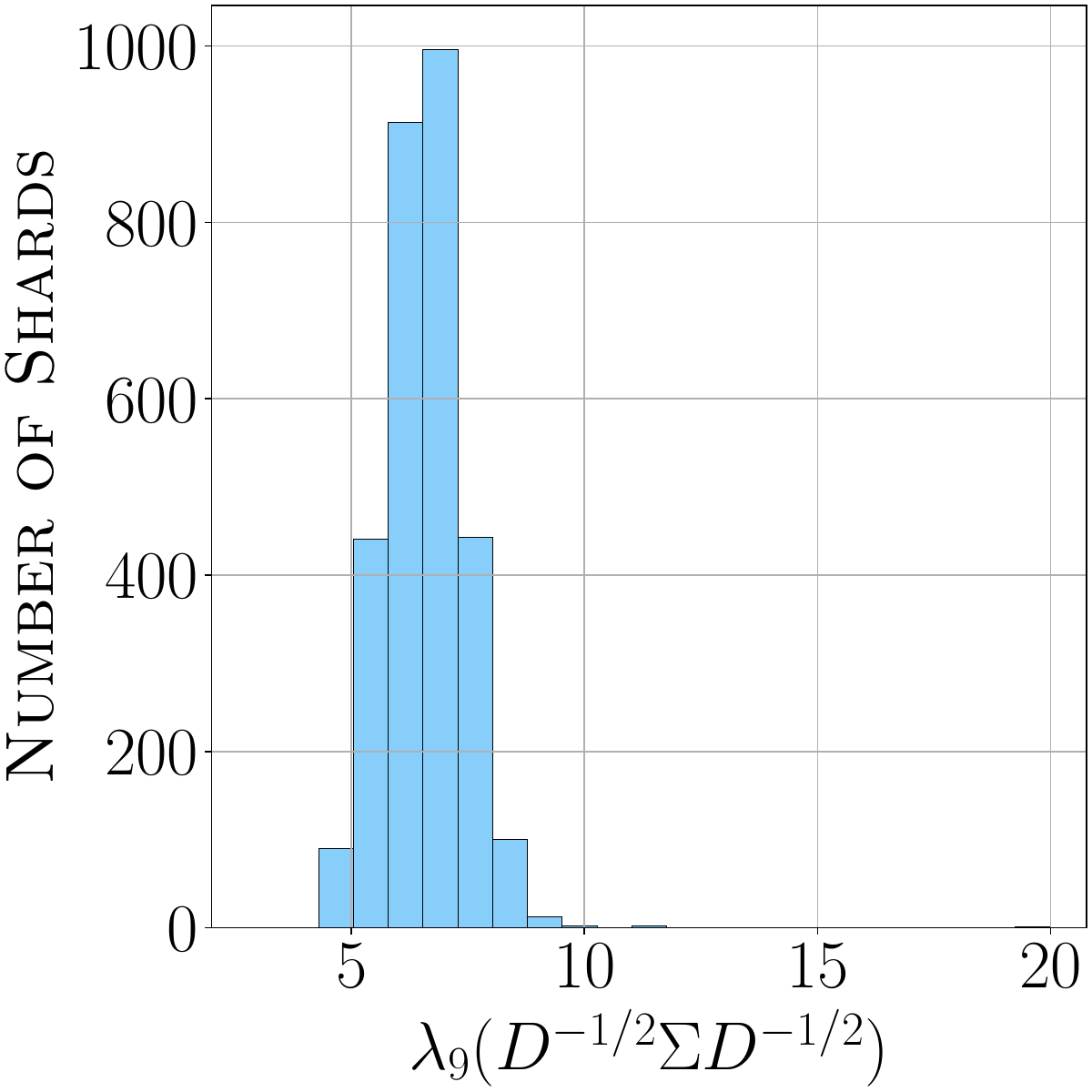}
    }
}
\centerline{
    \subfloat[\music]{
        \includegraphics[width=0.30\linewidth]{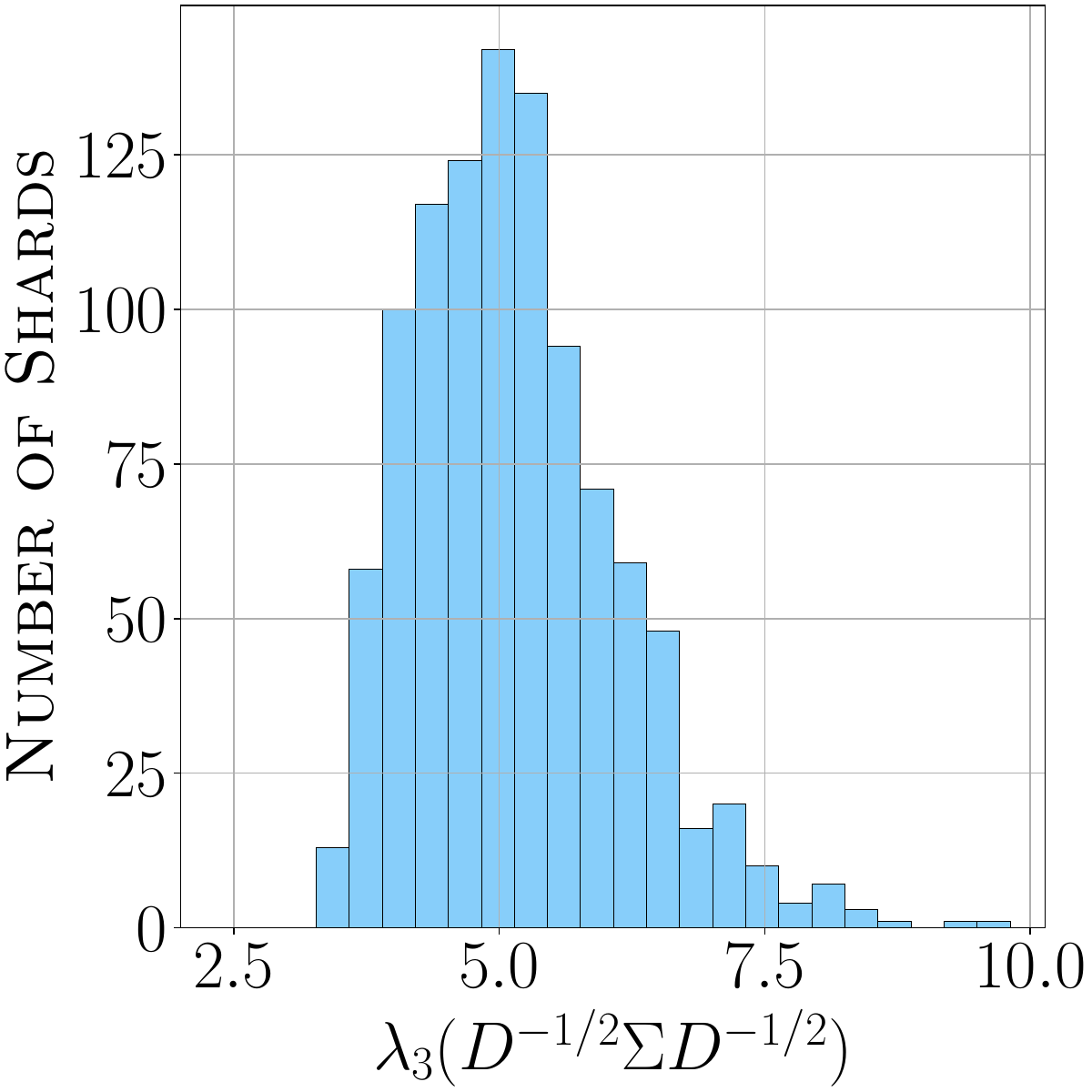}
    }
    \subfloat[\nq]{
        \includegraphics[width=0.30\linewidth]{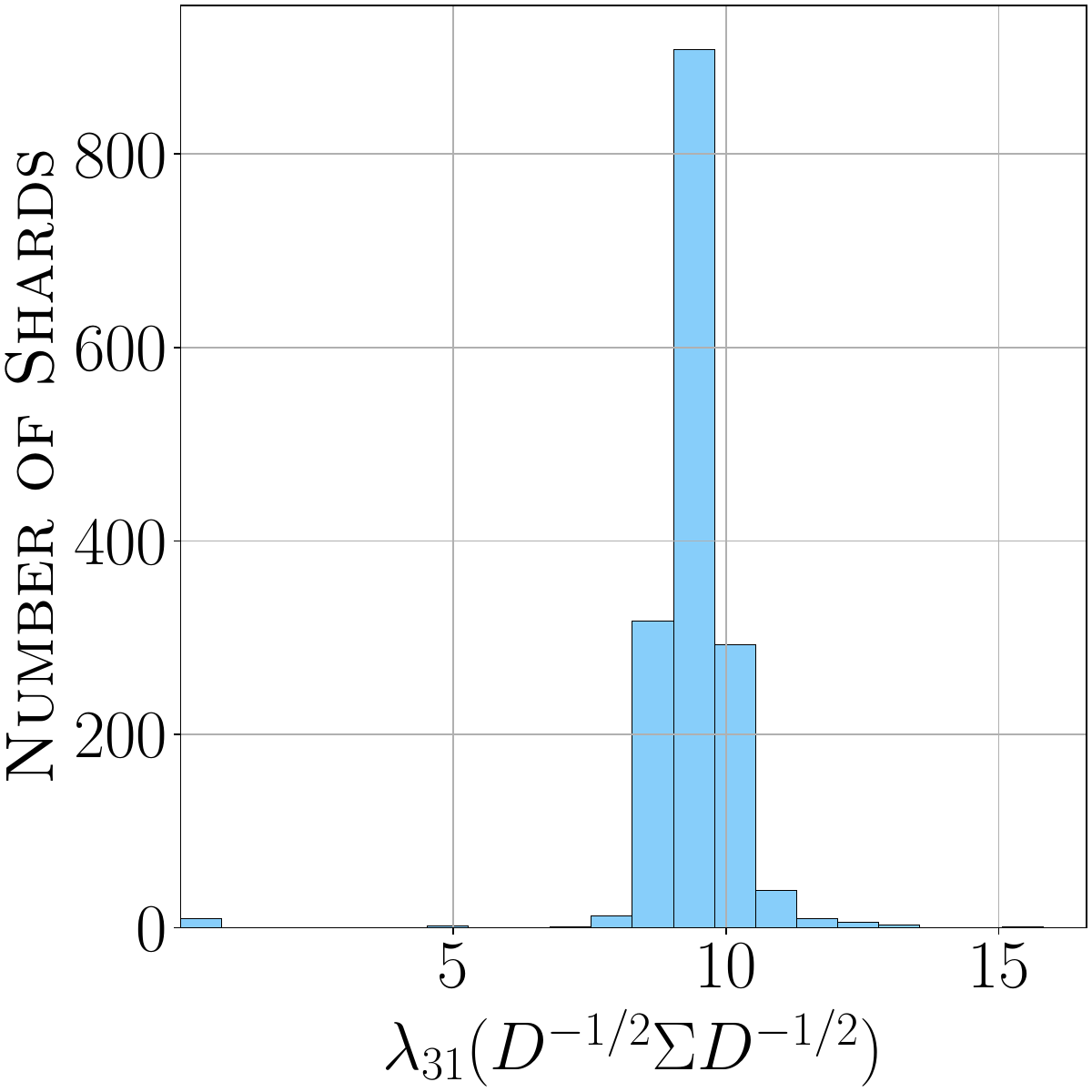}
    }
    \subfloat[\textimage]{
        \includegraphics[width=0.30\linewidth]{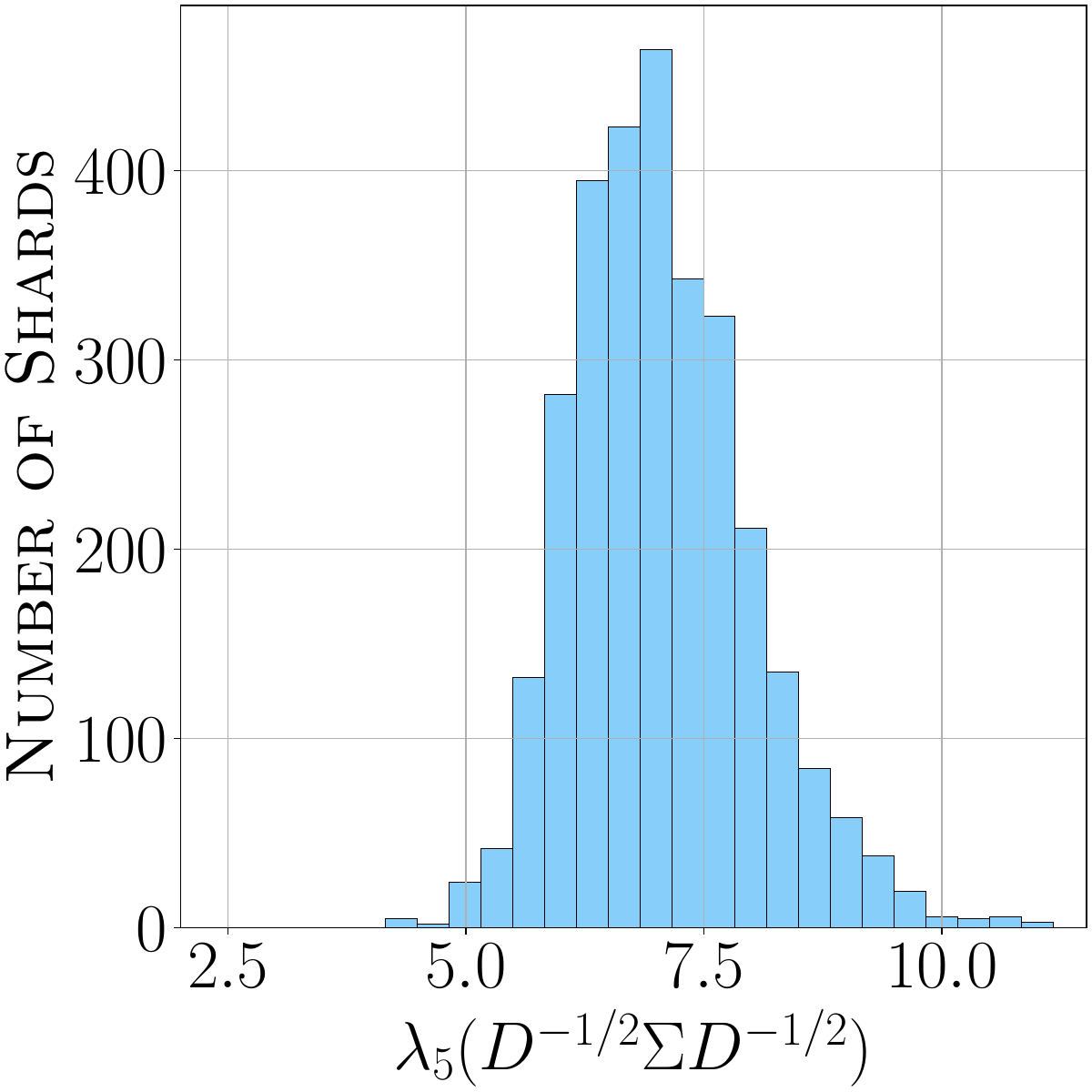}
    }
}
\caption{Histogram of the $(t + 1)$-th eigenvalue. For each dataset, we pick the partitioning and $t$ from Table~\ref{table:router-configuration}. Plots show that almost all shards for all datasets have $(t+1)$-th eigenvalue bounded away from $1$, except for a few shards for \nq.}
\label{figure:t_plus_1_eigenvalue:full}
\end{center}
\end{figure*}

\subsection{Assumptions of Lemma \ref{lemma:sketch-guarantee}}
Recall that we make two assumptions about the covariance matrix $\Sigma$ and the eigendecomposition
of its symmetrization, $D^{-1/2} \Sigma D^{-1/2}$: 
\begin{enumerate}
    \item The $(t+1)$-th eigenvalue of $D^{-1/2} \Sigma D^{-1/2}$ is greater than or equal to $1$. In particular, letting $D^{-1/2} \Sigma D^{-1/2} = Q(I + \Lambda) Q^\top$ be the orthogonal eigendecomposition of the symmetrization, we assume $1 + \Lambda_{t + 1} \geq 1$. 
    \item The diagonal of $\Sigma$ has the property that $\min_{i \in [d]} \Sigma_{ii} \geq (1 - \epsilon) \max_{i \in [d]} \Sigma_{ii}$ for some $\epsilon \in (0, 1]$. 
\end{enumerate}
\paragraph{First assumption.} Notice that $D^{-1/2} \Sigma D^{-1/2} = I + D^{-1/2}RD^{-1/2}$ is symmetric PSD, so we have that $\text{tr}(D^{-1/2} \Sigma D^{-1/2}) = \sum_{i = 1}^d 1 + \Lambda_i = d$. Hence by definition there must exist some $t$ for which $1 + \Lambda_{t+1} \geq 1$. While in the worse case we cannot hope for the existence of an eigenvalue larger than this, in practice, including for the datasets we consider in this work, it can be shown that in fact the eigenvalues of $D^{-1/2} \Sigma D^{-1/2}$ are larger than $1$ across shards and datasets--see Figure \ref{figure:t_plus_1_eigenvalue:full}. 

\paragraph{Second assumption.} While in the worst case, the diagonal of $\Sigma$ can have arbitrarily large entries compared to its smallest entries, in practice this is rarely the case. While we do not explore how to remove this assumption, there are several mechanisms to do so in practice such as applying random rotations or pseudo-random rotations \cite{ailon2009fast,woodruff2014sketching,ailon2013almost} to the data points in each shard before using Algorithm \ref{alg:full}. It is well known (e.g. Lemma 1 in \cite{ailon2009fast}) that after applying such transforms, the coordinates of the vectors are ``roughly equal,'' thereby ensuring that the diagonal of the covariance has entries of comparable magnitude. We leave the exploration of removing this assumption to future work.
\section{Datasets}
\label{appendix:datasets}

The following is a complete description of the datasets used in this work:
\begin{itemize}[leftmargin=*]
    \item \textbf{\textimage}: A cross-modal dataset, where data and query points may have different distributions in a shared space~\citep{pmlr-v176-simhadri22a}. We use a subset consisting of $10$ million $200$-dimensional data points along with a subset of $10{,}000$ test queries. The dataset is available under the terms of the Creative Commons Attribution 4.0 International license.
    \item \textbf{\music}: $1$ million $100$-dimensional points~\citep{music-dataset} with $1{,}000$ queries. To the best of our knowledge, the dataset comes with no information on the license under which it is made available (per https://github.com/stanis-morozov/ip-nsw?tab=readme-ov-file).
    \item \textbf{\deep}: Subset of $10$ million $96$-dimensional points from the billion deep image features dataset~\citep{deep-dataset} with $10{,}000$ queries. The dataset is available under the terms of Apache license 2.0.
    \item \textbf{\glove}: $1{.}2$ million, $200$-dimensional word embeddings trained on tweets~\citep{pennington-etal-2014-glove} with $10{,}000$ test queries. The dataset is available under the terms of the Public Domain Dedication and license v1.0.
    \item \textbf{\msmarco}: \textsc{Ms Marco} Passage Retrieval v1~\citep{nguyen2016msmarco} is a question-answering dataset consisting of $8{.}8$ million short passages in English. We use the ``dev'' set of queries for retrieval, made up of $6{,}980$ questions. We embed individual passages and queries using the \textsc{all-MiniLM-L6-v2} model\footnote{Checkpoint at \url{https://huggingface.co/sentence-transformers/all-MiniLM-L6-v2}.} to form a $384$-dimensional vector collection. The dataset is available under the terms of the Creative Commons Attribution 4.0 International license. The model used to embed the dataset is available under the terms of Apache license 2.0.
    \item \textbf{\nq}: $2{.}7$ million, $1{,}536$-dimensional embeddings of the Natural Questions dataset~\citep{kwiatkowski-etal-2019-natural} with the \textsc{Ada-002} model.\footnote{\url{https://openai.com/index/new-and-improved-embedding-model/}} The dataset is available under the terms of Apache license 2.0.
\end{itemize}

We note that the last four datasets are intended for cosine similarity search. As such we normalize these collections prior to indexing, reducing the task to MIPS of Equation~(\ref{equation:mips}).

\section{Effect of hyper-parameters on \optimist}
\label{appendix:optimist-hyperparameters}

\begin{figure*}
\begin{center}
\centerline{
    \subfloat[\nq]{
        \includegraphics[width=0.28\linewidth]{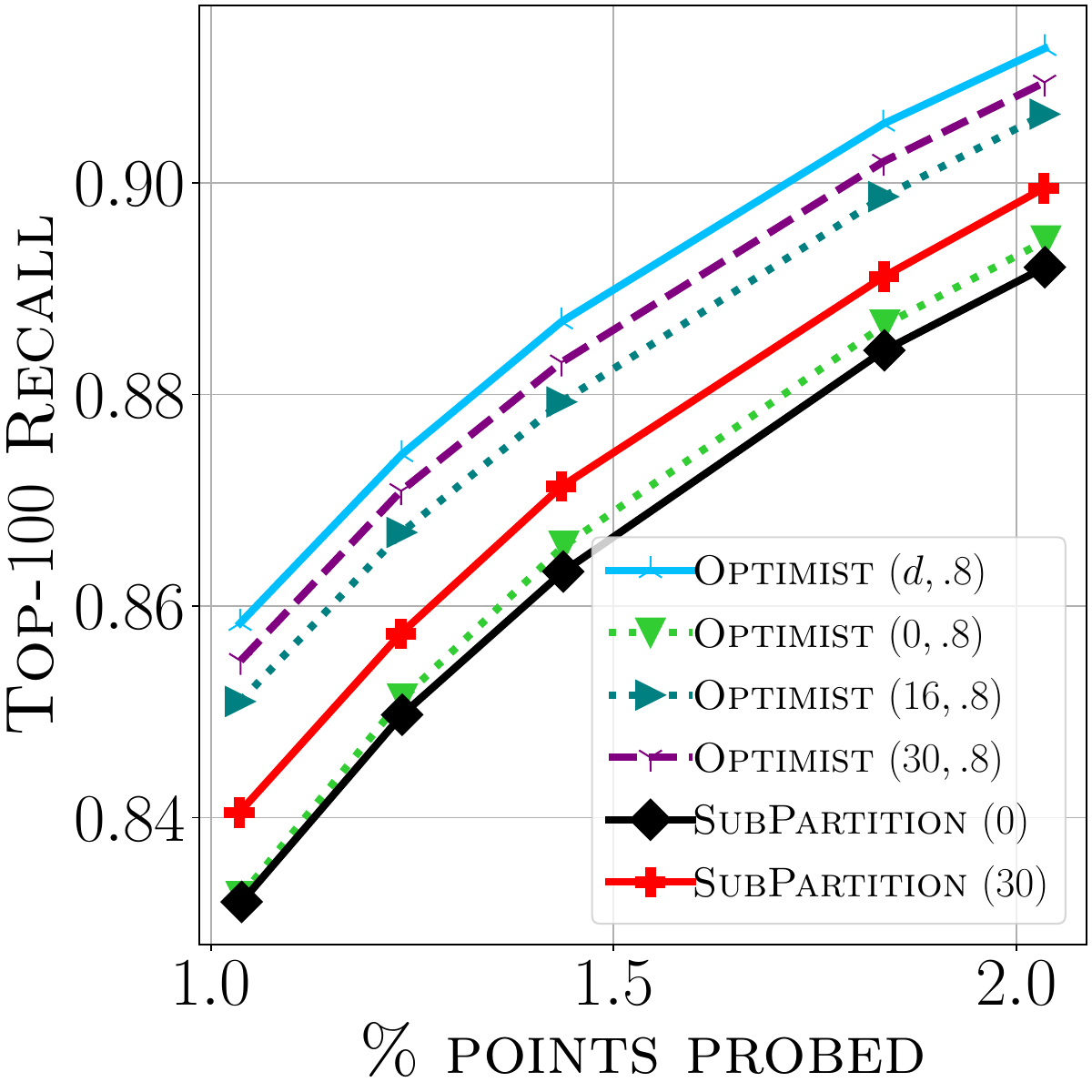}
    }
    \subfloat[\glove]{
        \includegraphics[width=0.28\linewidth]{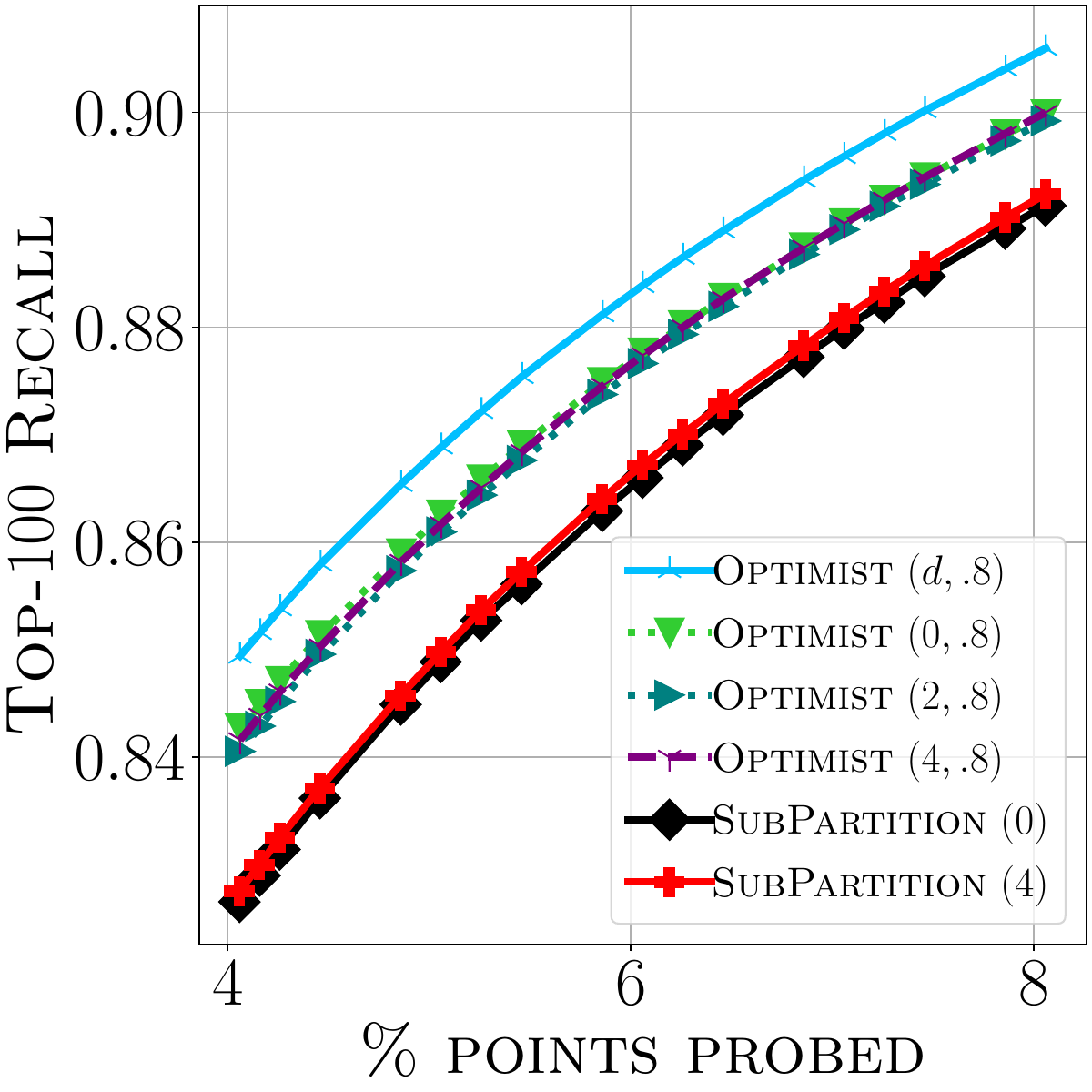}
    }
    \subfloat[\msmarco]{
        \includegraphics[width=0.28\linewidth]{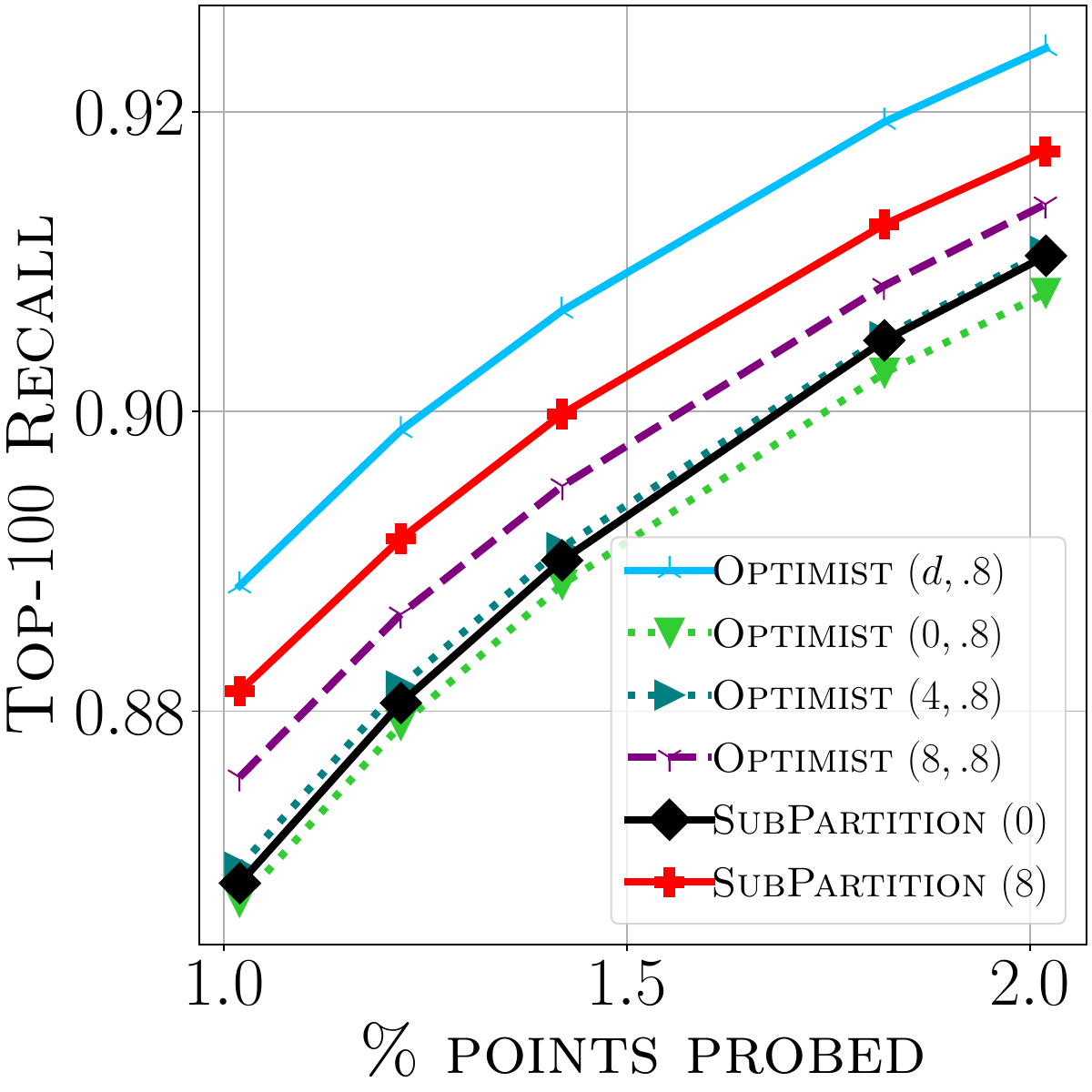}
    }
}
\centerline{
    \subfloat[\deep]{
        \includegraphics[width=0.28\linewidth]{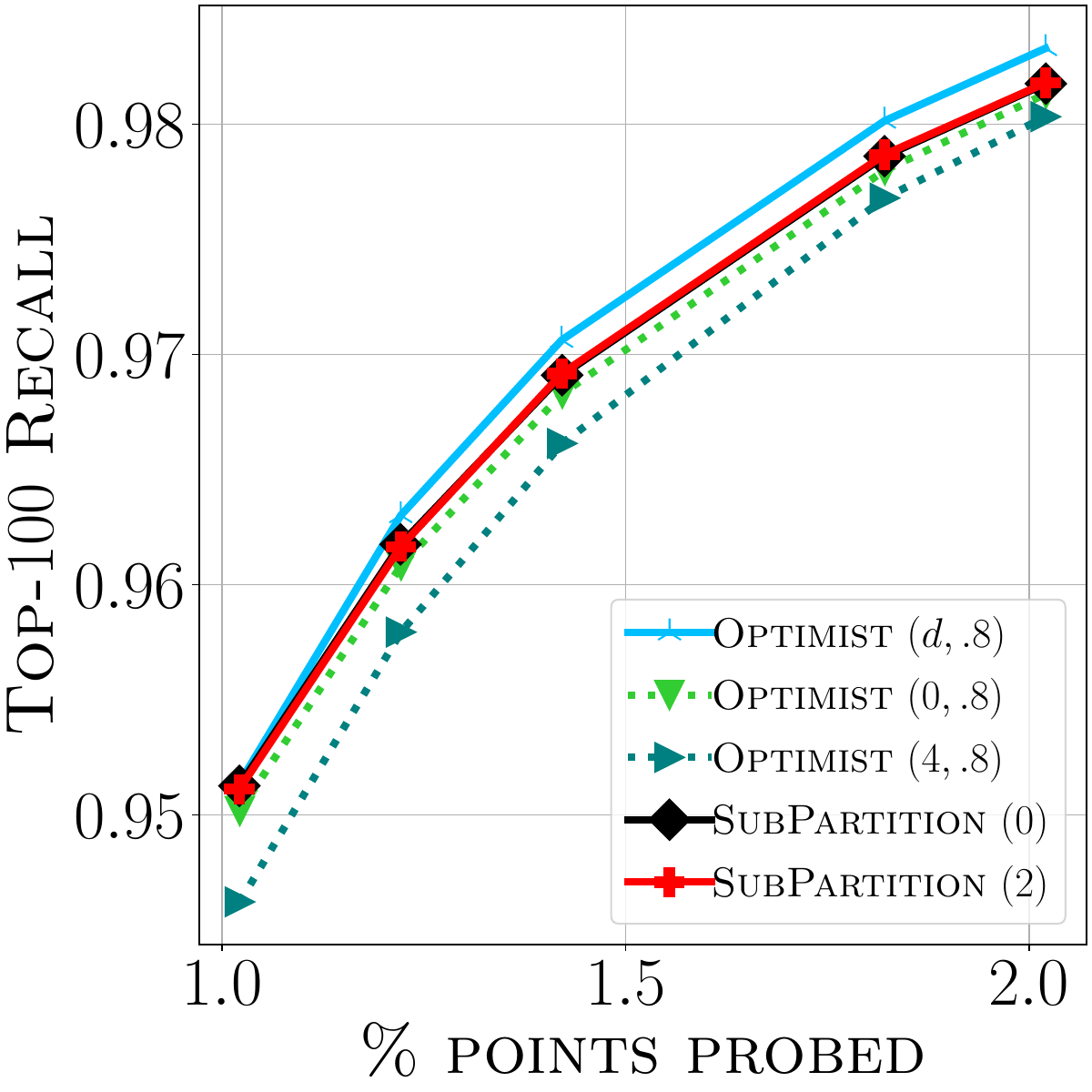}
    }
    \subfloat[\music]{
        \includegraphics[width=0.28\linewidth]{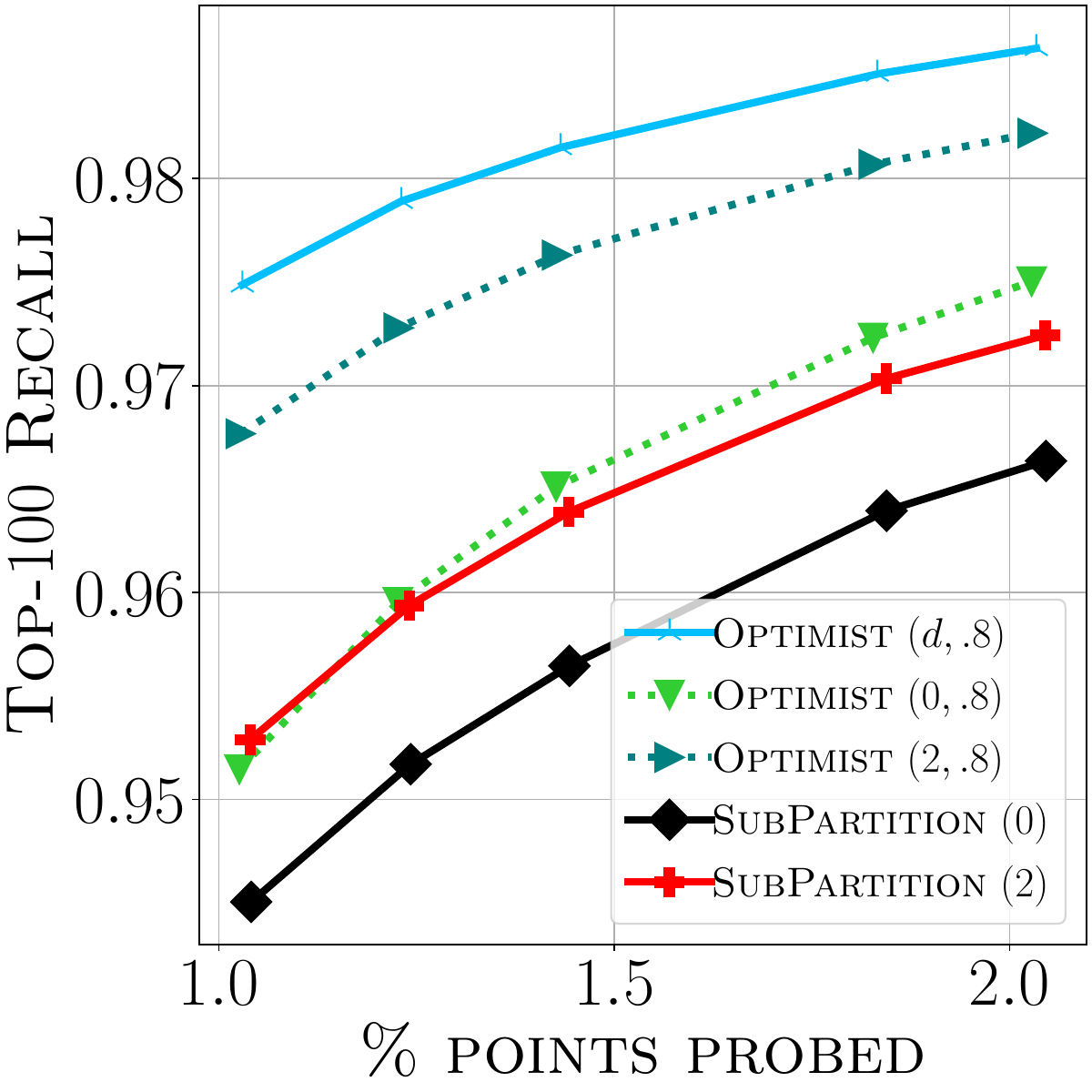}
    }
    \subfloat[\textimage]{
        \includegraphics[width=0.28\linewidth]{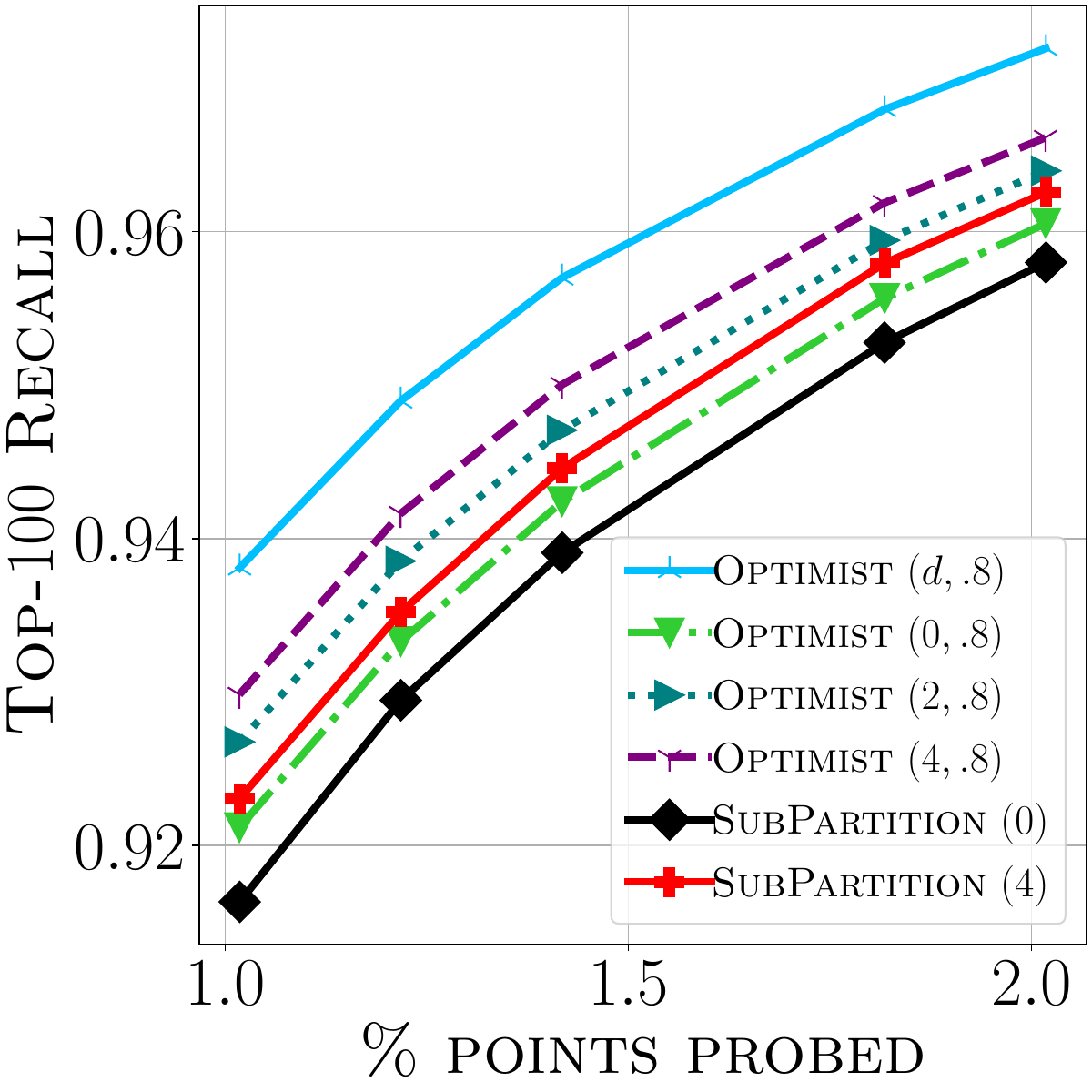}
    }
}
\caption{Top-$100$ recall vs. volume of probed data as we change the rank parameter ($t$). $\optimist(0, \cdot)$ is a sketch that is the diagonals only. Partitioning is with Spherical KMeans.}
\label{figure:rank_sweep:full}
\end{center}
\end{figure*}

Recall that \optimist takes two parameters:
$t$, the rank of the covariance sketch, and $\delta$, the degree of optimism.
We examine the effect of these parameters on the performance of \optimist.

Figure~\ref{figure:rank_sweep:full} visualizes the role played by $t$.
It comes as no surprise that larger values of $t$ lead to a better approximation of the covariance matrix.
What we found interesting, however, is the remarkable effectiveness of a sketch that simply retains the diagonal
of the covariance, denoted by $\optimist(0, \cdot)$, in the settings of $k$ we experimented with
(i.e., $k \in \{ 1, 10, 100\}$).

In the same figure, we have also included two configurations of \subpartition:
one with $2$ sub-partitions, $\subpartition(0)$,
and another with $t+2$ sub-partitions, $\subpartition(t)$, for the largest $t$.
These help put the performance of \optimist with various ranks in perspective.
In particular, we give the \subpartition baseline the same amount of information
and contrast its recall with \optimist.

\begin{figure*}
\begin{center}
\centerline{
    \subfloat[\nq]{
        \includegraphics[width=0.28\linewidth]{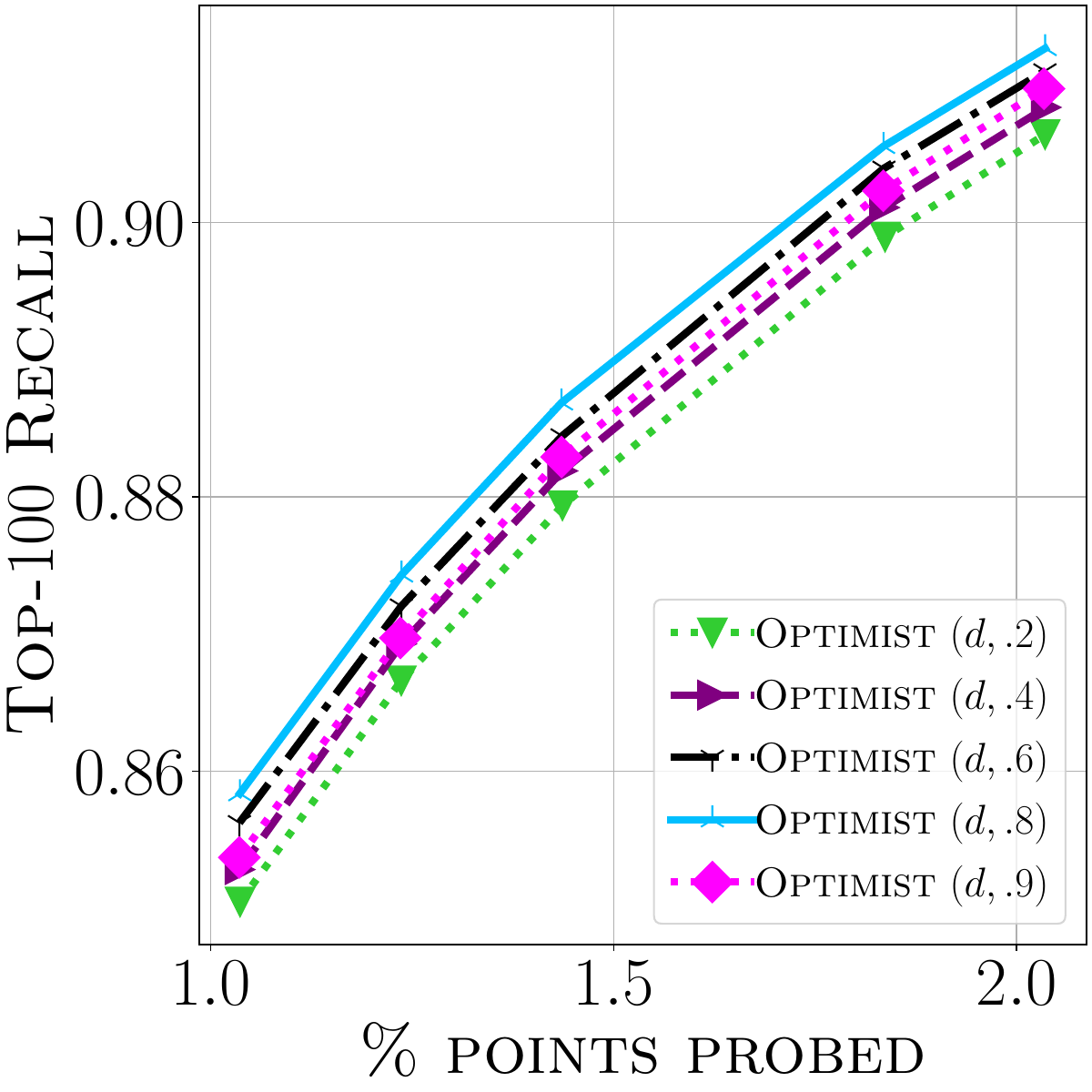}
    }
    \subfloat[\glove]{
        \includegraphics[width=0.28\linewidth]{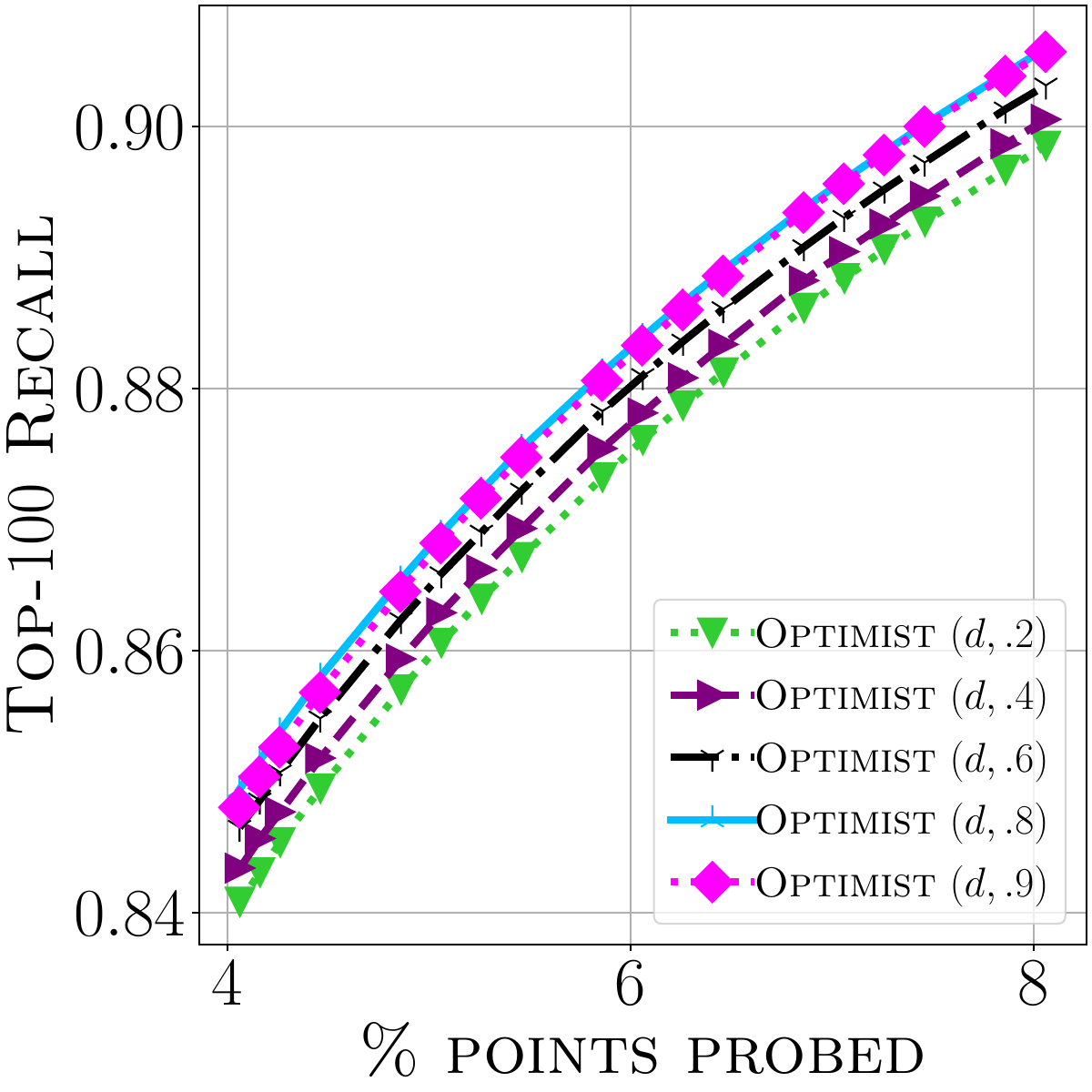}
    }
    \subfloat[\msmarco]{
        \includegraphics[width=0.28\linewidth]{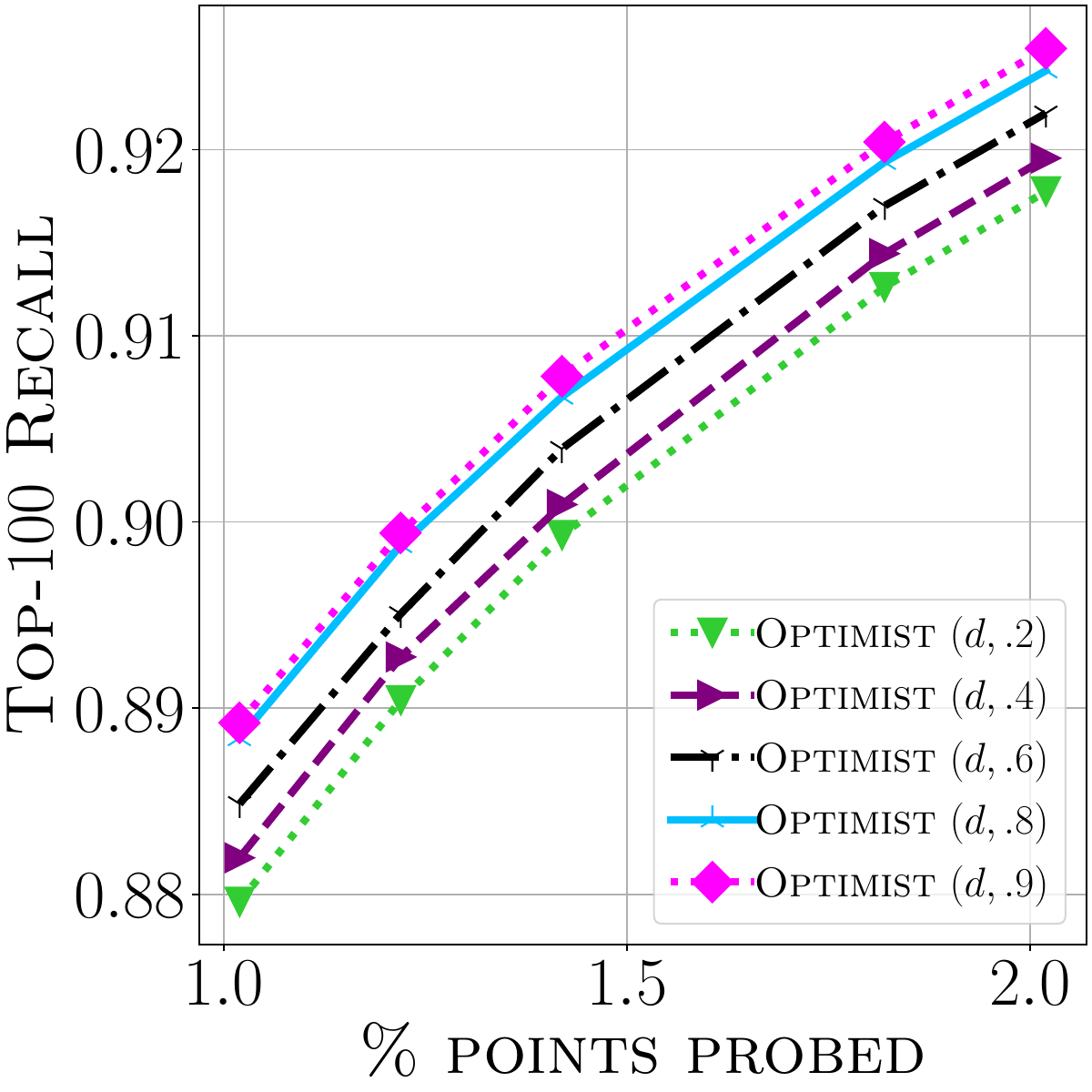}
    }
}
\centerline{
    \subfloat[\deep]{
        \includegraphics[width=0.28\linewidth]{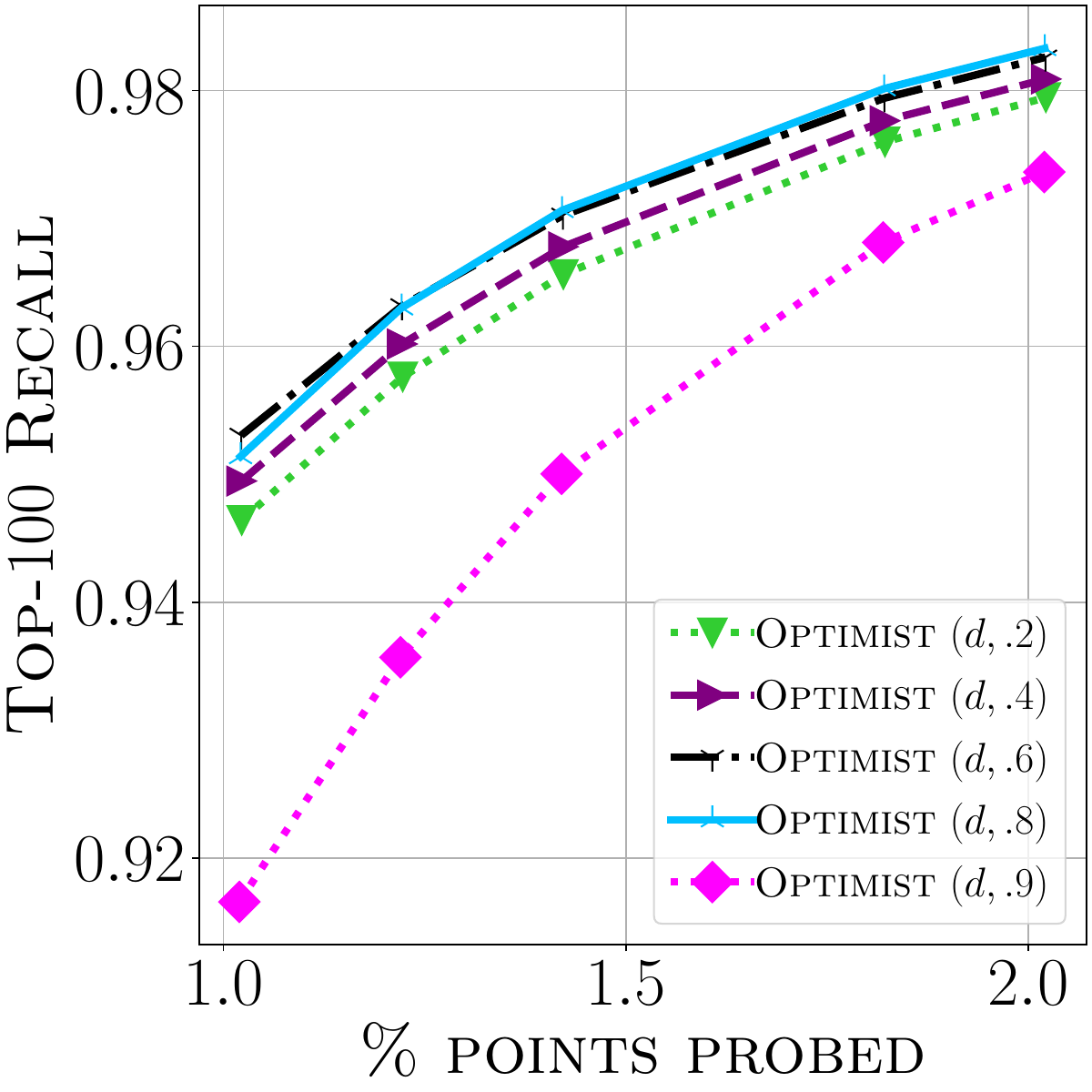}
    }
    \subfloat[\music]{
        \includegraphics[width=0.28\linewidth]{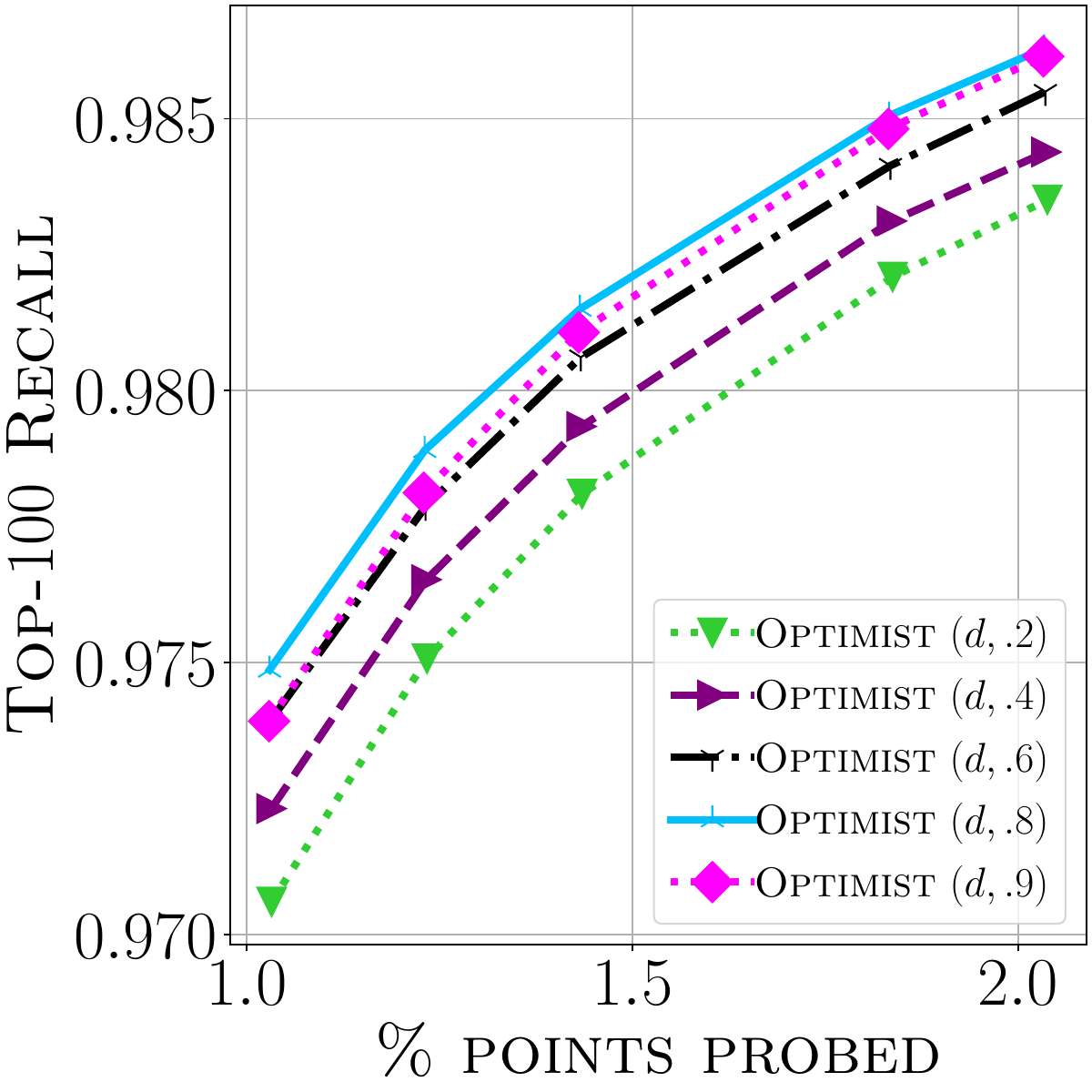}
    }
    \subfloat[\textimage]{
        \includegraphics[width=0.28\linewidth]{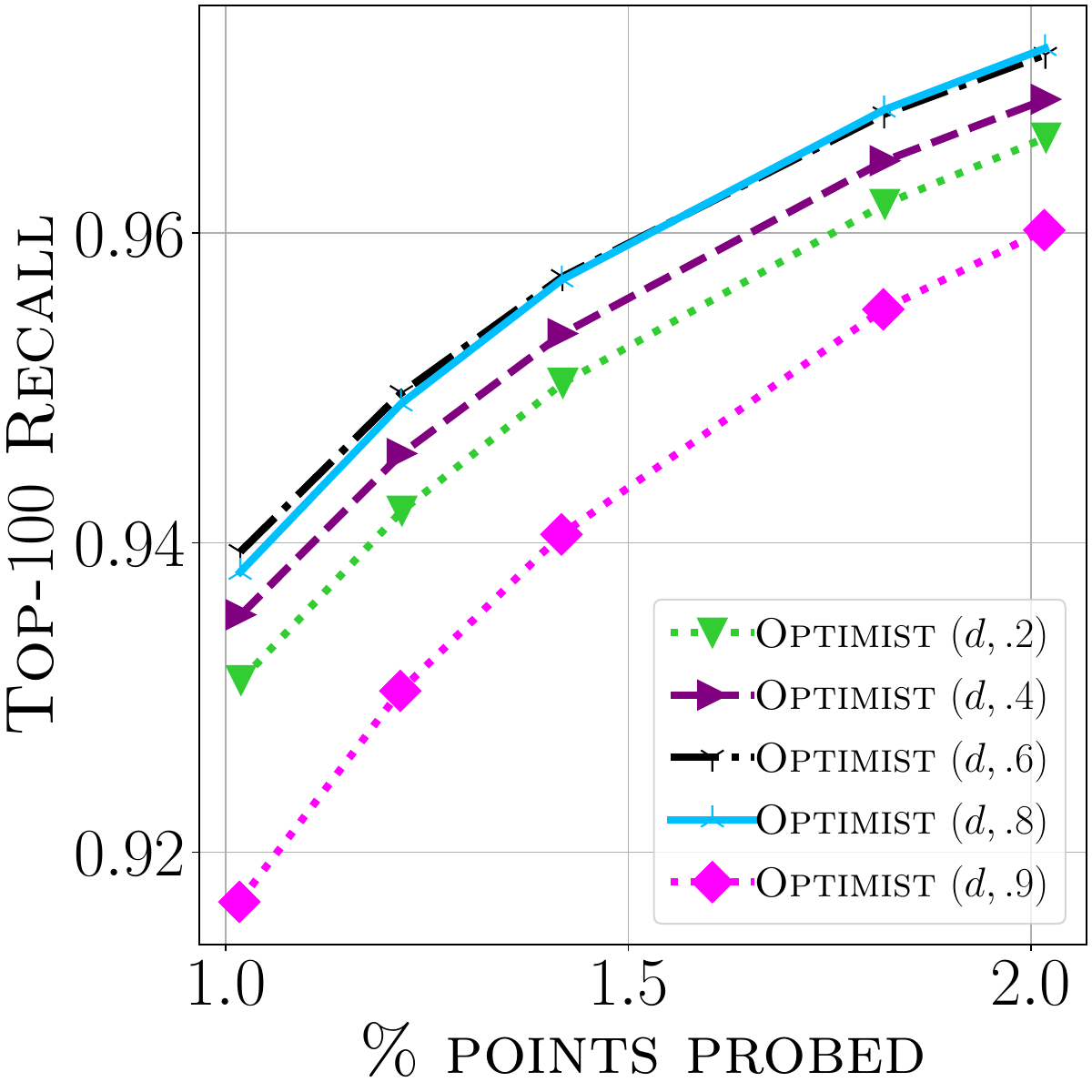}
    }
}
\caption{Top-$100$ recall vs. volume of probed data, comparing a range of values of $\delta$. As $\delta \rightarrow 1$, \optimist becomes more optimistic. Partitioning is with Spherical KMeans.}
\label{figure:delta_sweep:full}
\end{center}
\end{figure*}

We turn to Figure~\ref{figure:delta_sweep:full} to understand the impact of $\delta$. It is clear that encouraging \optimist to be too optimistic can lead to sub-optimal performance. That is because of our reliance on the Chebyshev's inequality, which can prove too loose, leading to an overestimation of the maximum value. Interestingly, $\delta \in (0.6, 0.8)$ appears to yield better recall across datasets.

\newpage
\section{Experiments with spherical KMeans}
\label{appendix:experiments:spherical-kmeans}

\begin{figure}[h]
\begin{center}
\centerline{
    \subfloat[\nq]{
        \includegraphics[width=0.3\linewidth]{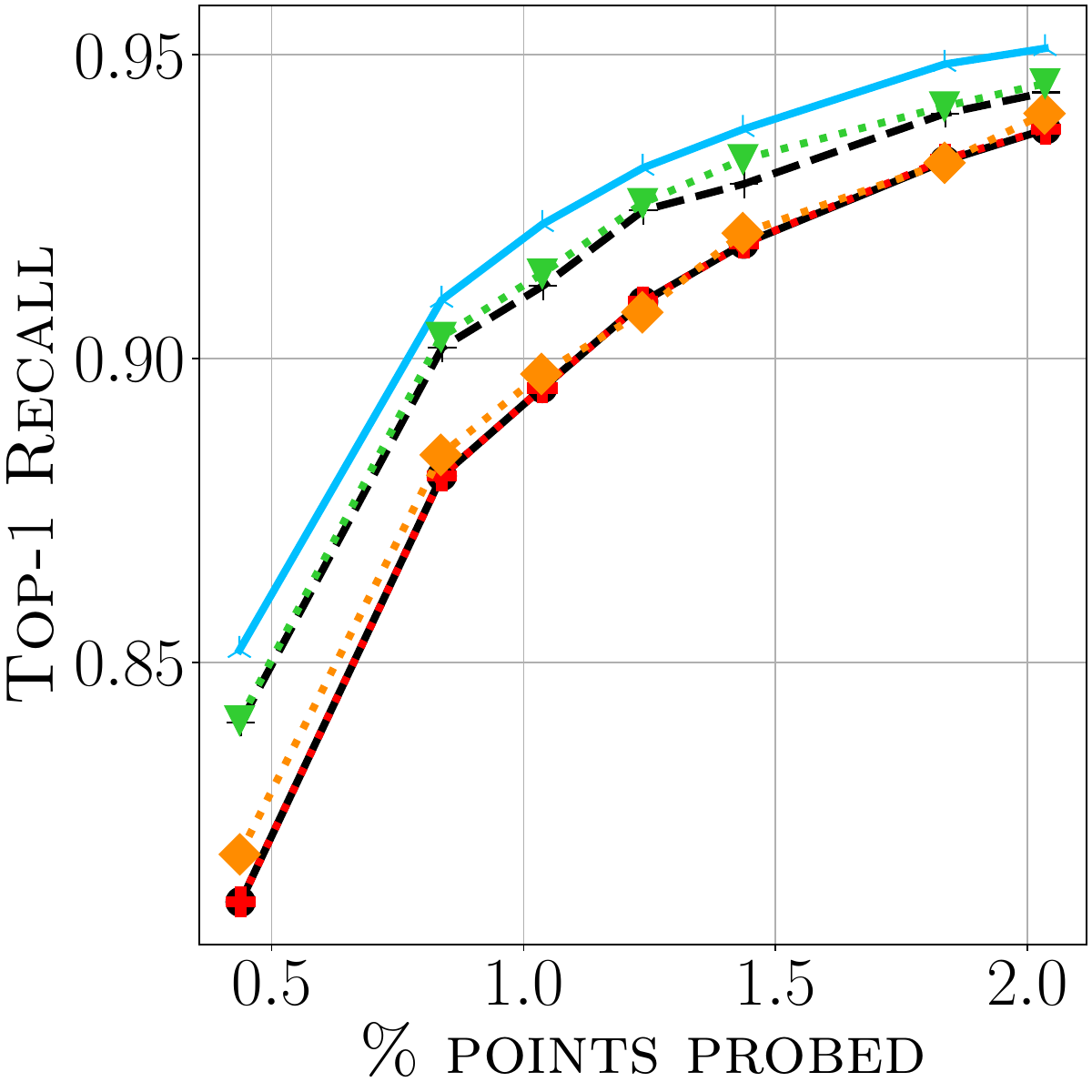}
        \includegraphics[width=0.3\linewidth]{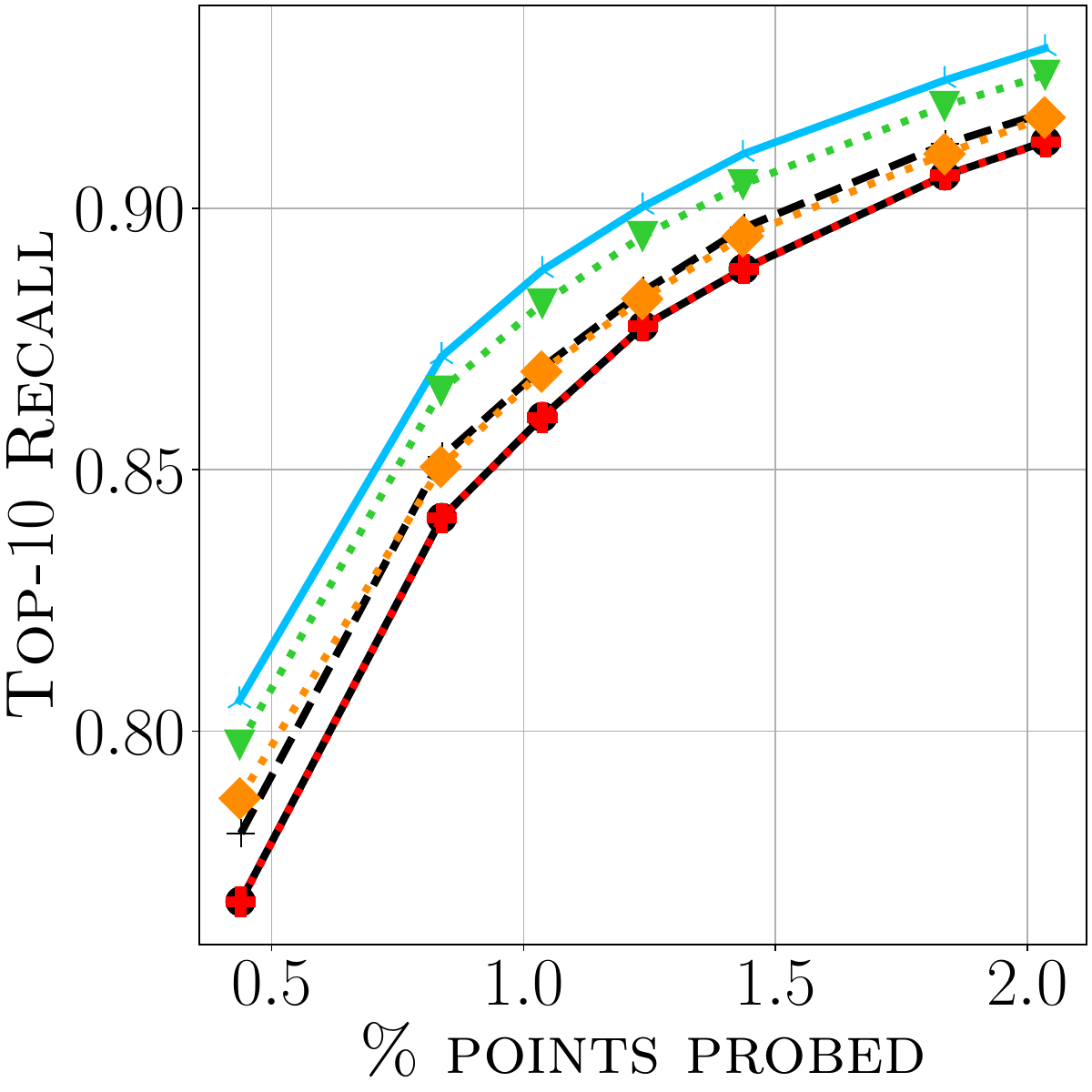}
        \includegraphics[width=0.3\linewidth]{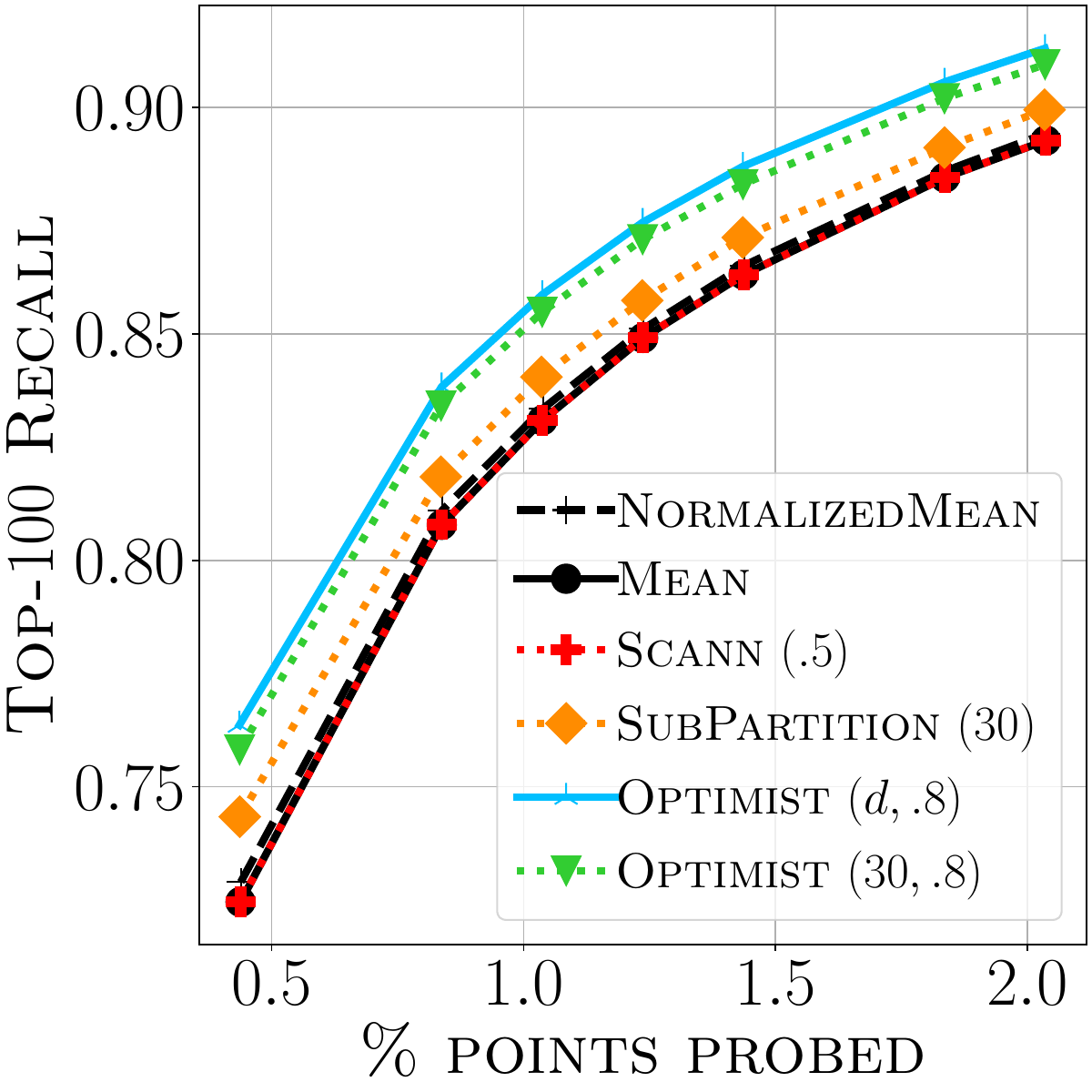}
    }
}
\centerline{
    \subfloat[\music]{
        \includegraphics[width=0.3\linewidth]{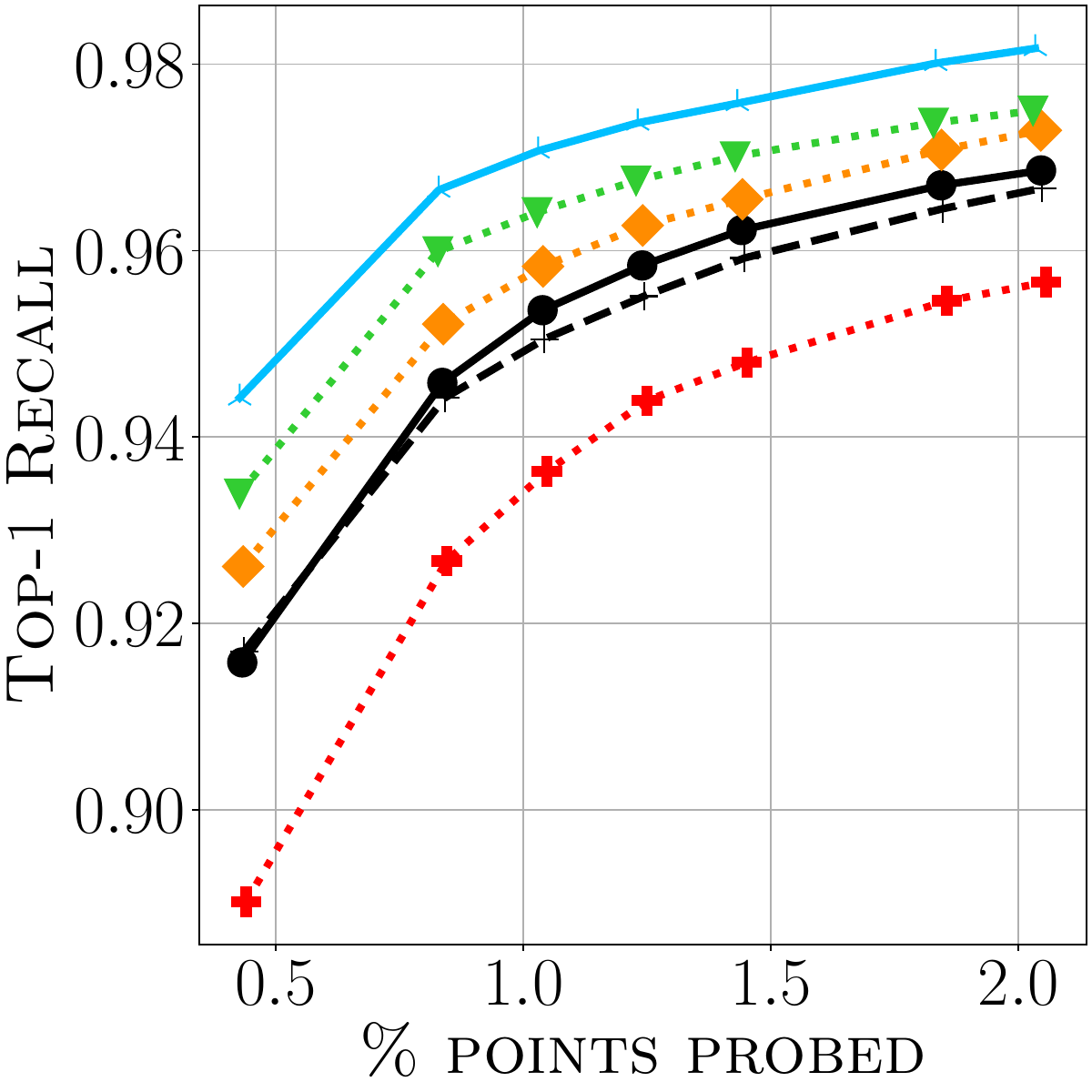}
        \includegraphics[width=0.3\linewidth]{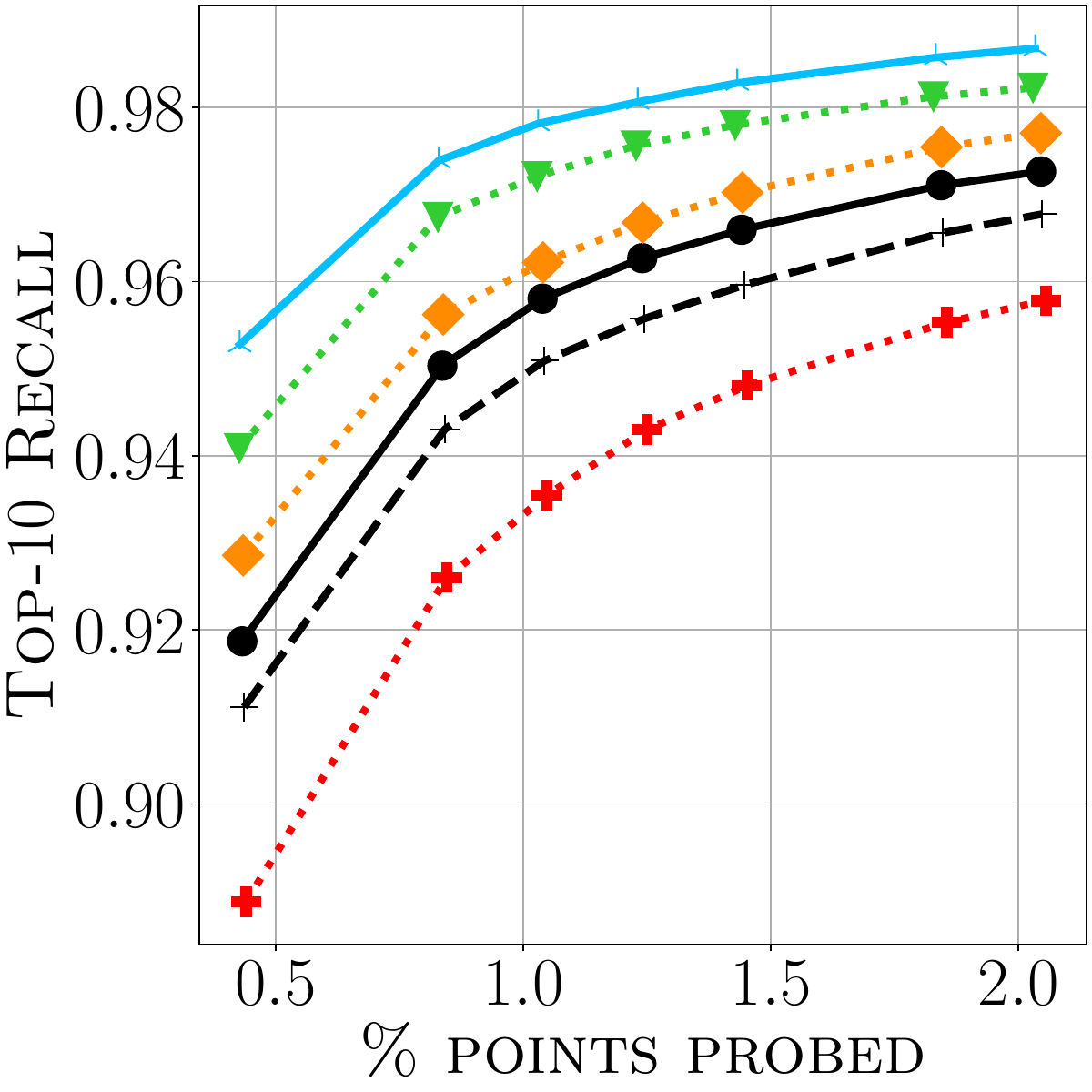}
        \includegraphics[width=0.3\linewidth]{figures/spherical_kmeans/music_100_mips_top100.pdf}
    }
}
\centerline{
    \subfloat[\glove]{
        \includegraphics[width=0.3\linewidth]{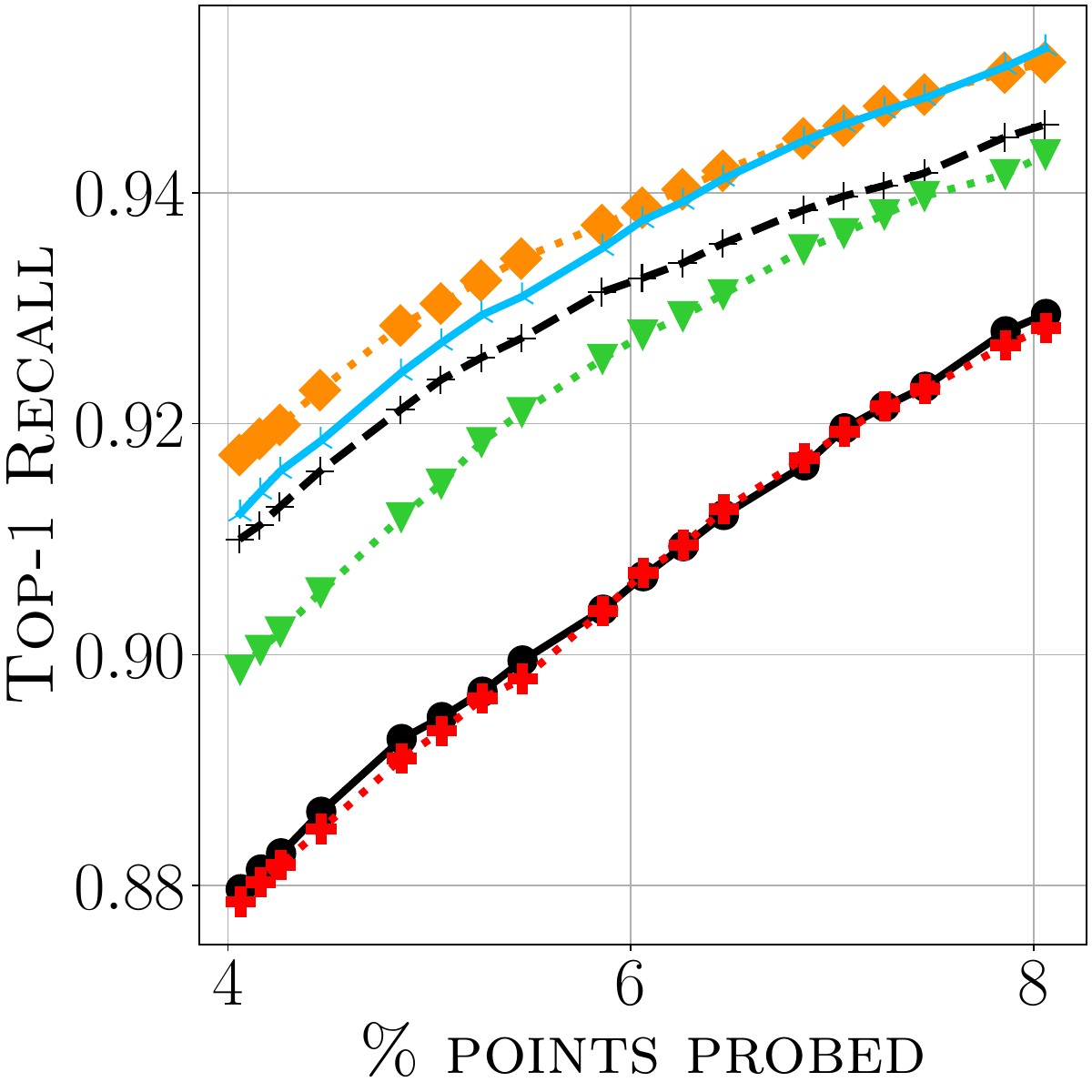}
        \includegraphics[width=0.3\linewidth]{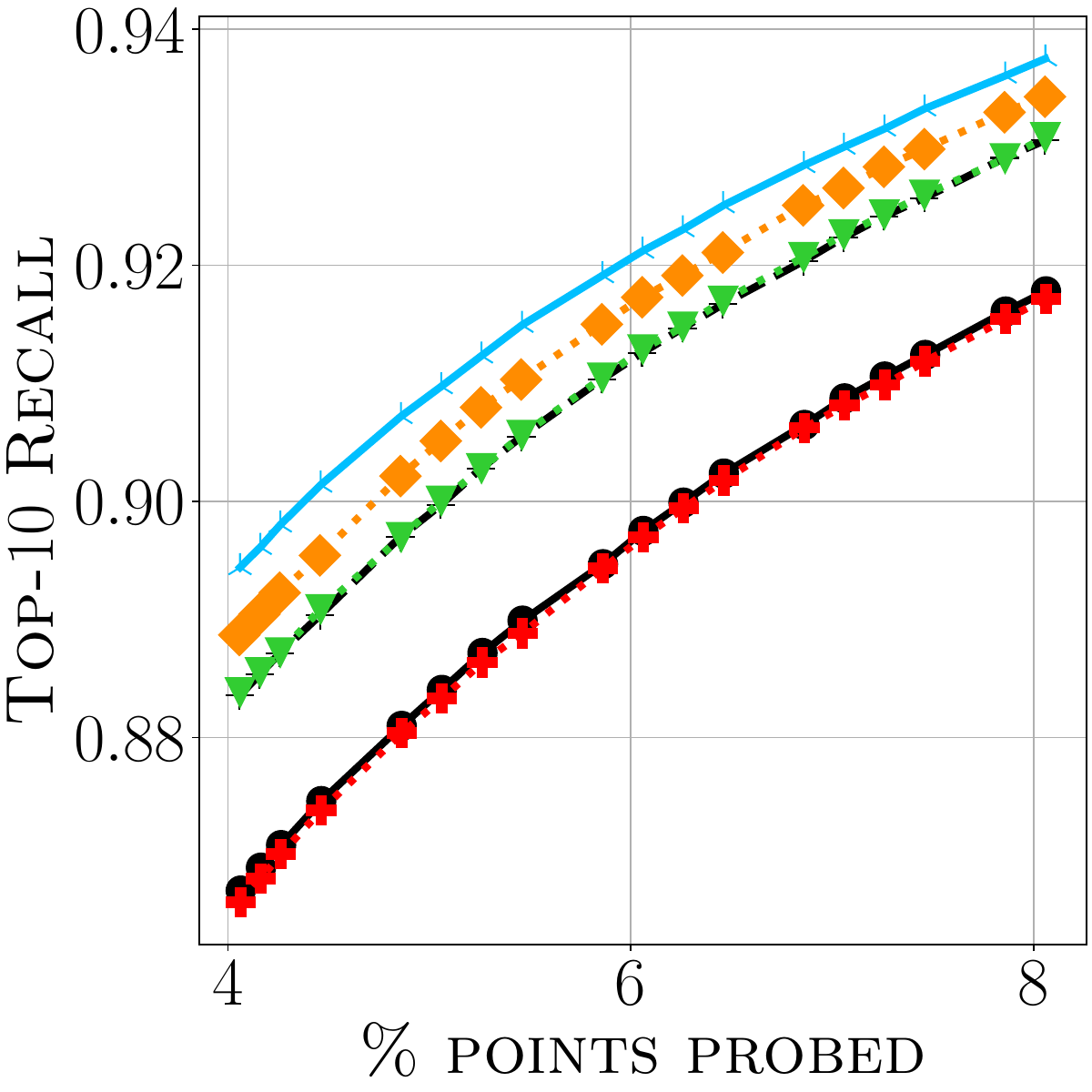}
        \includegraphics[width=0.3\linewidth]{figures/spherical_kmeans/glove_200_cosine_top100.pdf}
    }
}
\caption{Top-$k$ recall vs. volume of probed data. Partitioning is with Spherical KMeans. \scann has parameter $T$, \subpartition $t$ (leading to $t+2$ sub-partitions per shard), and \optimist rank $t$ and degree of optimism $\delta$.}
\label{figure:spherical-kmeans:full}
\end{center}
\end{figure}

\begin{figure}[h]
\ContinuedFloat
\begin{center}
\centerline{
    \subfloat[\textimage]{
        \includegraphics[width=0.3\linewidth]{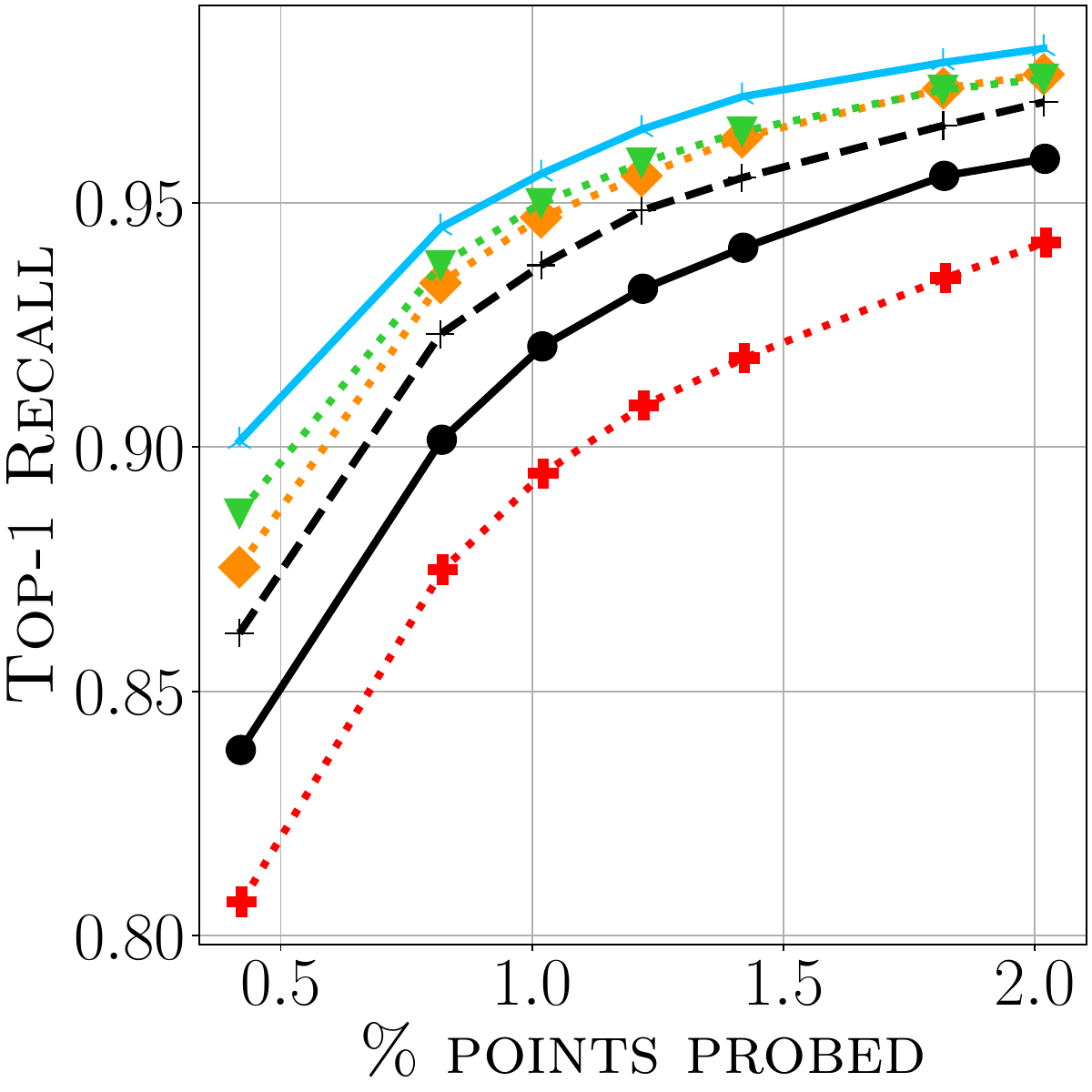}
        \includegraphics[width=0.3\linewidth]{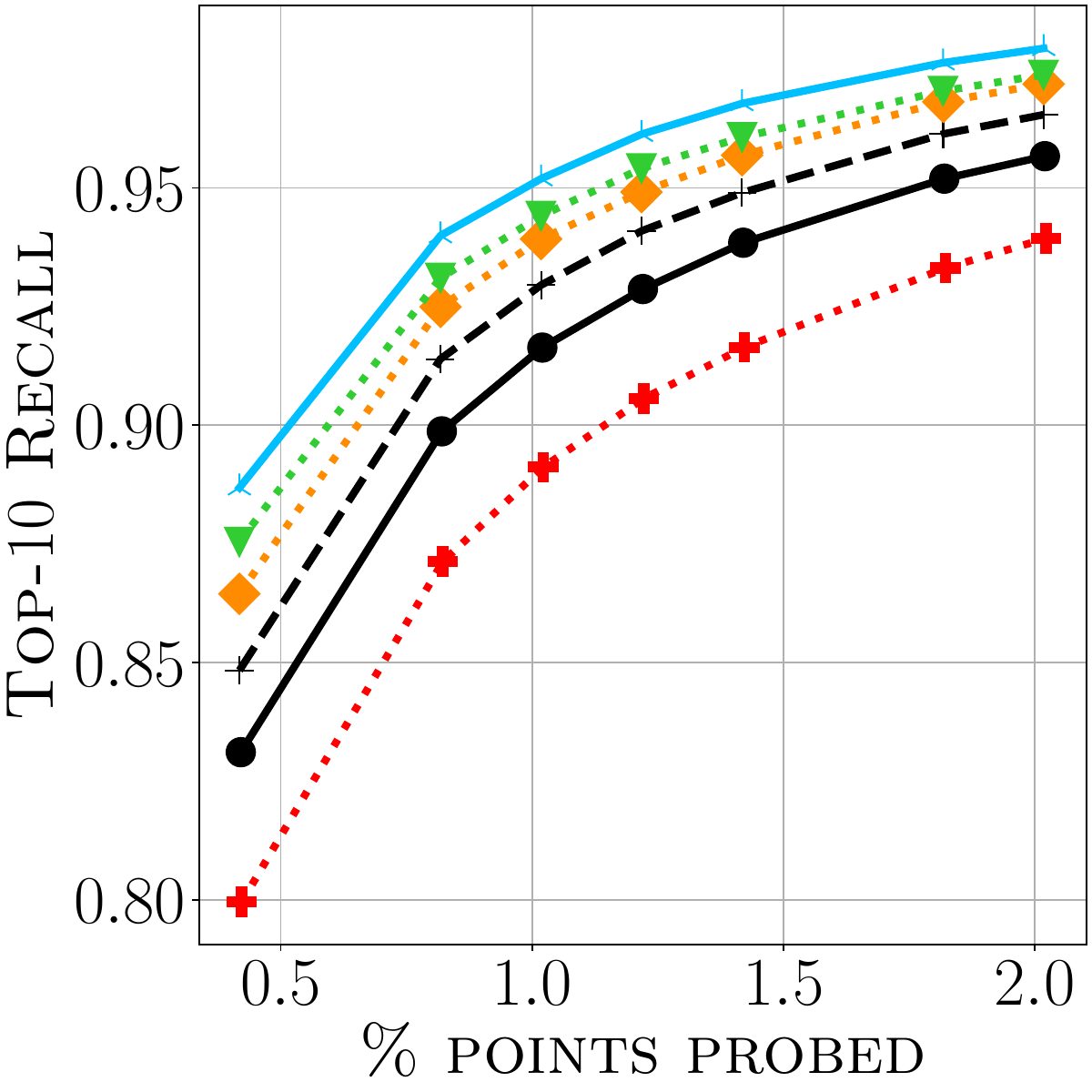}
        \includegraphics[width=0.3\linewidth]{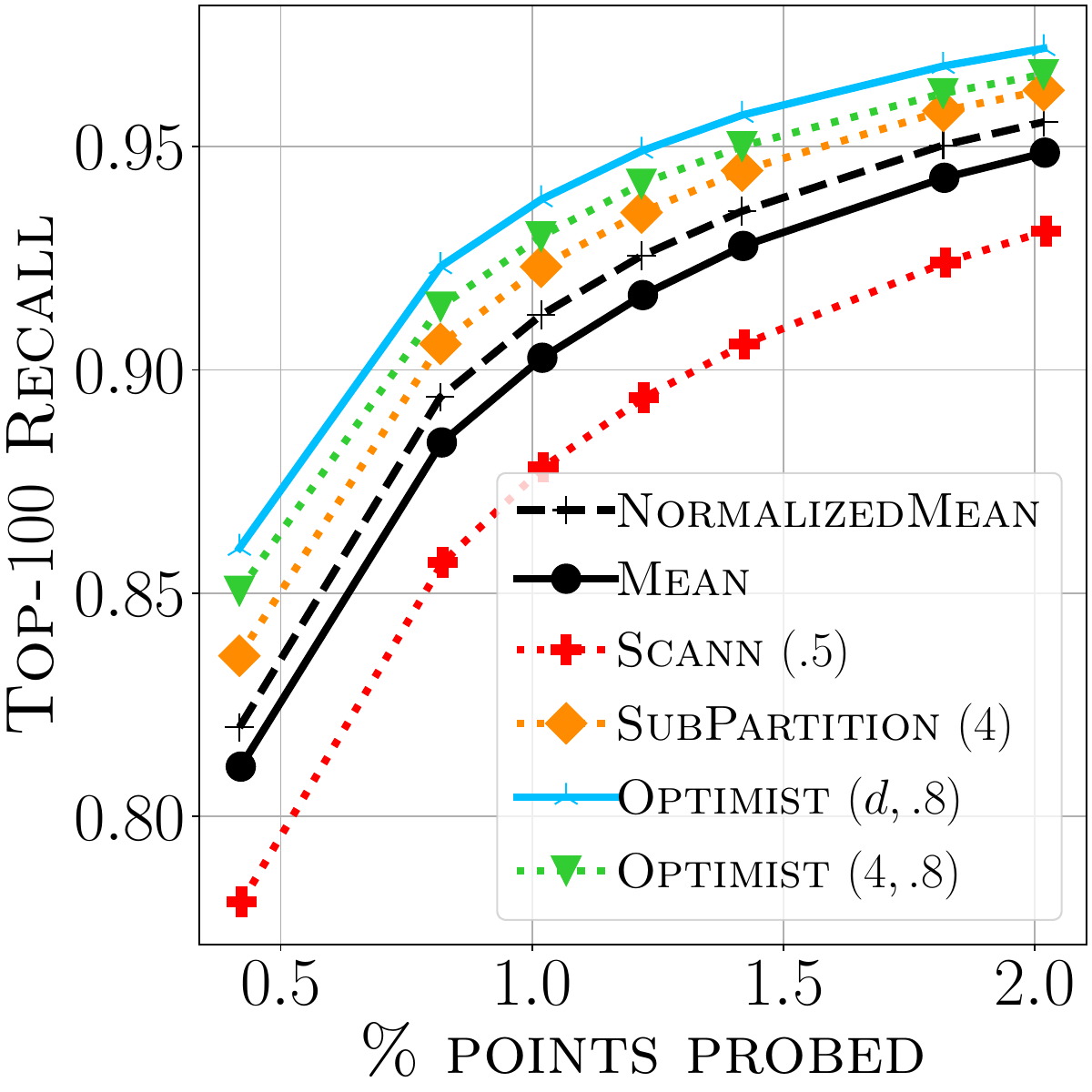}
    }
}
\centerline{
    \subfloat[\msmarco]{
        \includegraphics[width=0.3\linewidth]{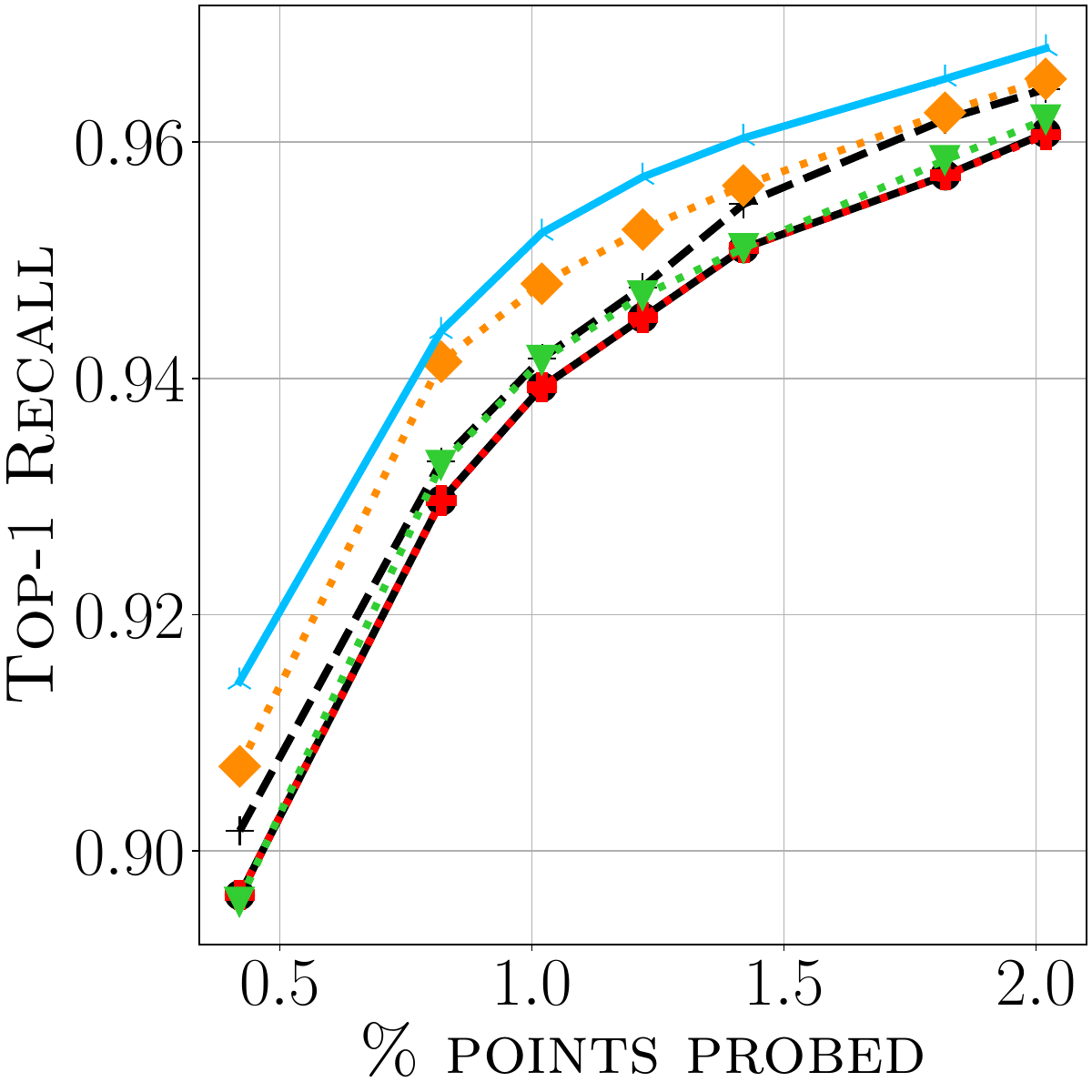}
        \includegraphics[width=0.3\linewidth]{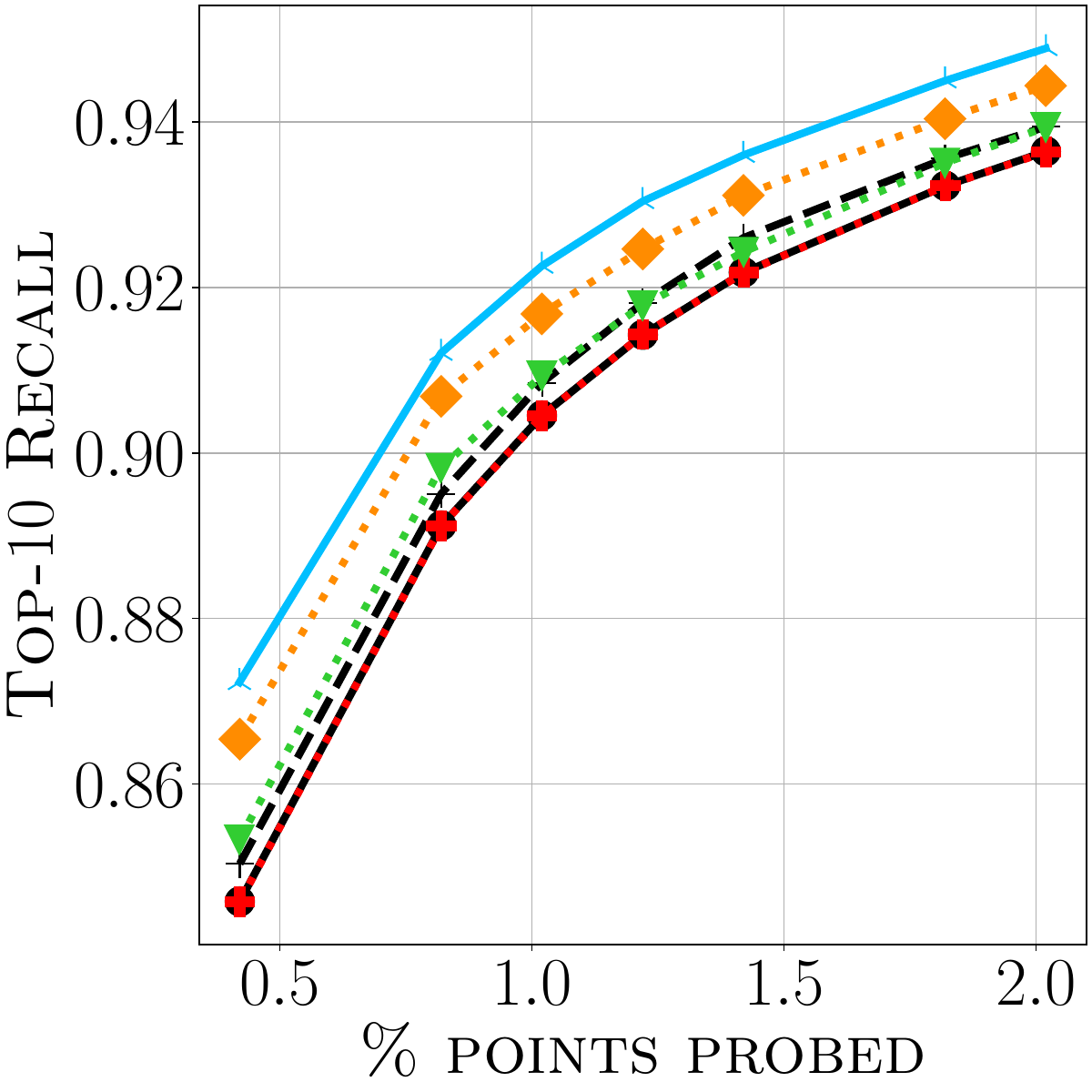}
        \includegraphics[width=0.3\linewidth]{figures/spherical_kmeans/msmarco_cosine_top100.pdf}
    }
}
\centerline{
    \subfloat[\deep]{
        \includegraphics[width=0.3\linewidth]{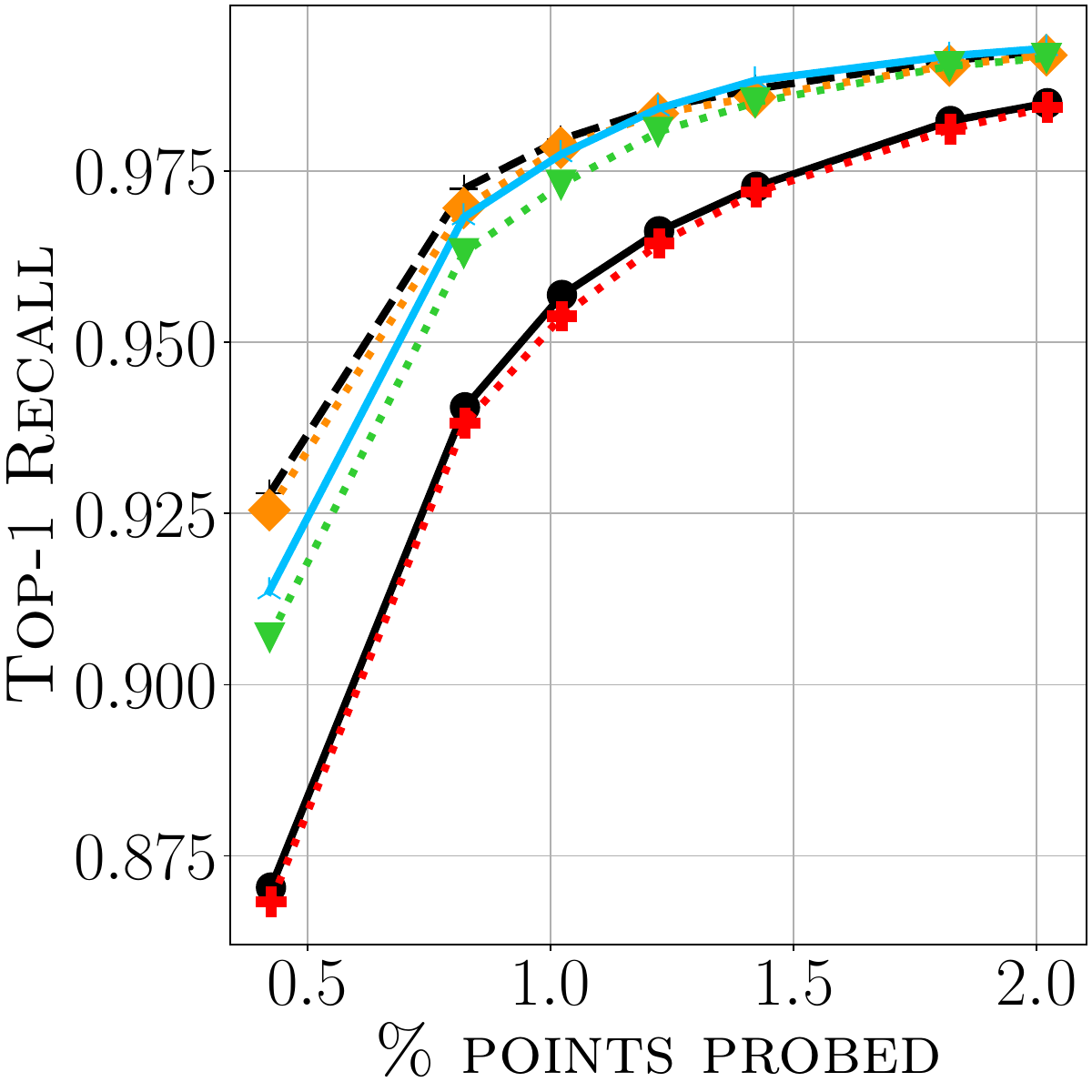}
        \includegraphics[width=0.3\linewidth]{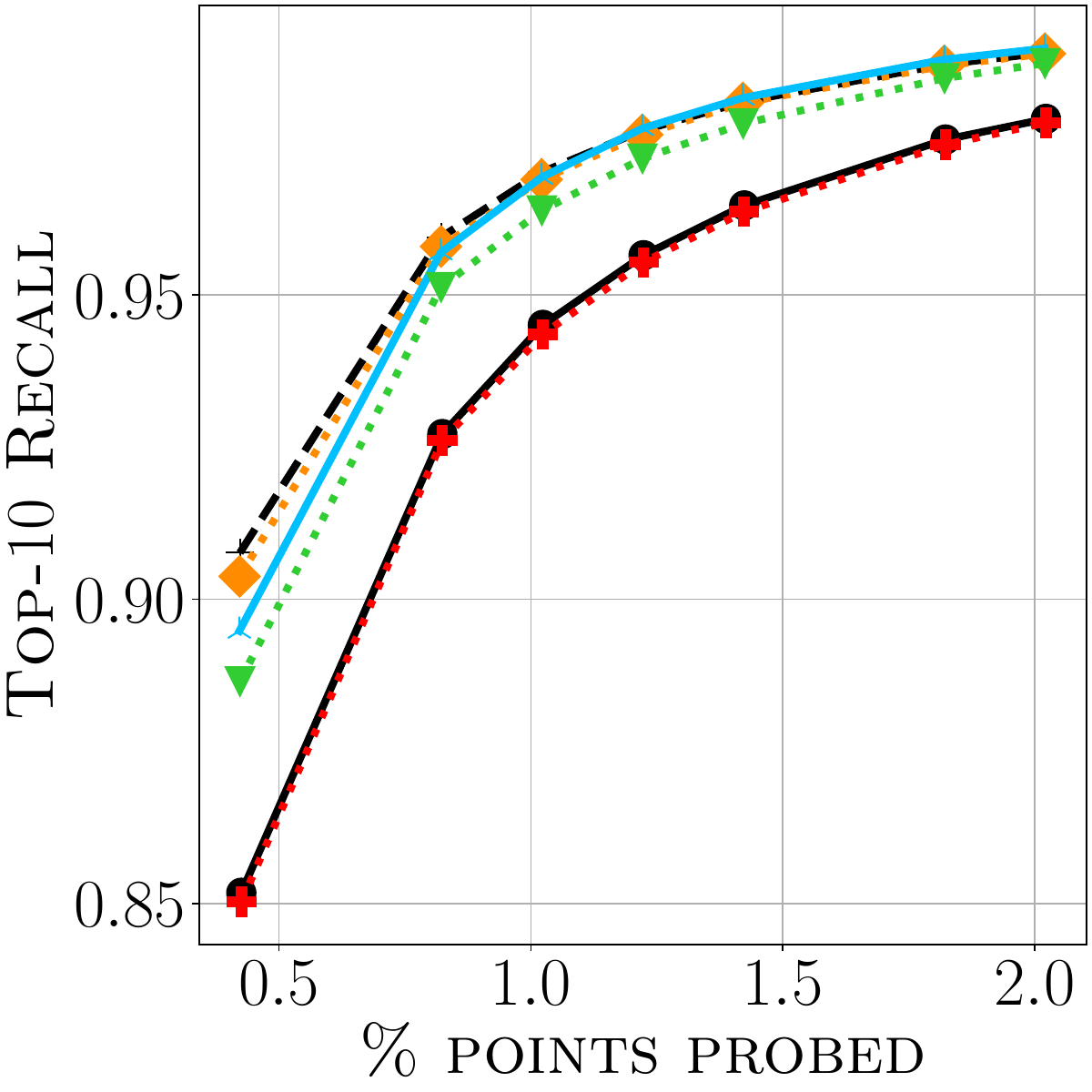}
        \includegraphics[width=0.3\linewidth]{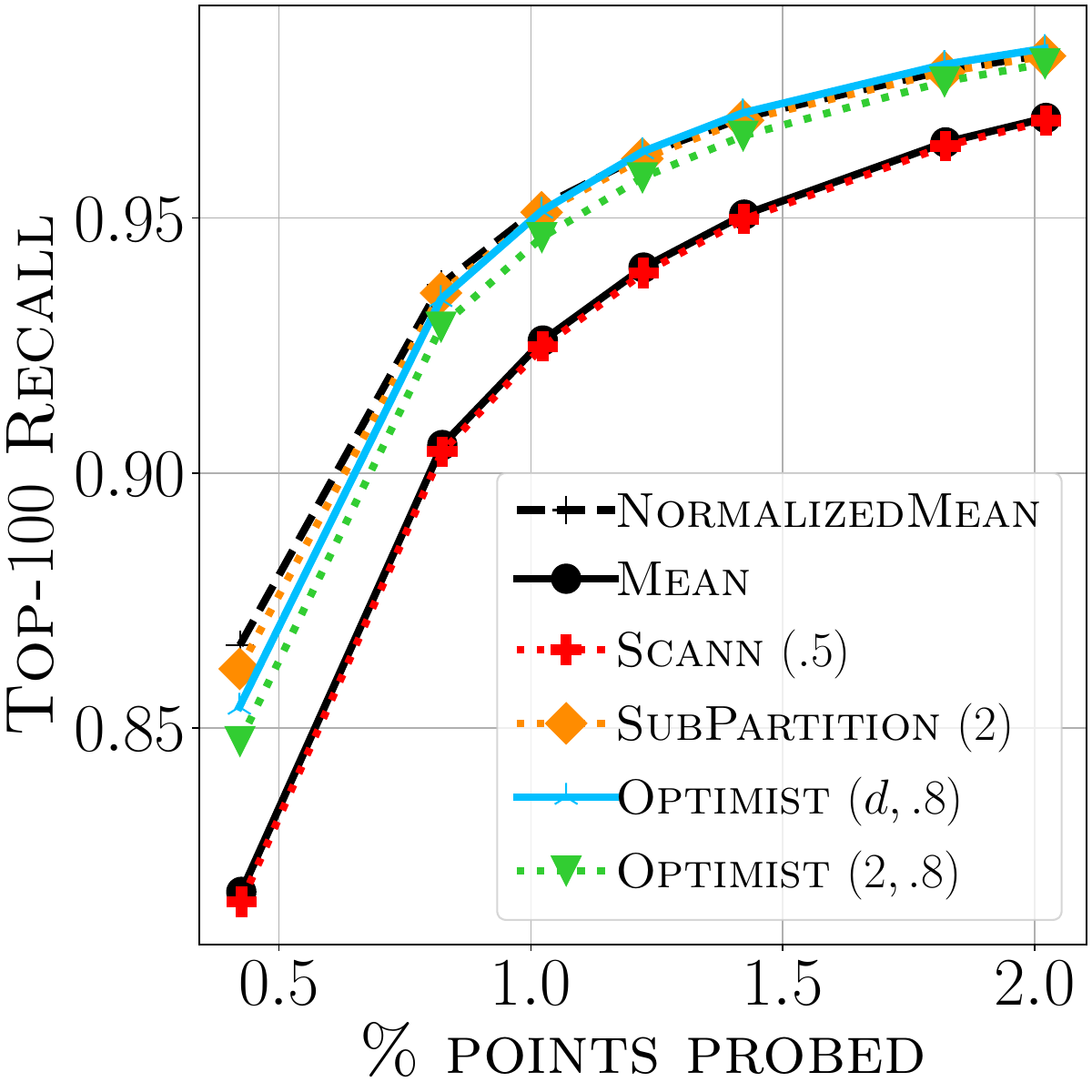}
    }
}
\caption{Top-$k$ recall vs. volume of probed data. Partitioning is with Spherical KMeans. \scann has parameter $T$, \subpartition $t$ (leading to $t+2$ sub-partitions per shard), and \optimist rank $t$ and degree of optimism $\delta$.}
\end{center}
\end{figure}

\FloatBarrier

\newpage
\section{Experiments with standard KMeans}
\label{appendix:experiments:standard-kmeans}

\begin{figure}[h]
\begin{center}
\centerline{
    \subfloat[\nq]{
        \includegraphics[width=0.3\linewidth]{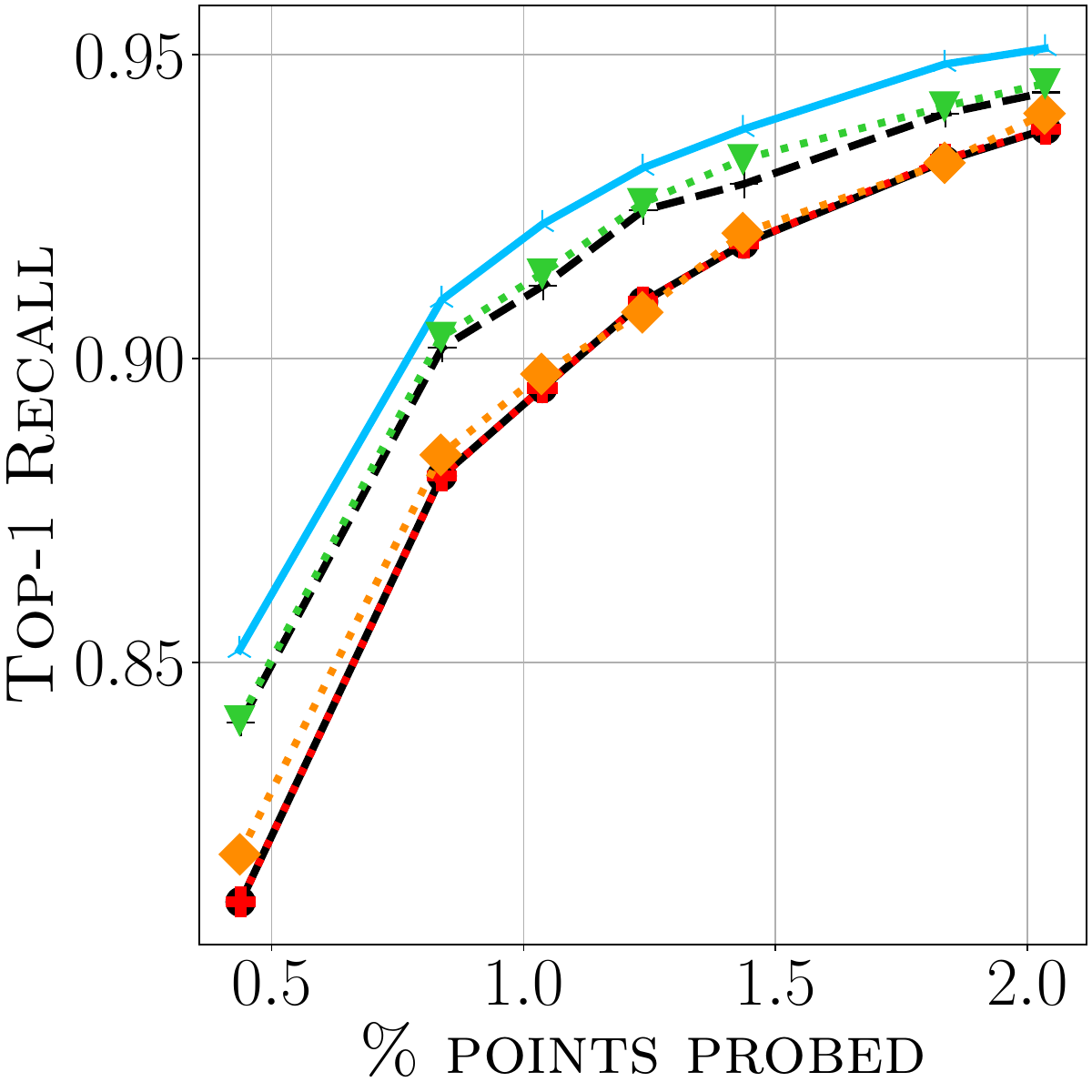}
        \includegraphics[width=0.3\linewidth]{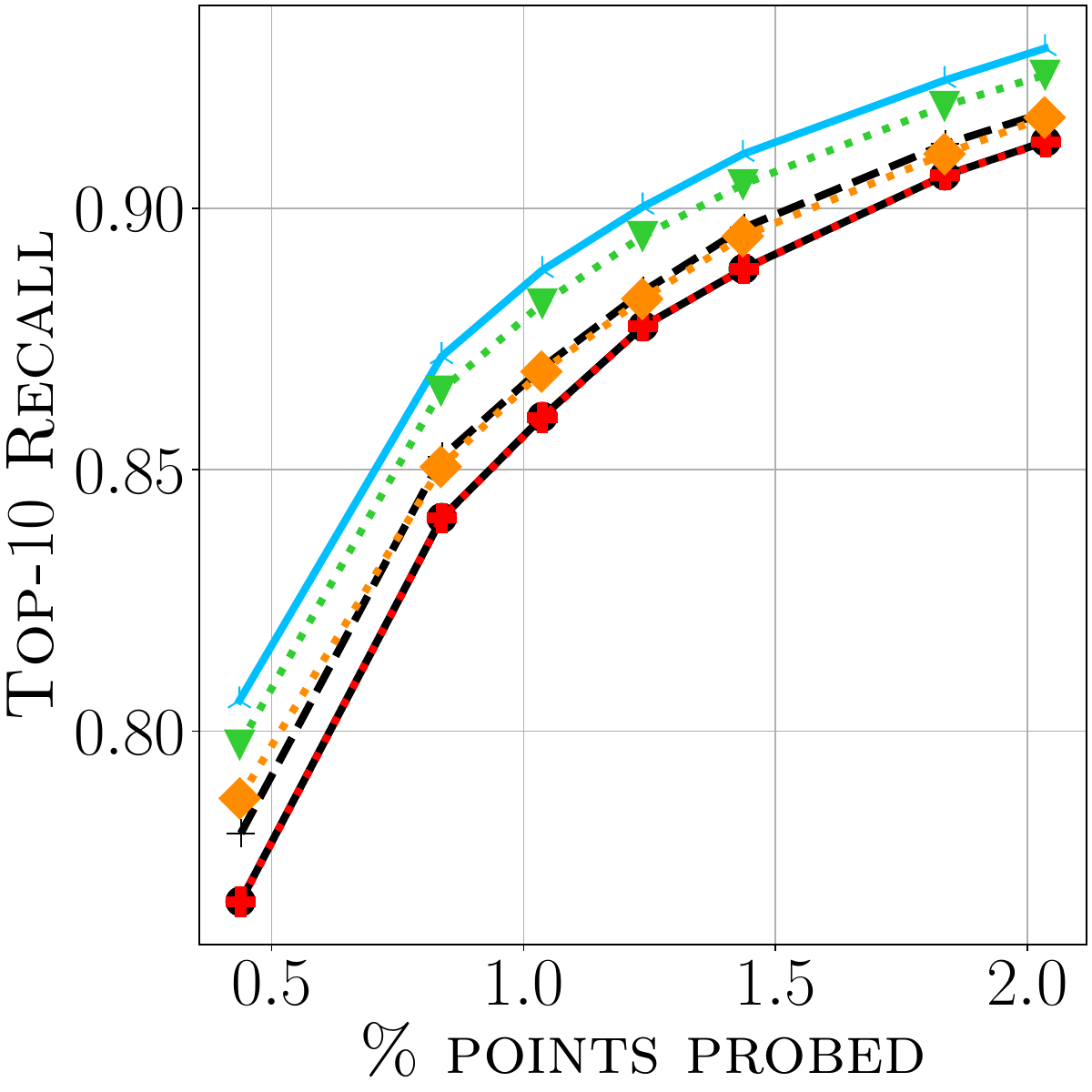}
        \includegraphics[width=0.3\linewidth]{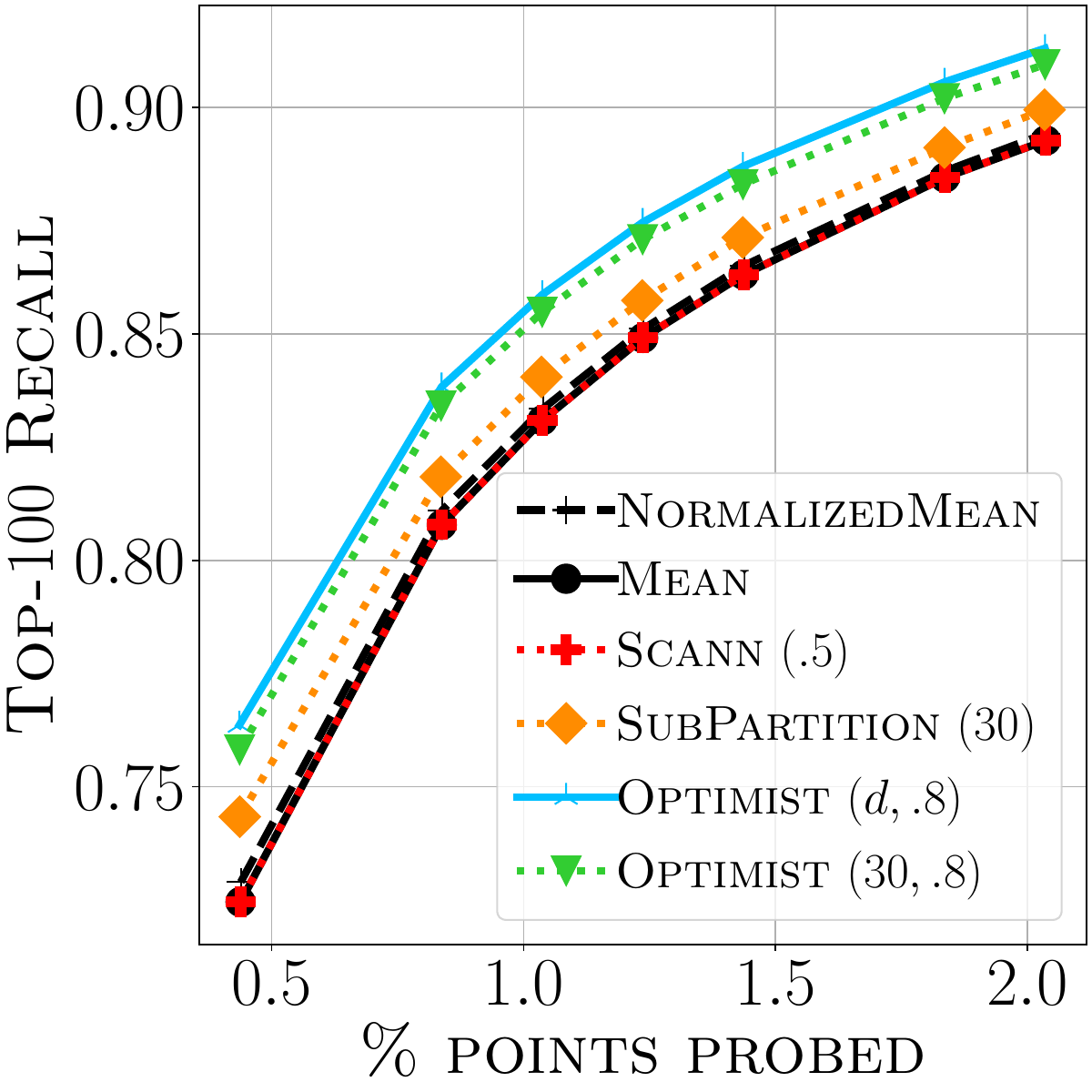}
    }
}
\centerline{
    \subfloat[\music]{
        \includegraphics[width=0.3\linewidth]{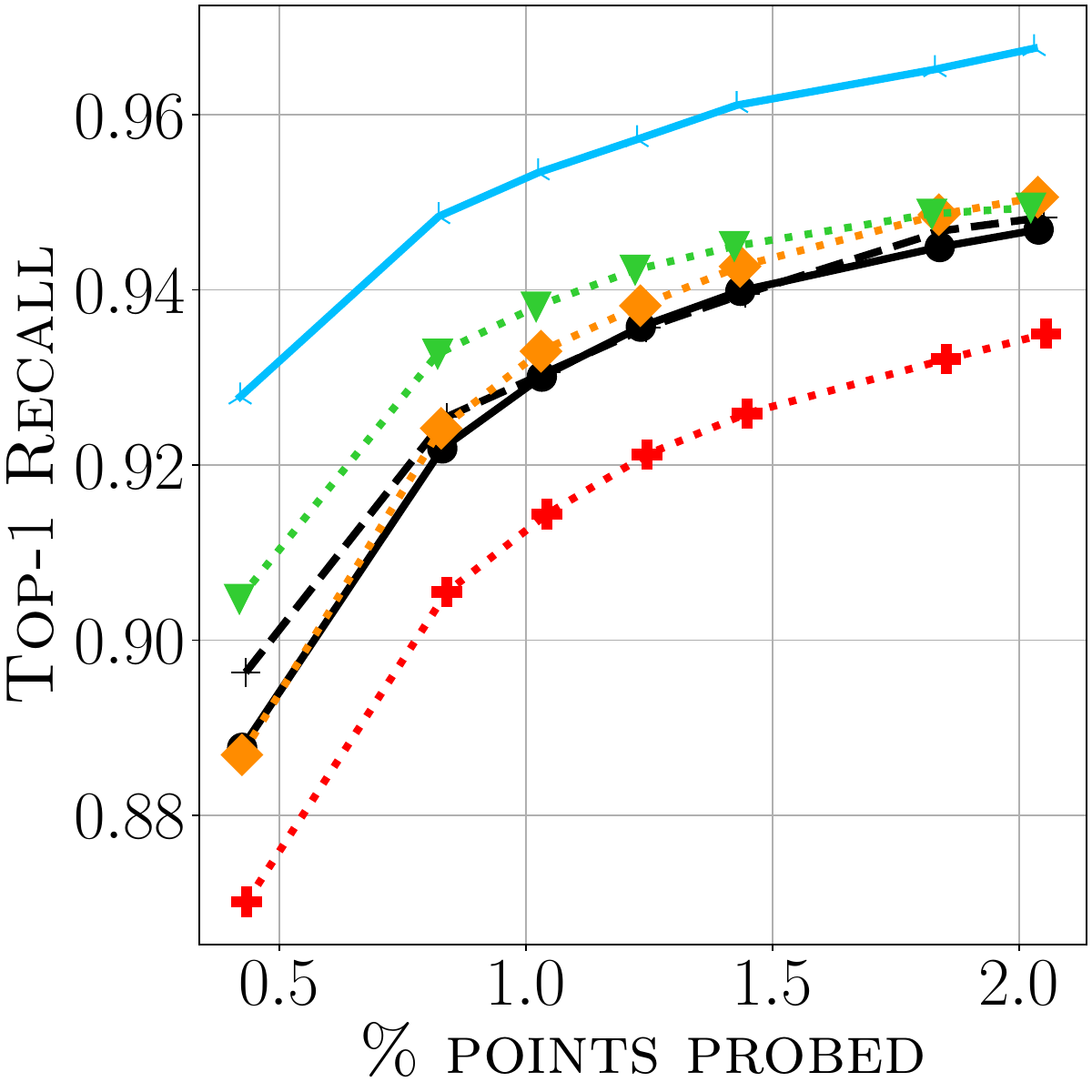}
        \includegraphics[width=0.3\linewidth]{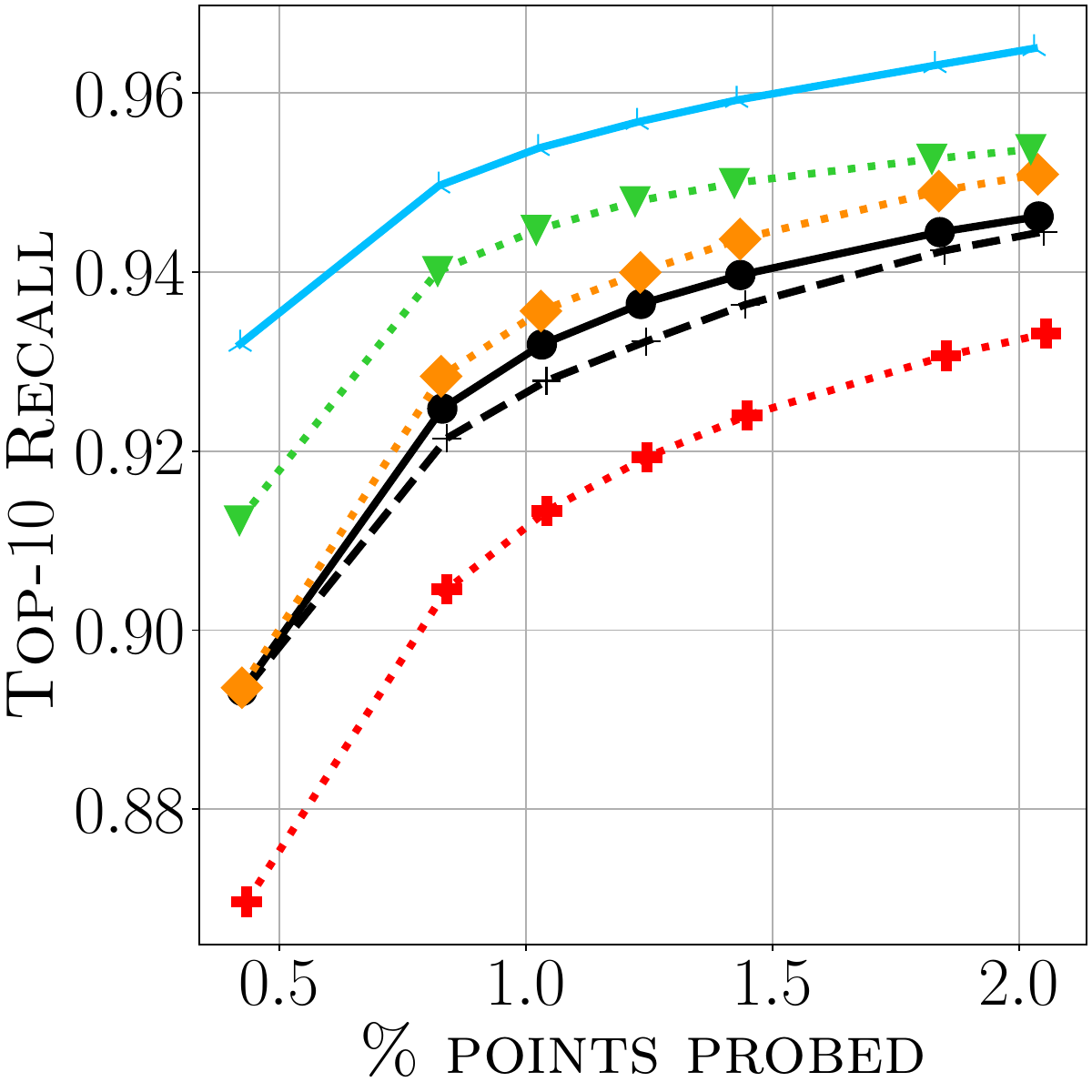}
        \includegraphics[width=0.3\linewidth]{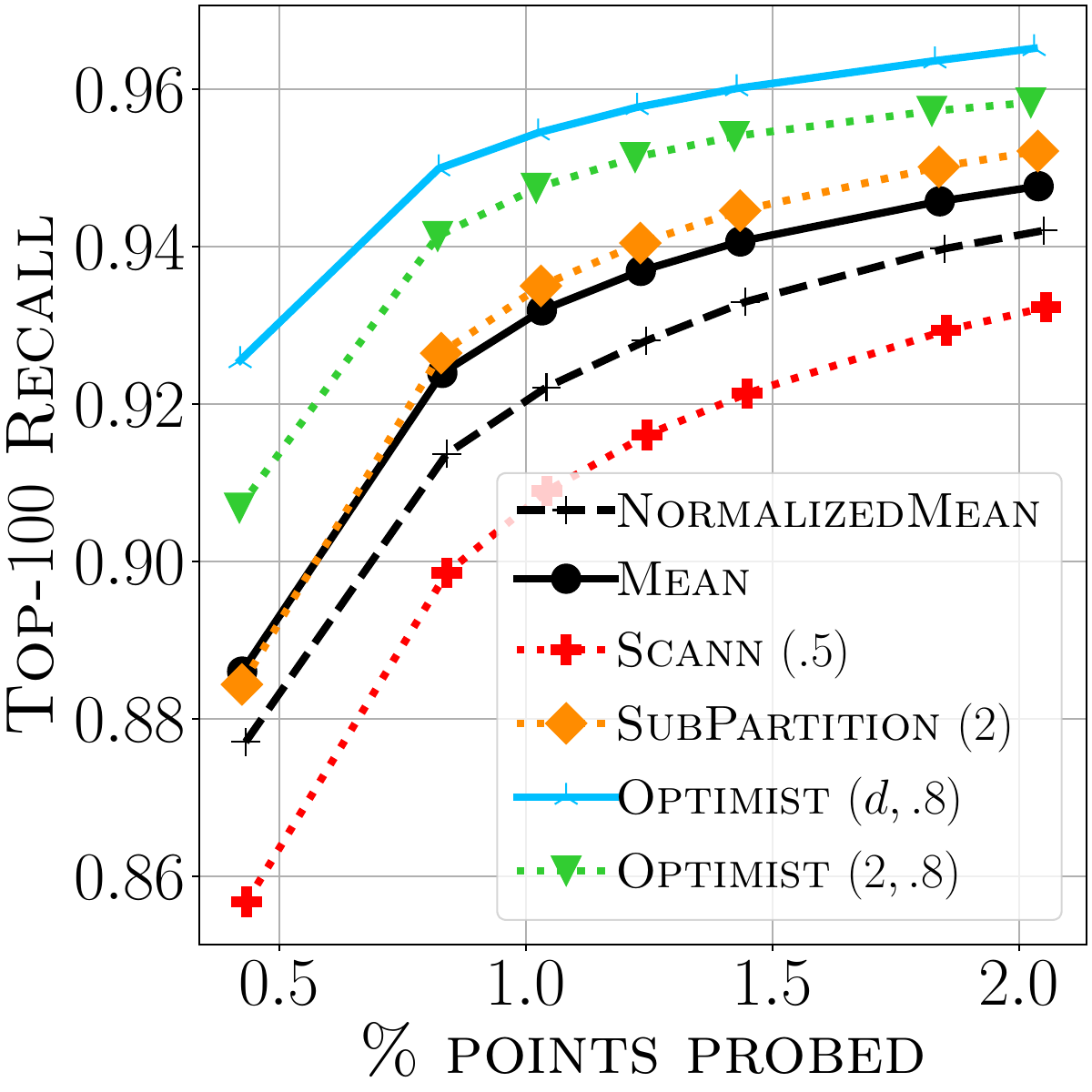}
    }
}
\centerline{
    \subfloat[\glove]{
        \includegraphics[width=0.3\linewidth]{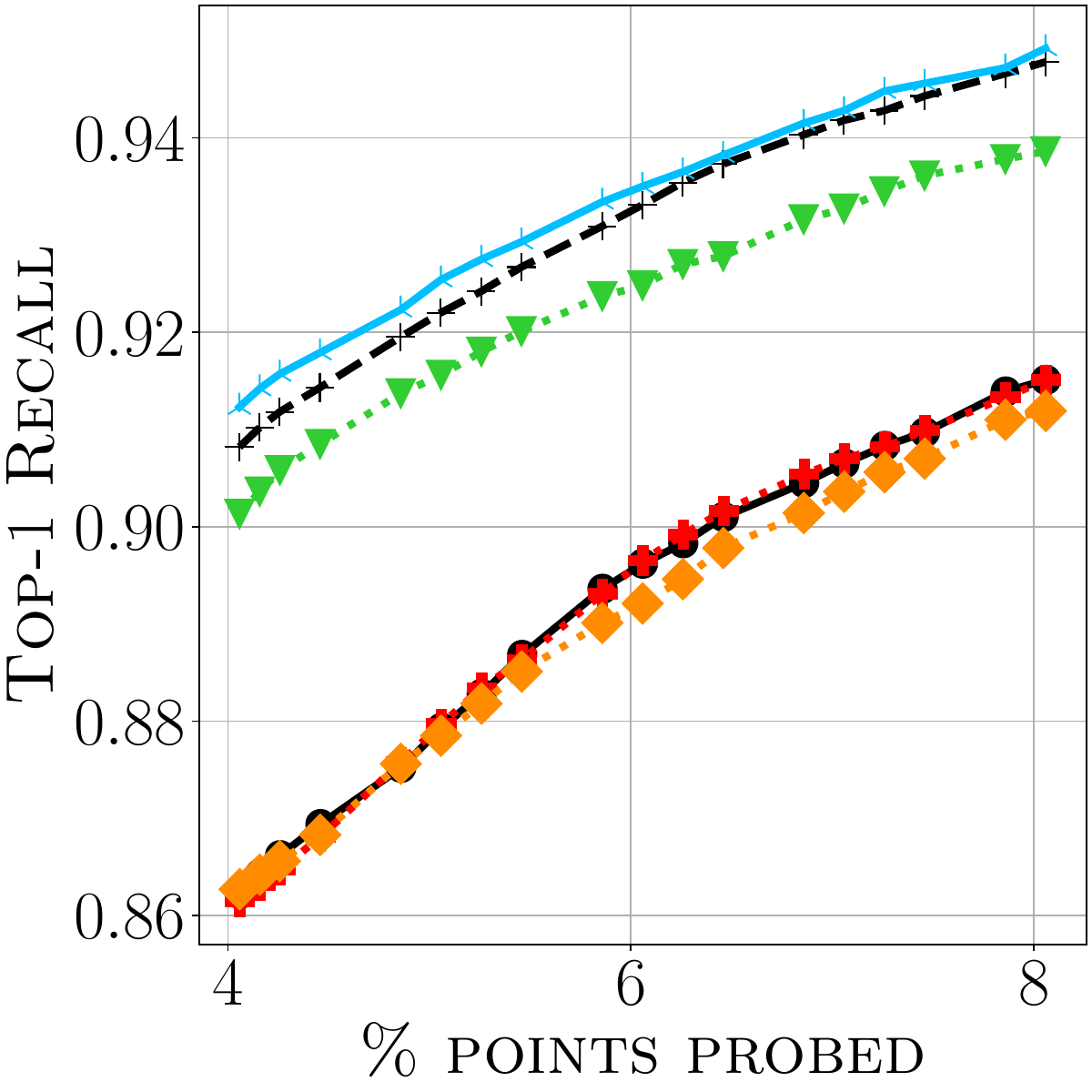}
        \includegraphics[width=0.3\linewidth]{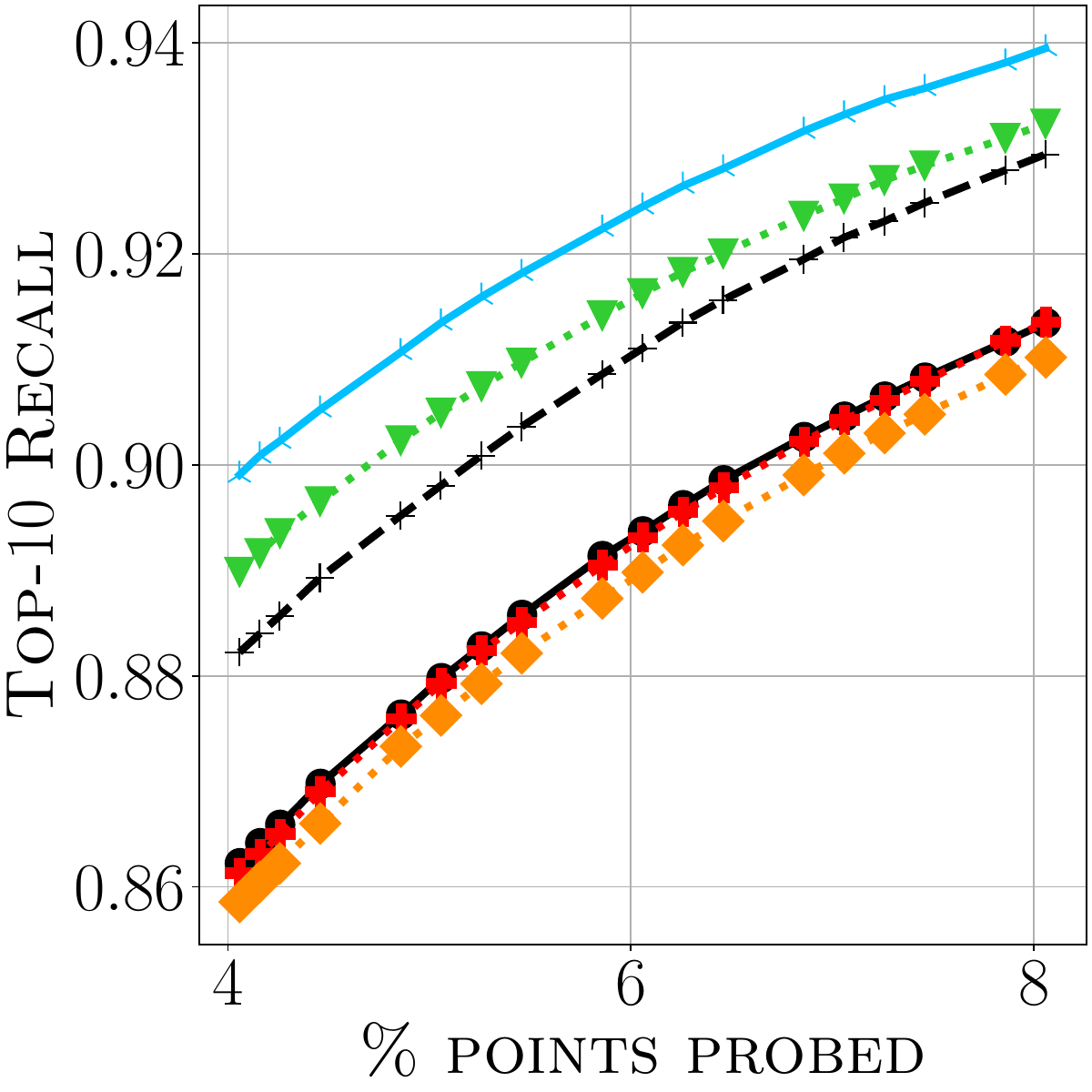}
        \includegraphics[width=0.3\linewidth]{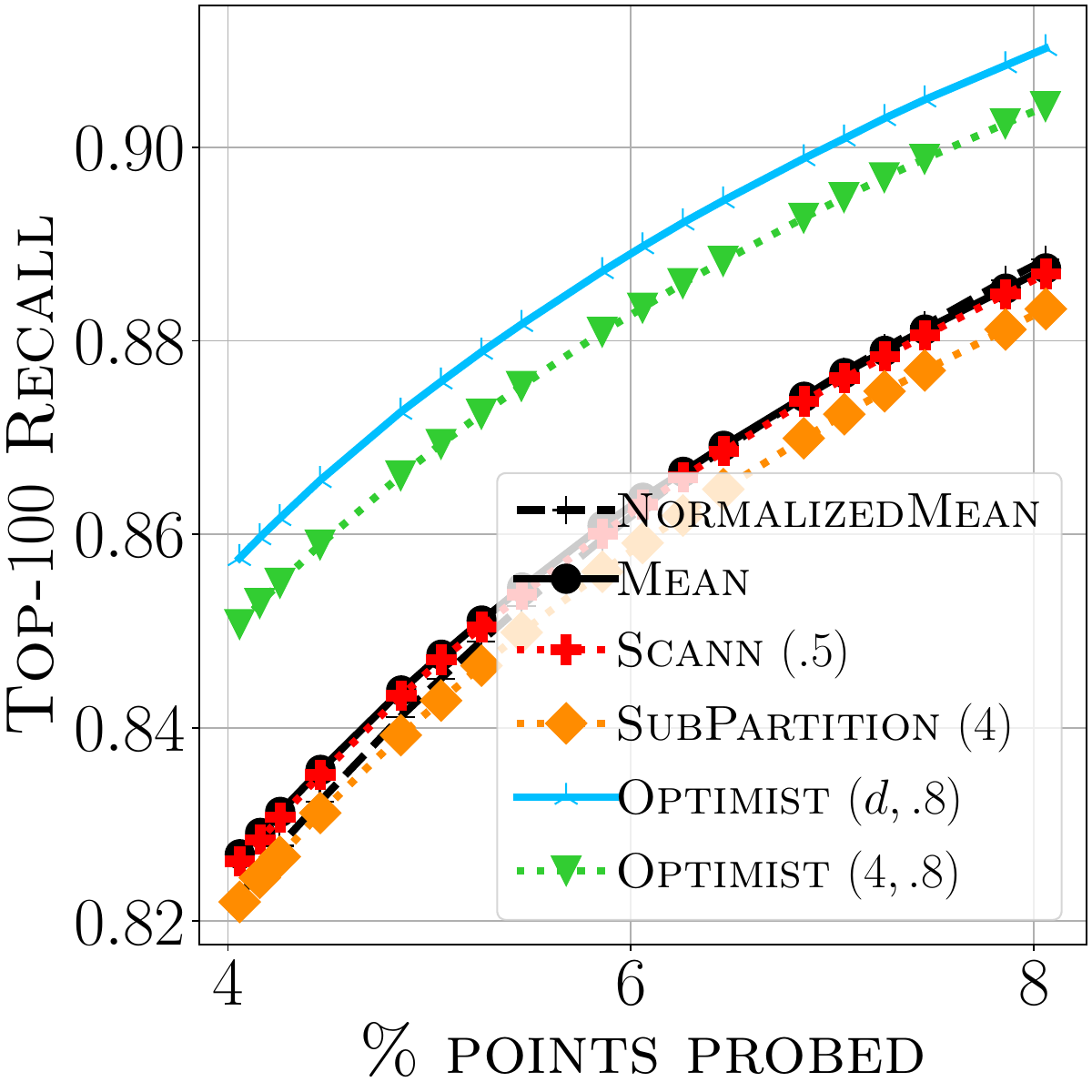}
    }
}
\caption{Top-$k$ recall vs. volume of probed data. Partitioning is with Standard KMeans. \scann has parameter $T$, \subpartition $t$ (leading to $t+2$ sub-partitions per shard), and \optimist rank $t$ and degree of optimism $\delta$.}
\label{figure:standard-kmeans:full}
\end{center}
\end{figure}

\begin{figure}[h]
\ContinuedFloat
\begin{center}
\centerline{
    \subfloat[\textimage]{
        \includegraphics[width=0.3\linewidth]{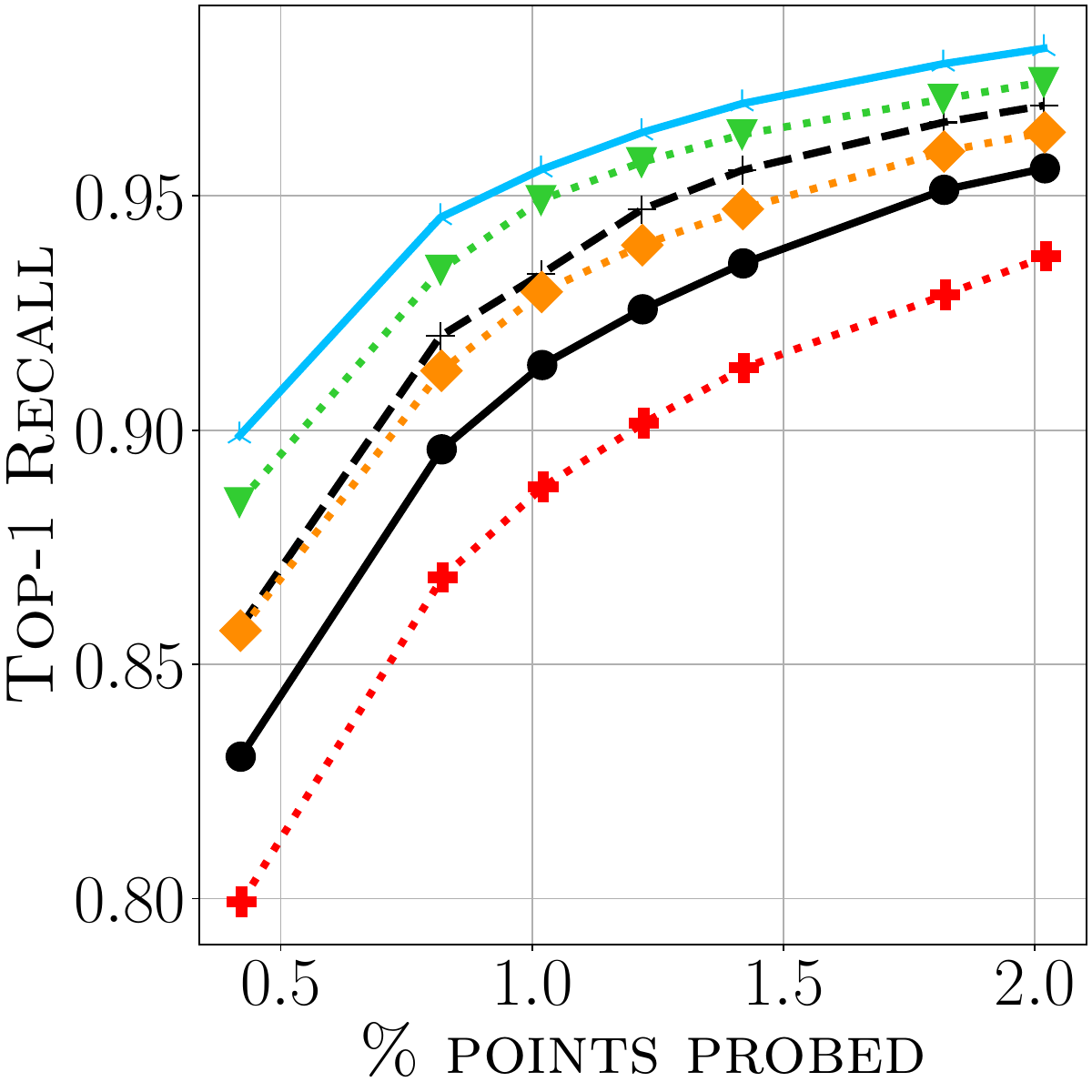}
        \includegraphics[width=0.3\linewidth]{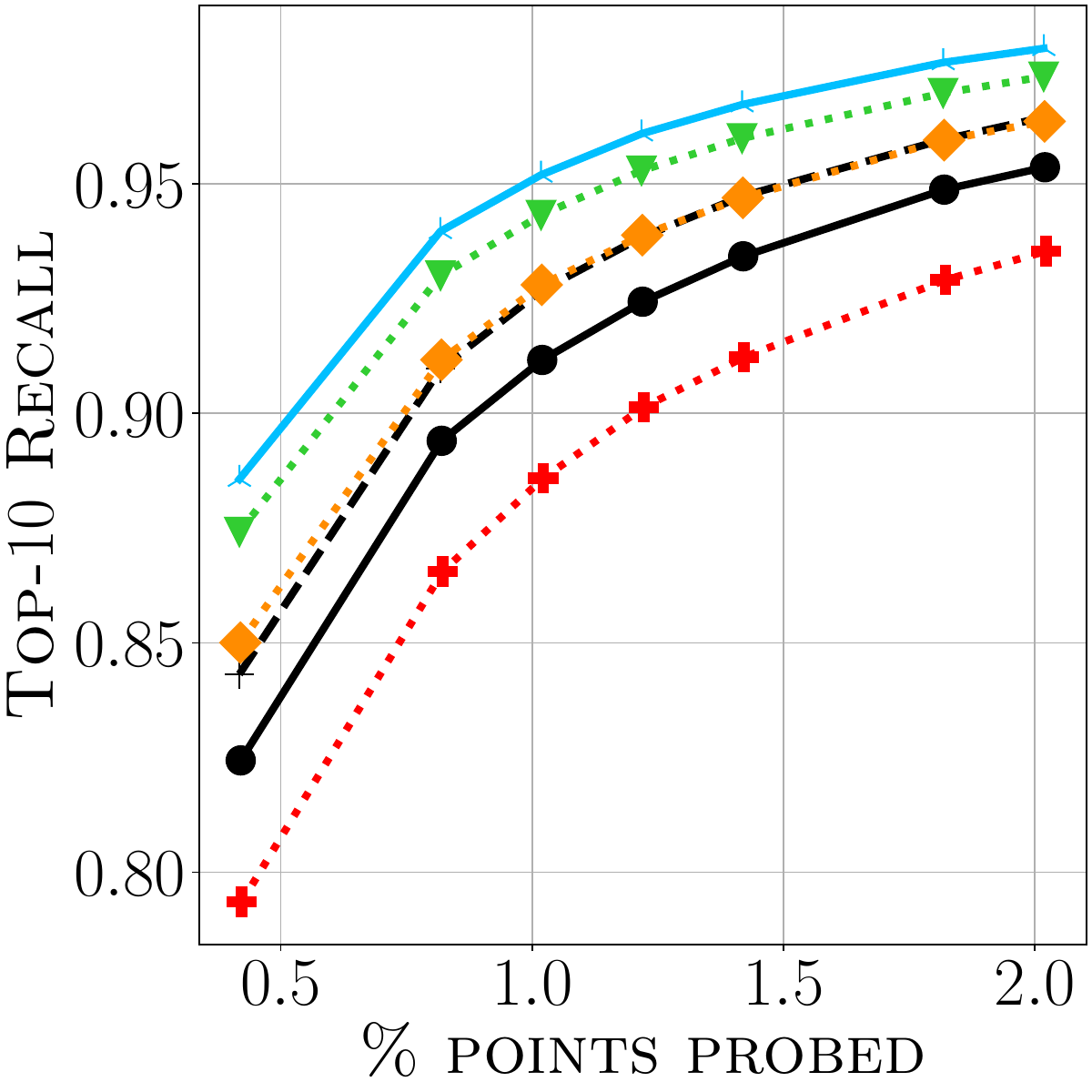}
        \includegraphics[width=0.3\linewidth]{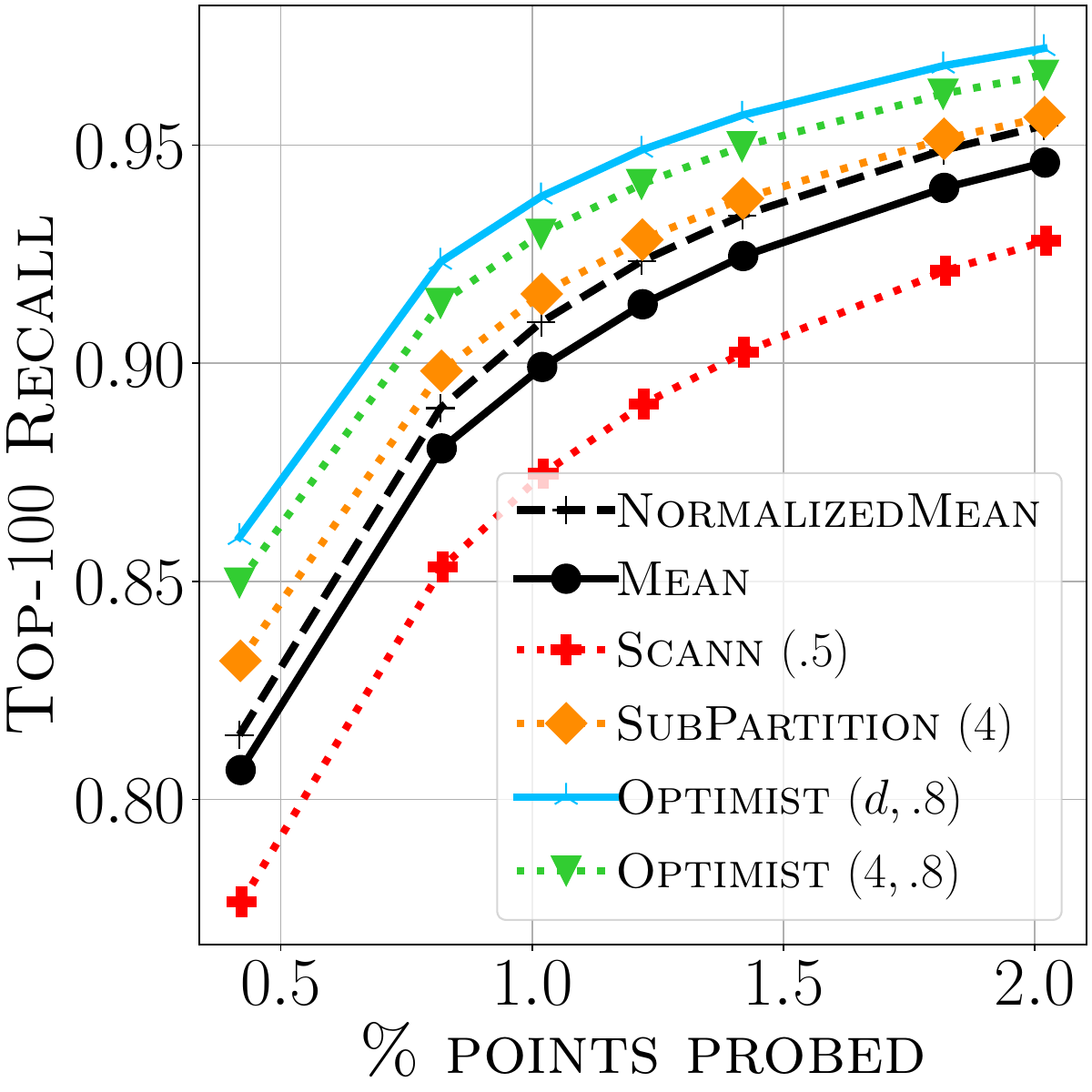}
    }
}
\centerline{
    \subfloat[\msmarco]{
        \includegraphics[width=0.3\linewidth]{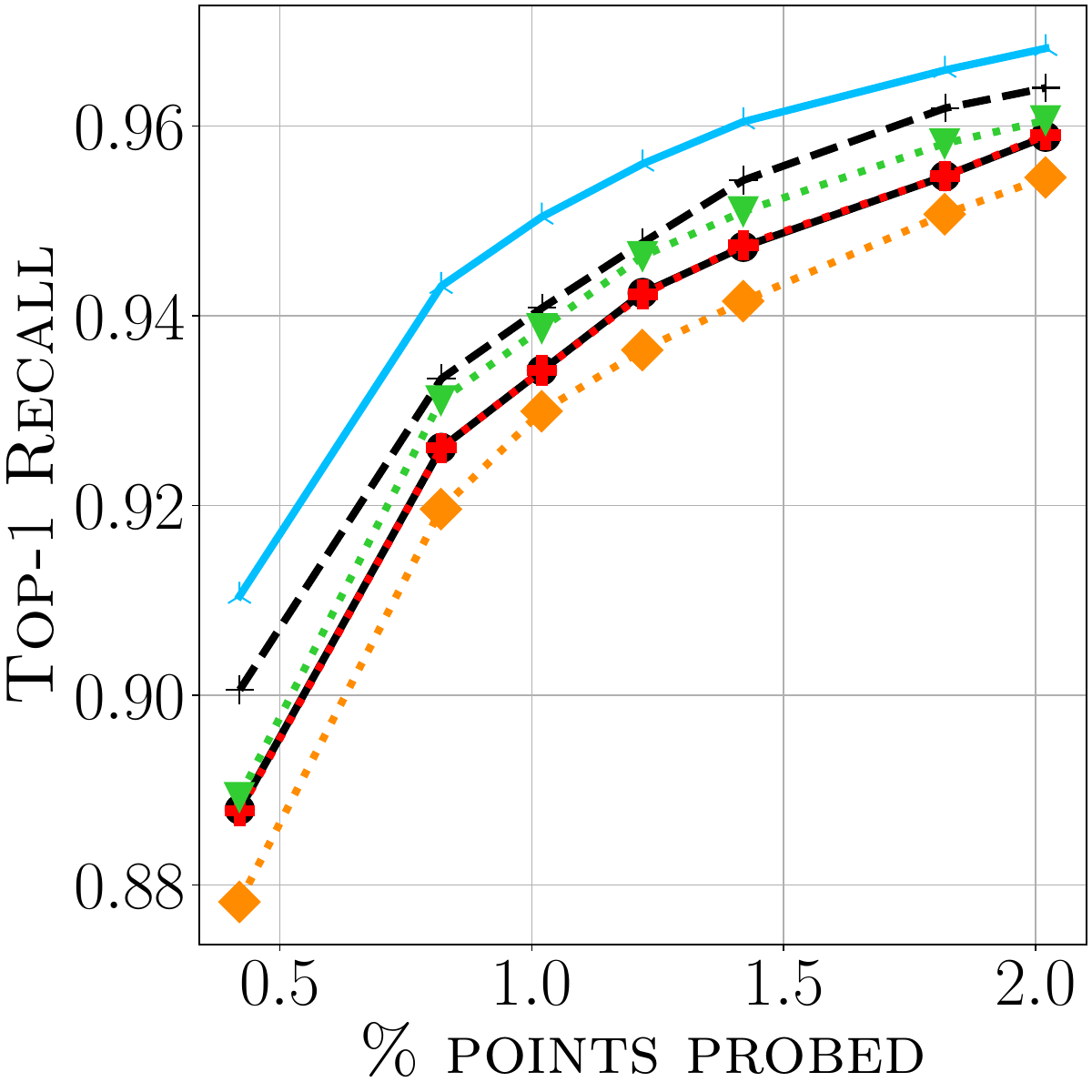}
        \includegraphics[width=0.3\linewidth]{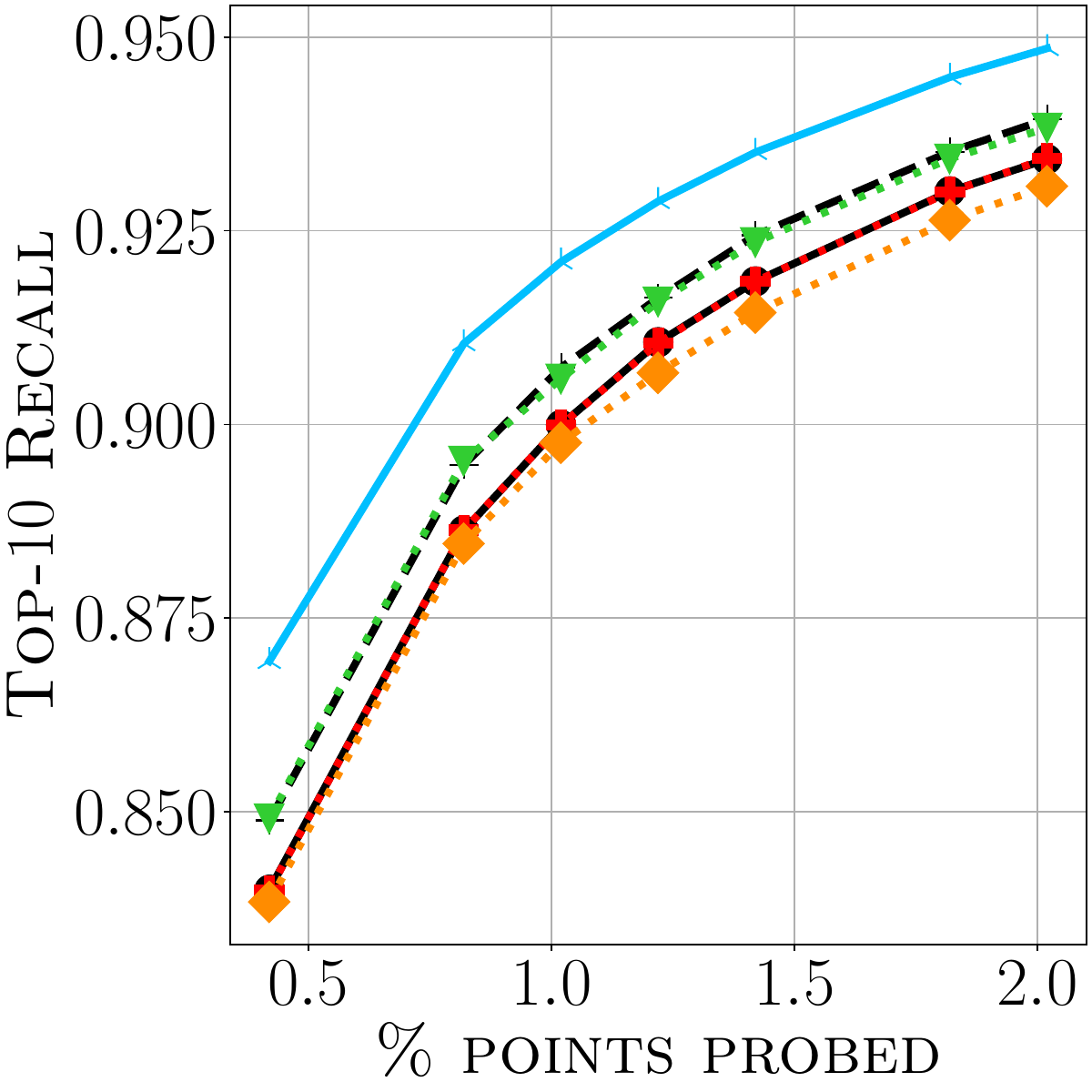}
        \includegraphics[width=0.3\linewidth]{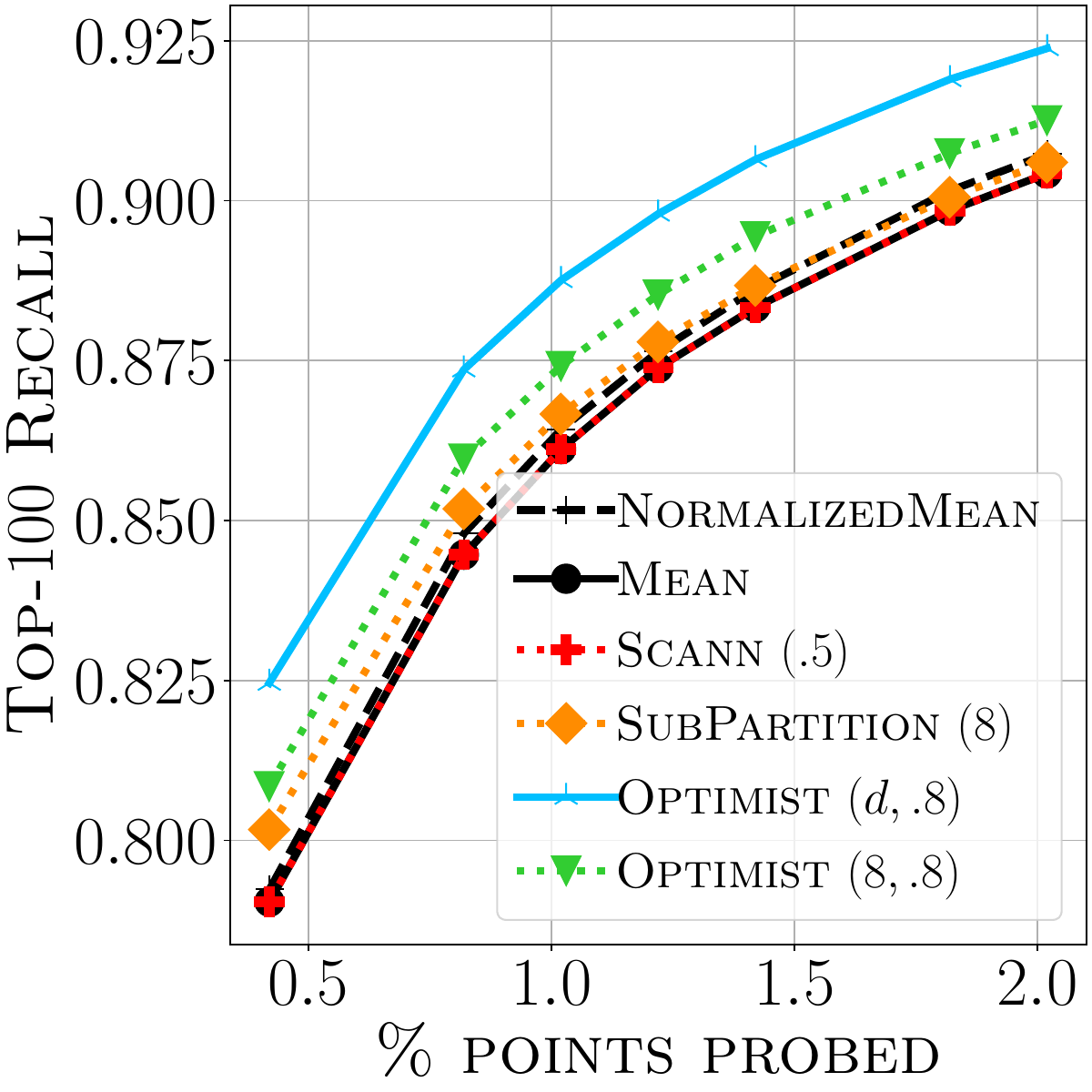}
    }
}
\centerline{
    \subfloat[\deep]{
        \includegraphics[width=0.3\linewidth]{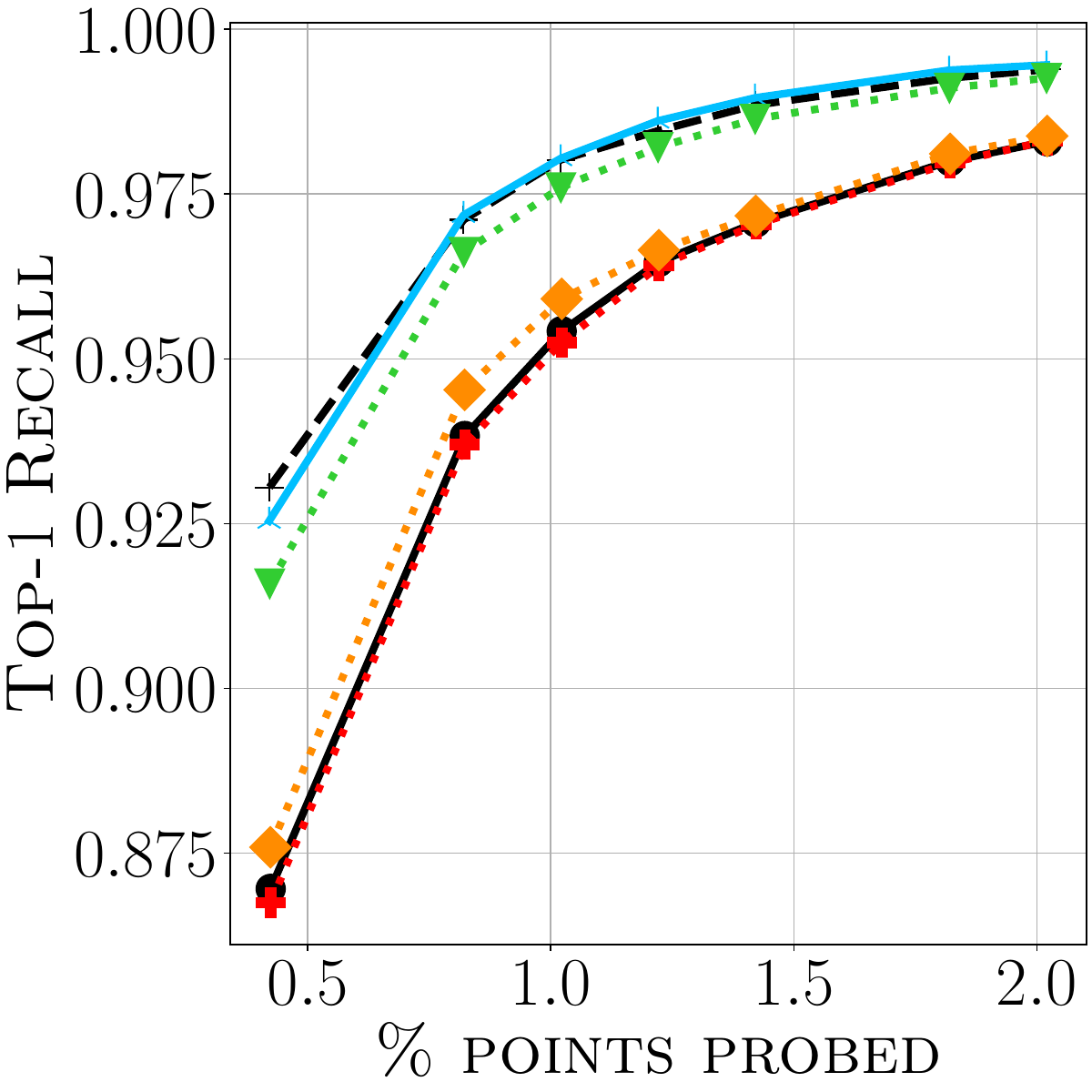}
        \includegraphics[width=0.3\linewidth]{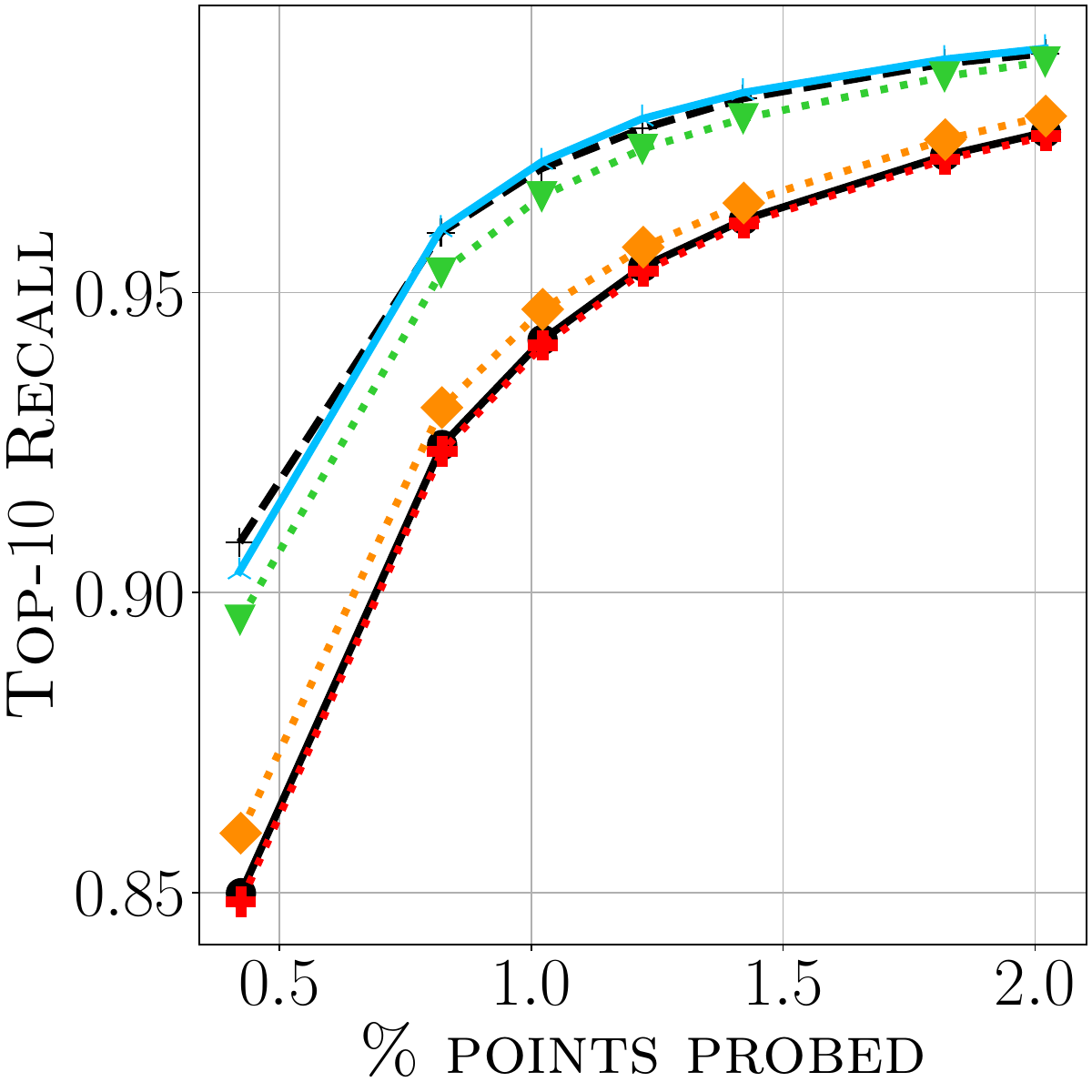}
        \includegraphics[width=0.3\linewidth]{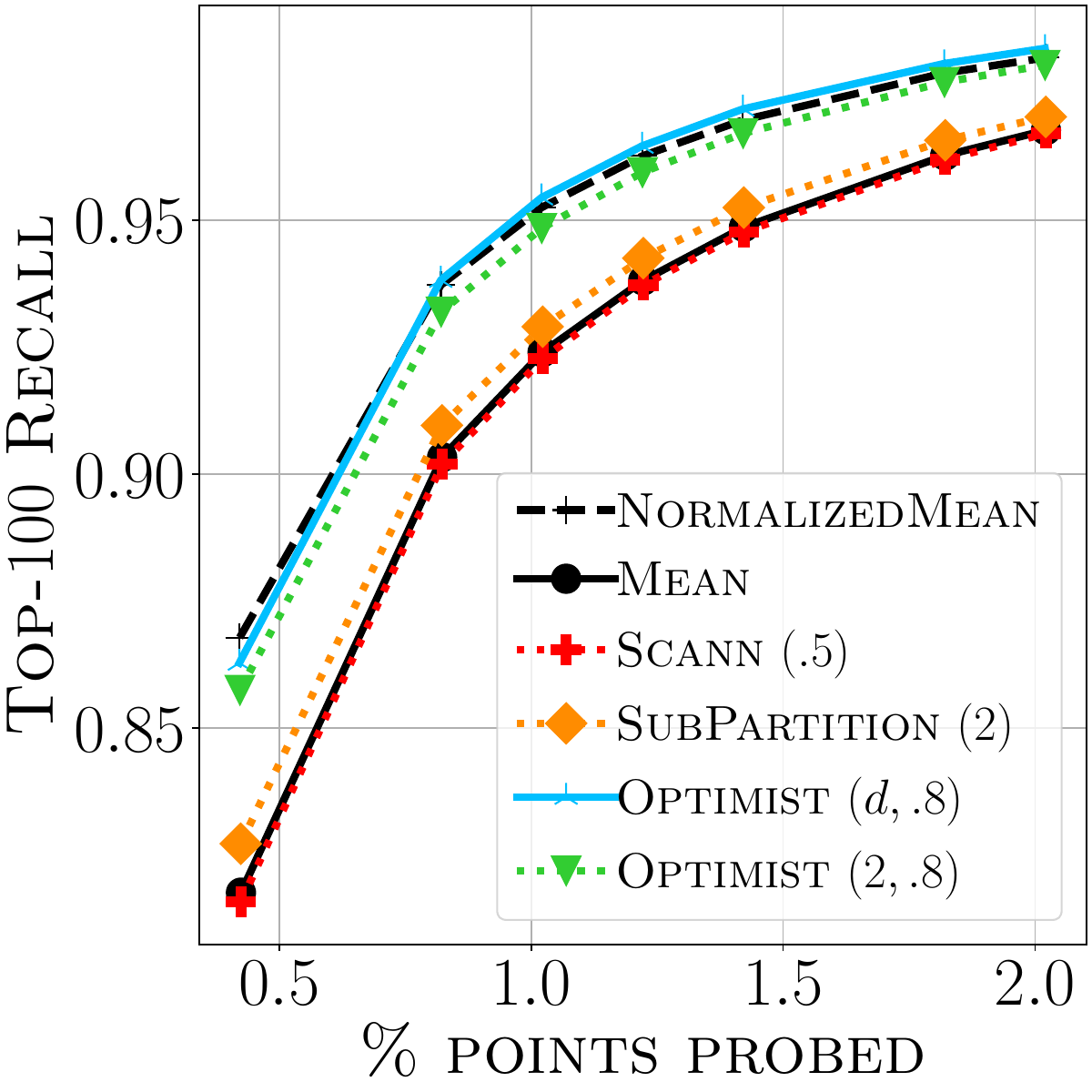}
    }
}
\caption{Top-$k$ recall vs. volume of probed data. Partitioning is with Standard KMeans. \scann has parameter $T$, \subpartition $t$ (leading to $t+2$ sub-partitions per shard), and \optimist rank $t$ and degree of optimism $\delta$.}
\end{center}
\end{figure}

\FloatBarrier
\newpage
\section{Experiments with GMM}
\label{appendix:experiments:gmm}

\begin{figure*}[ht]
\begin{center}
\centerline{
    \subfloat[\music]{
        \includegraphics[width=0.26\linewidth]{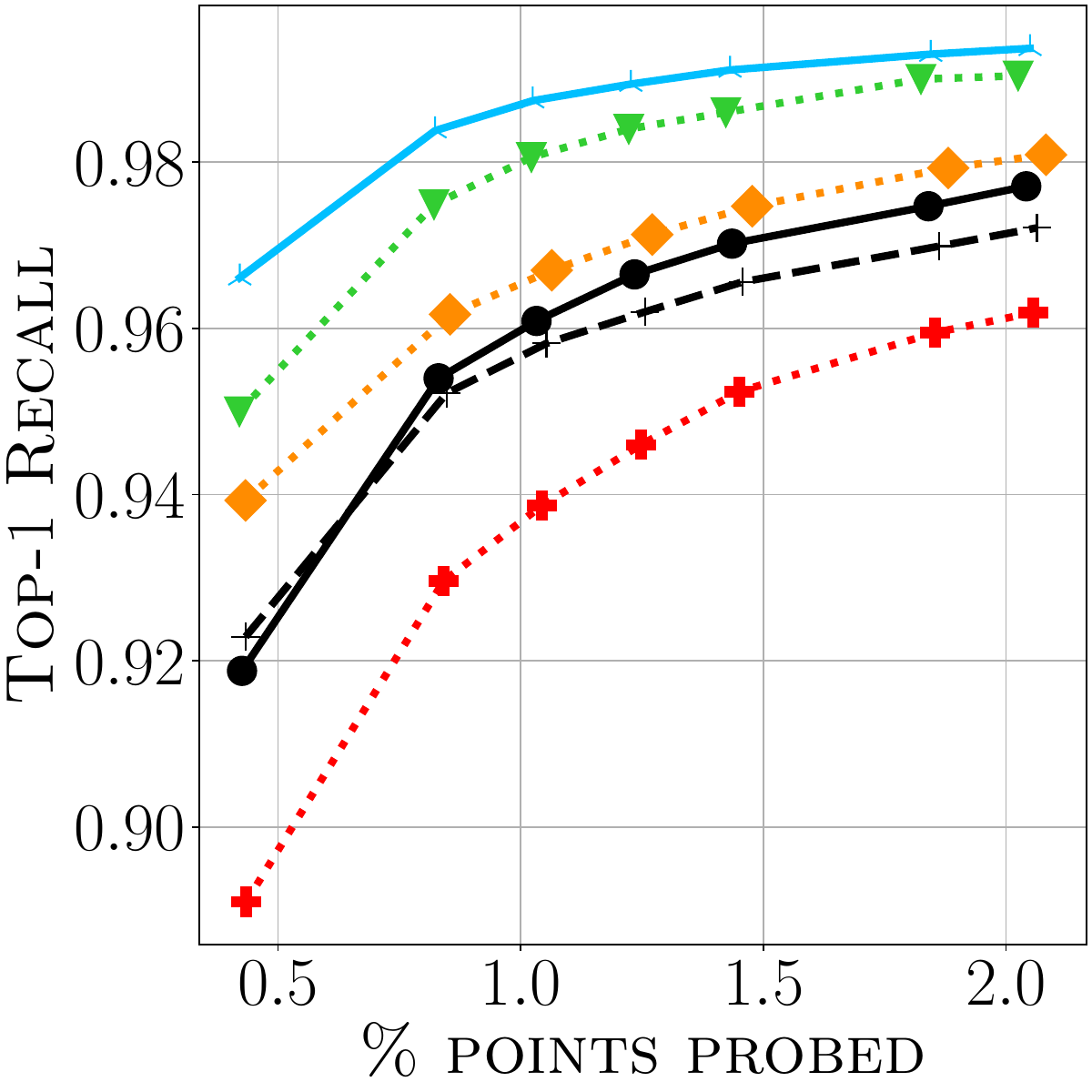}
        \includegraphics[width=0.26\linewidth]{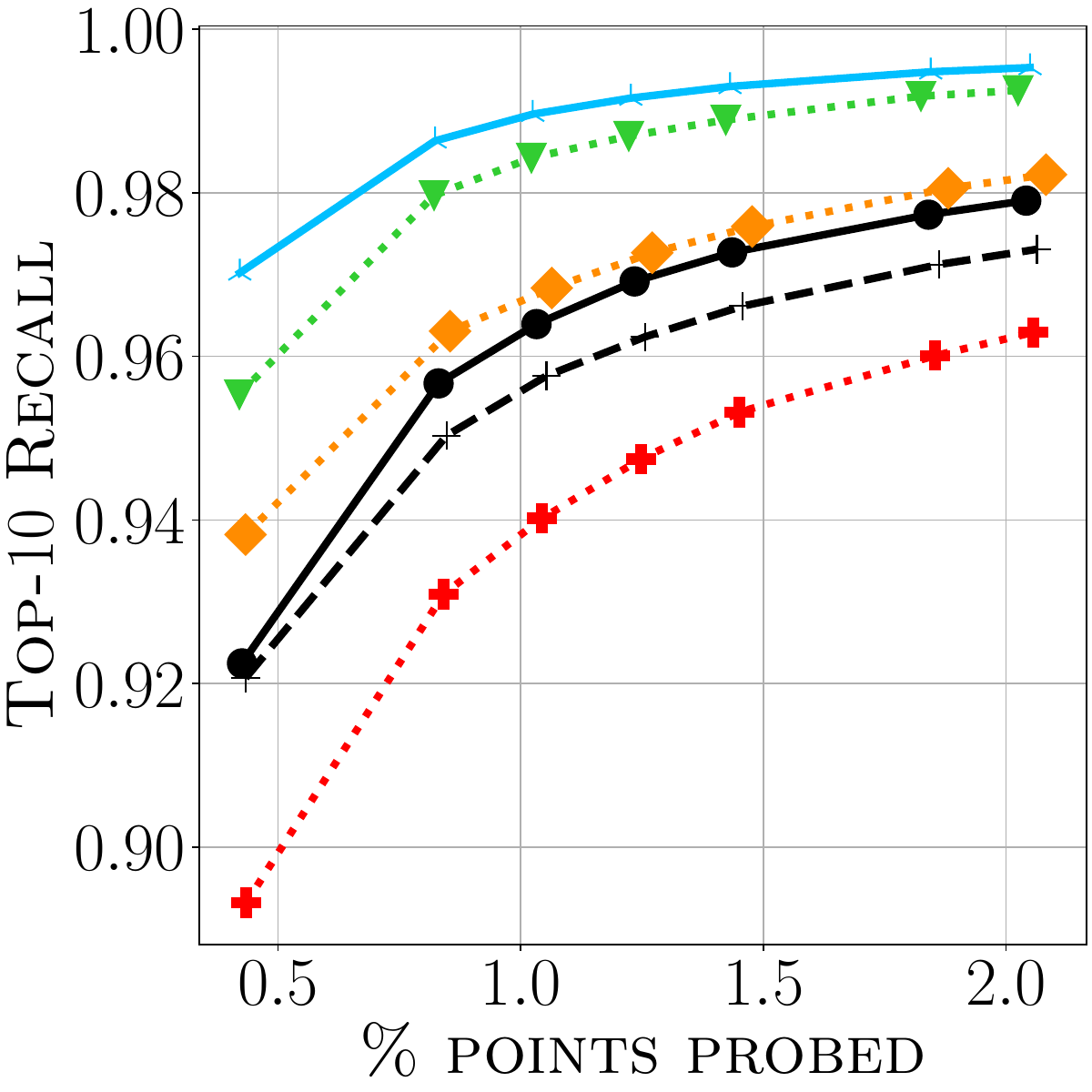}
        \includegraphics[width=0.26\linewidth]{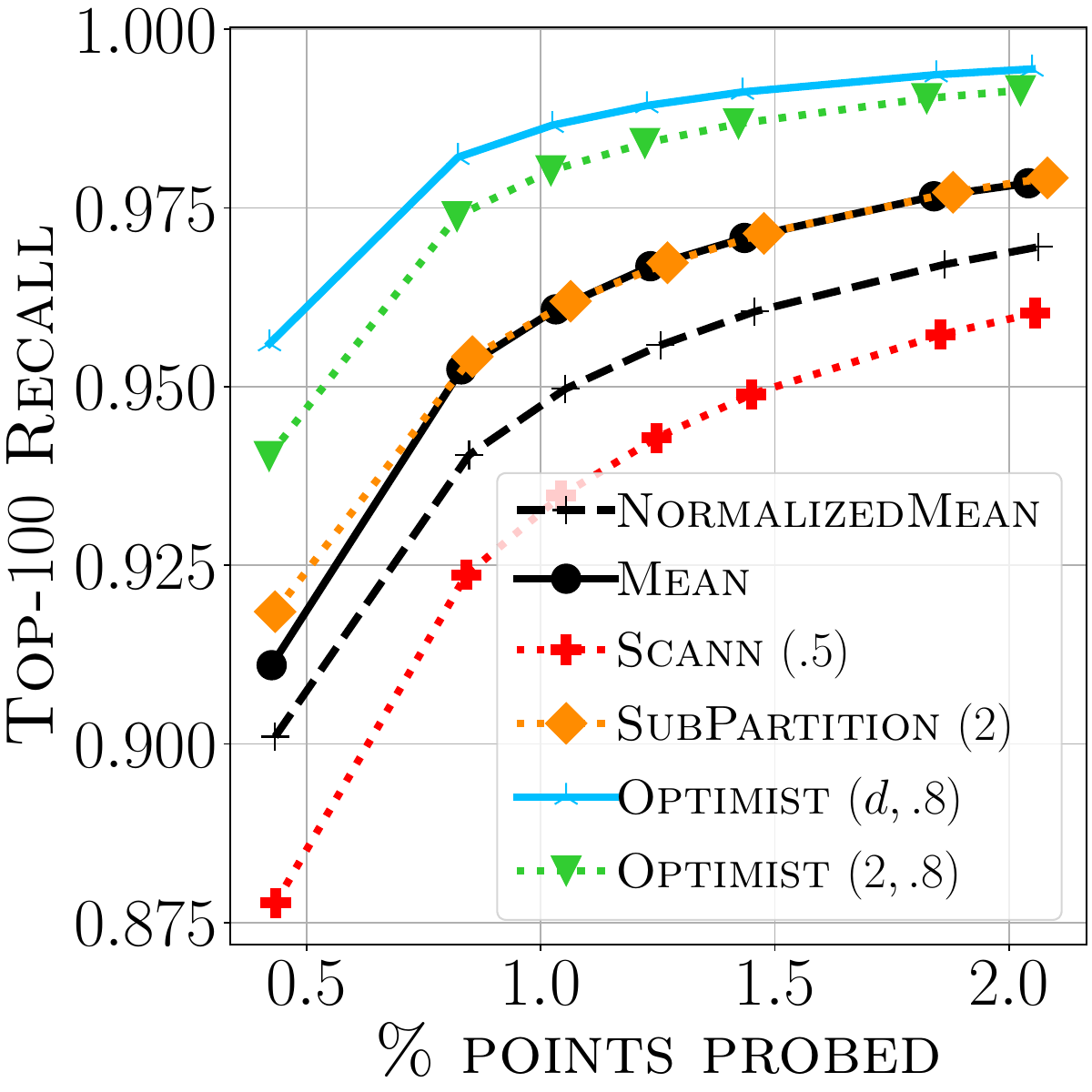}
    }
}
\centerline{
    \subfloat[\glove]{
        \includegraphics[width=0.26\linewidth]{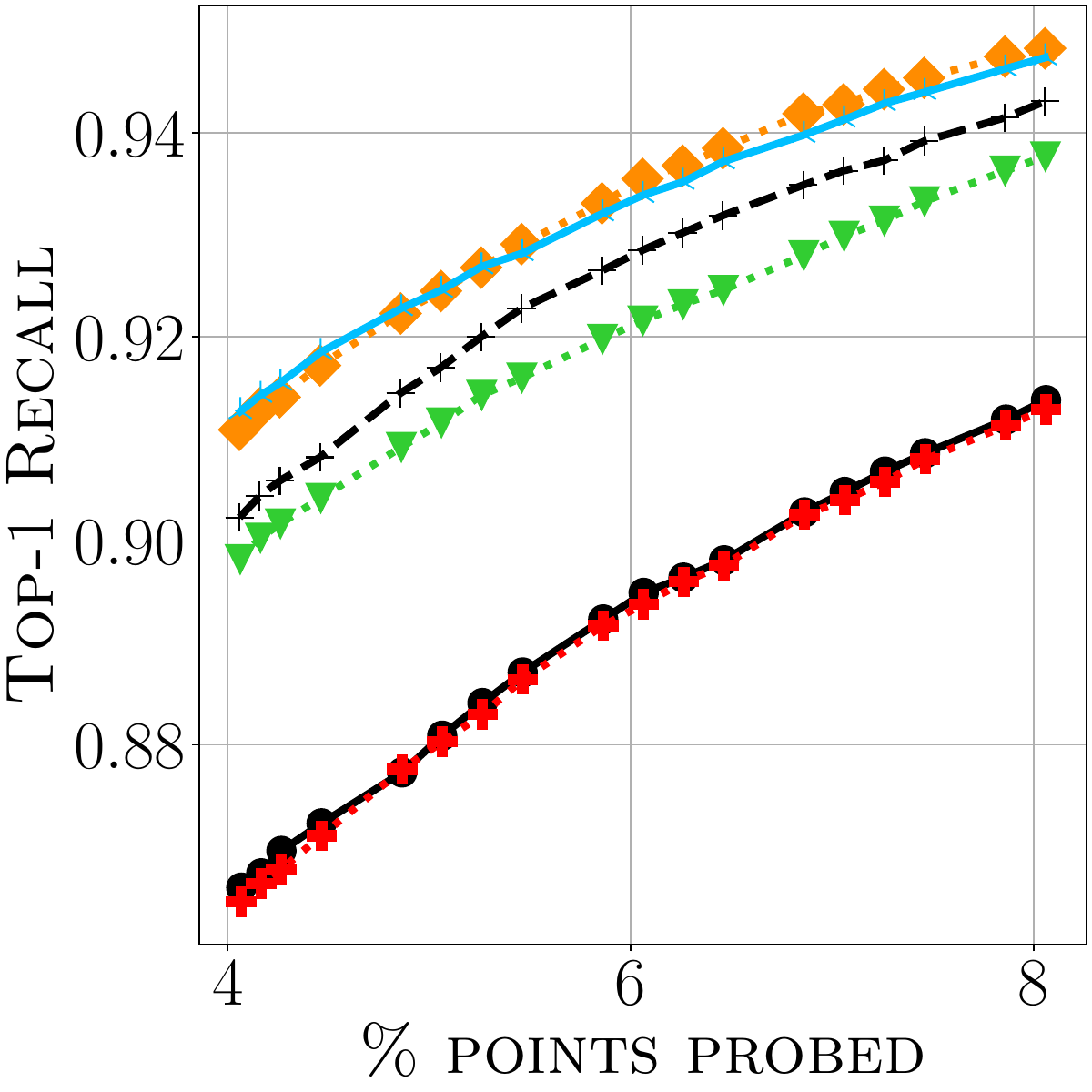}
        \includegraphics[width=0.26\linewidth]{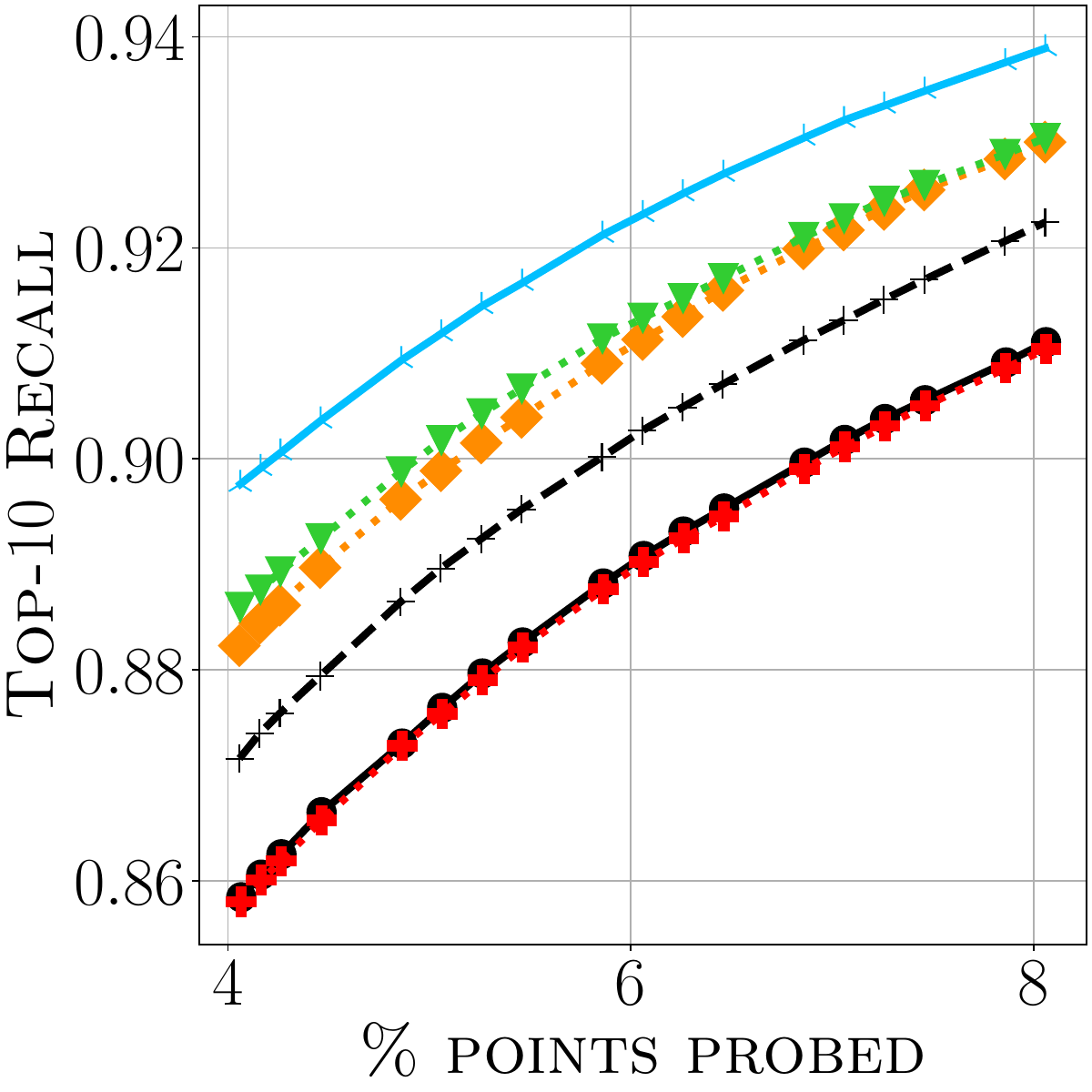}
        \includegraphics[width=0.26\linewidth]{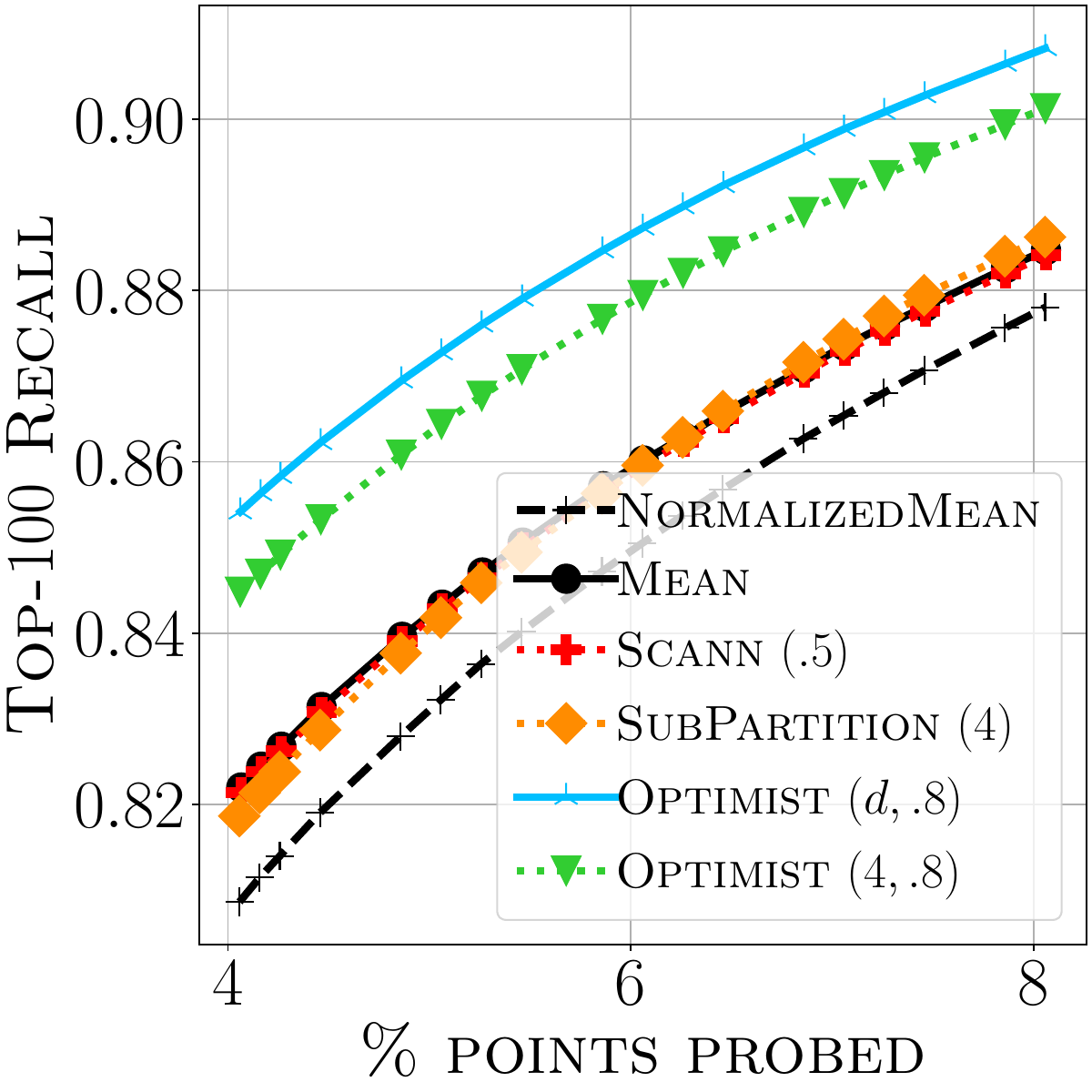}
    }
}
\centerline{
    \subfloat[\textimage]{
        \includegraphics[width=0.26\linewidth]{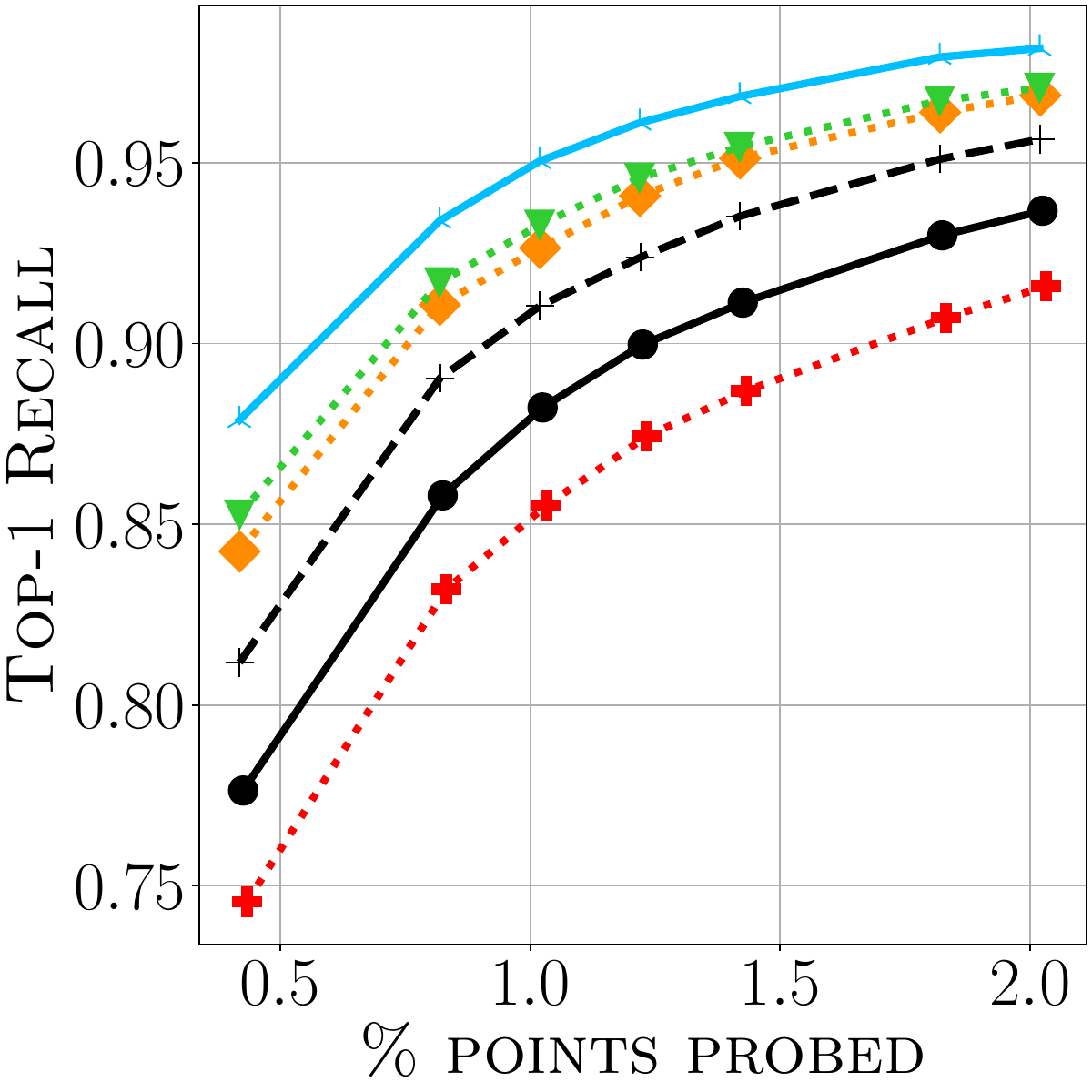}
        \includegraphics[width=0.26\linewidth]{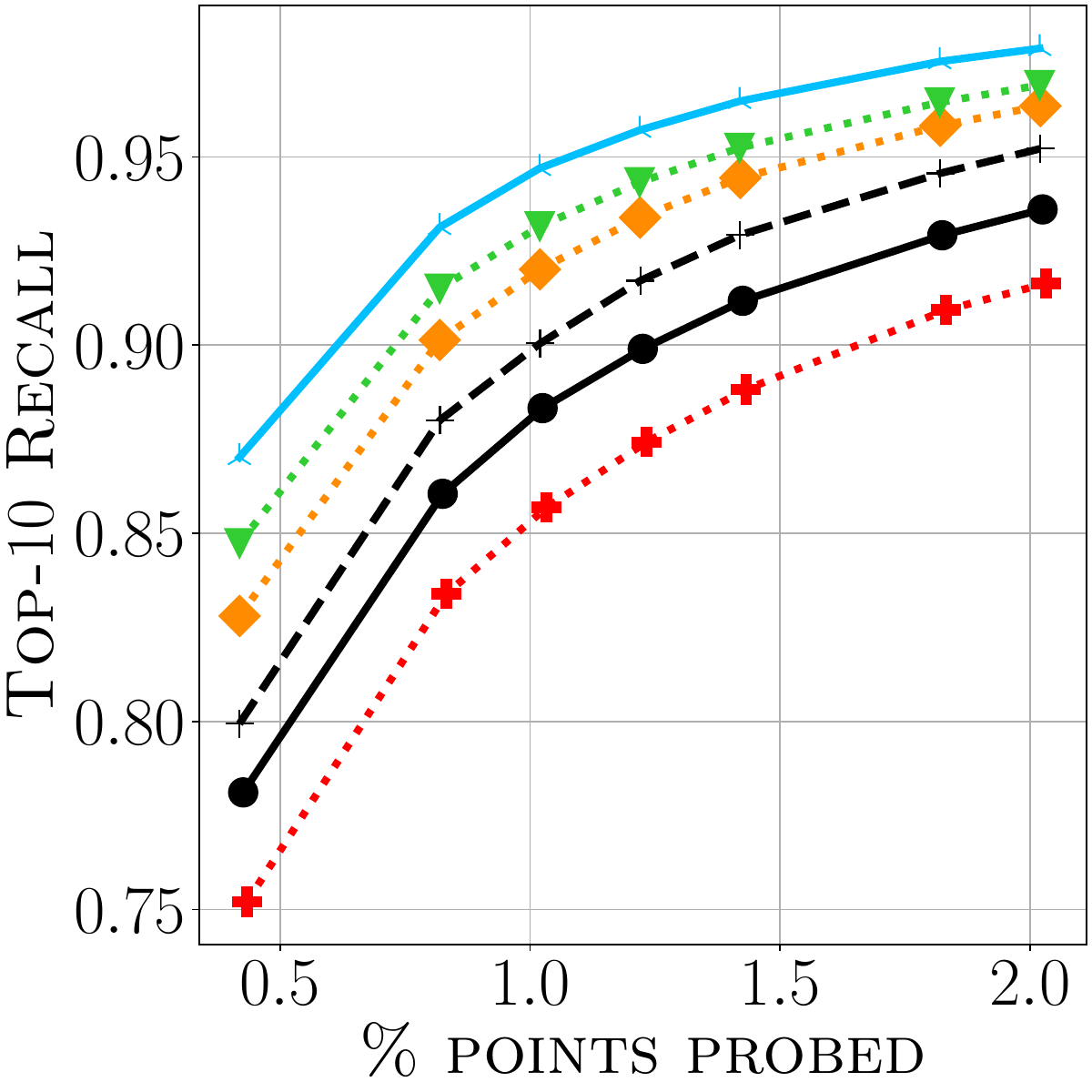}
        \includegraphics[width=0.26\linewidth]{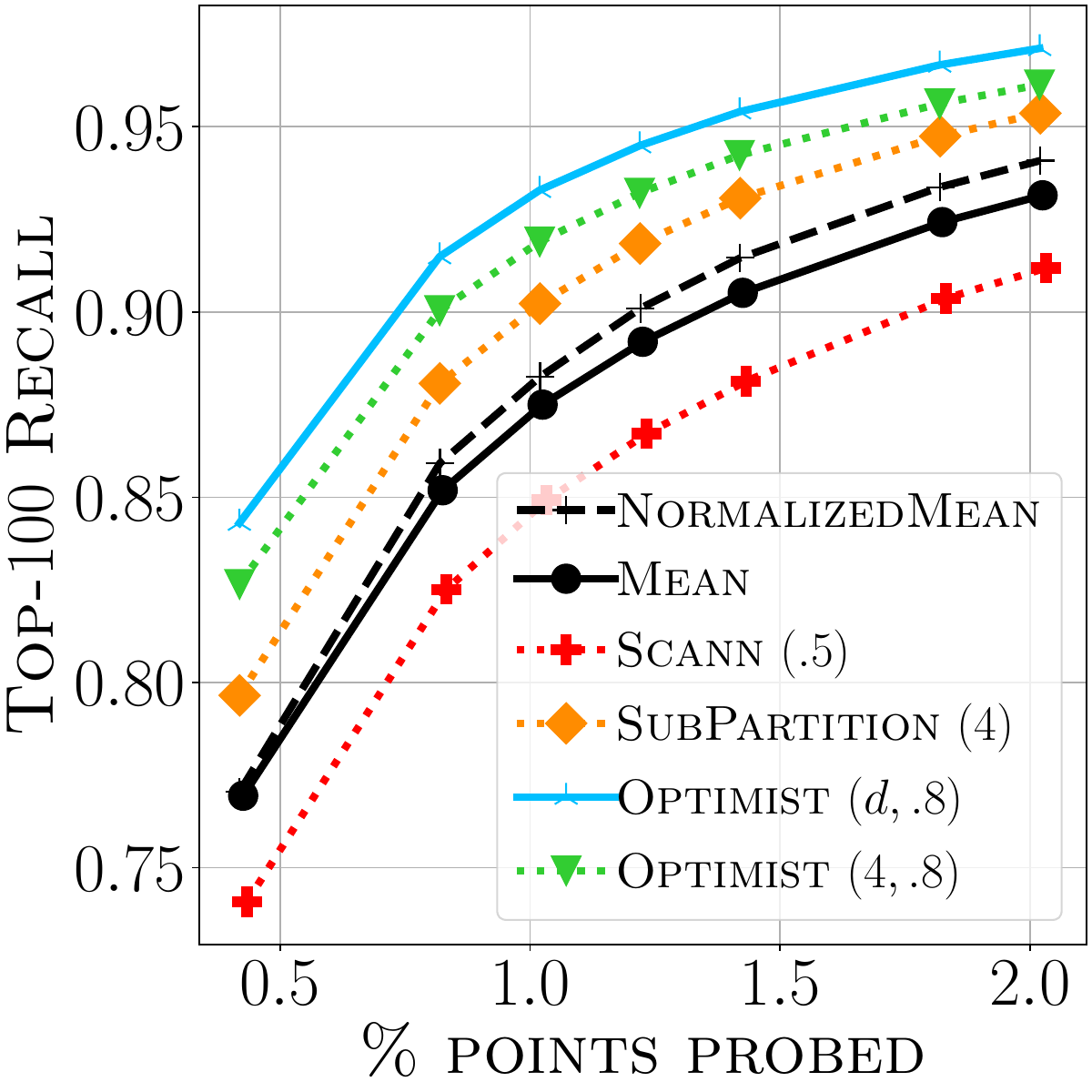}
    }
}
\centerline{
    \subfloat[\msmarco]{
        \includegraphics[width=0.26\linewidth]{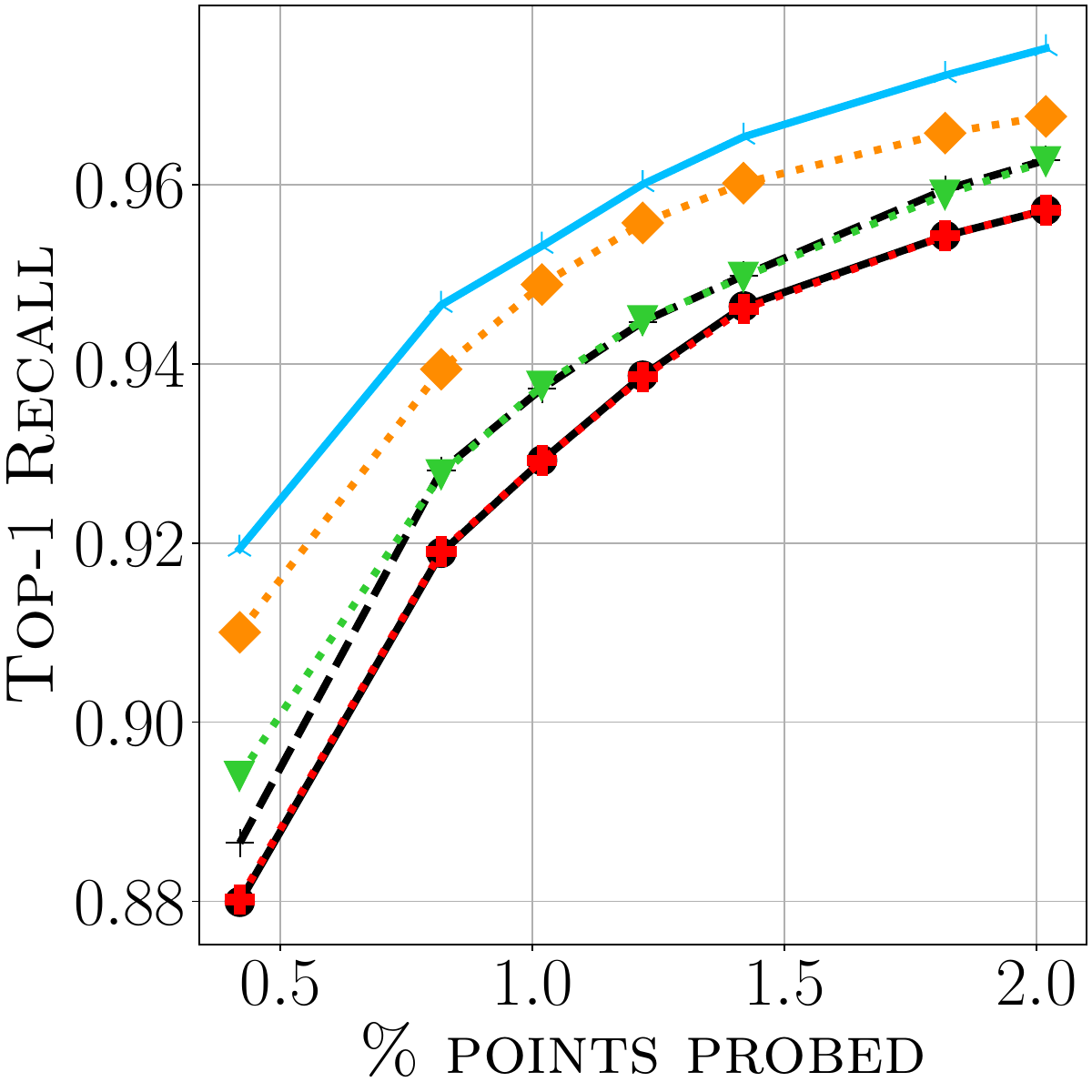}
        \includegraphics[width=0.26\linewidth]{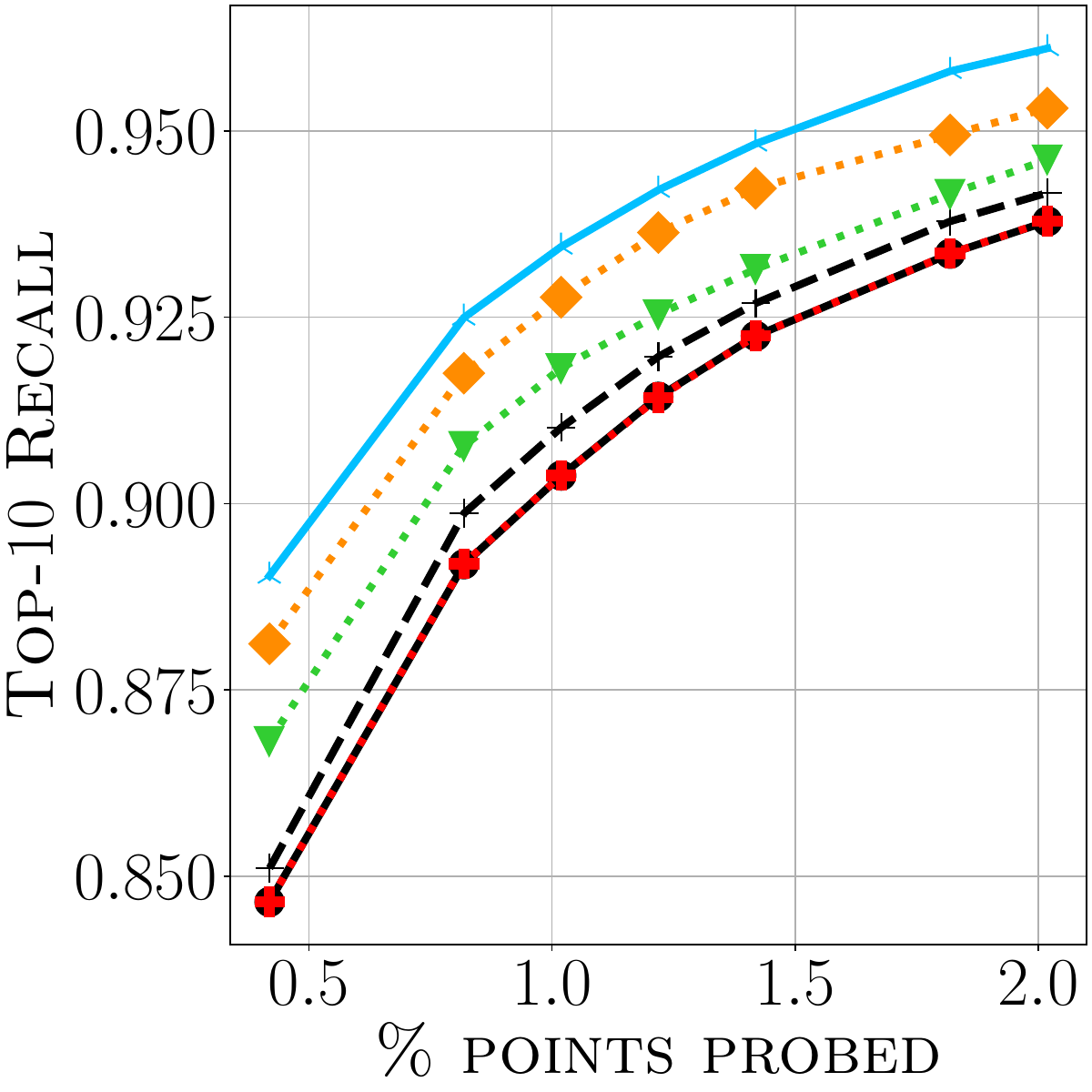}
        \includegraphics[width=0.26\linewidth]{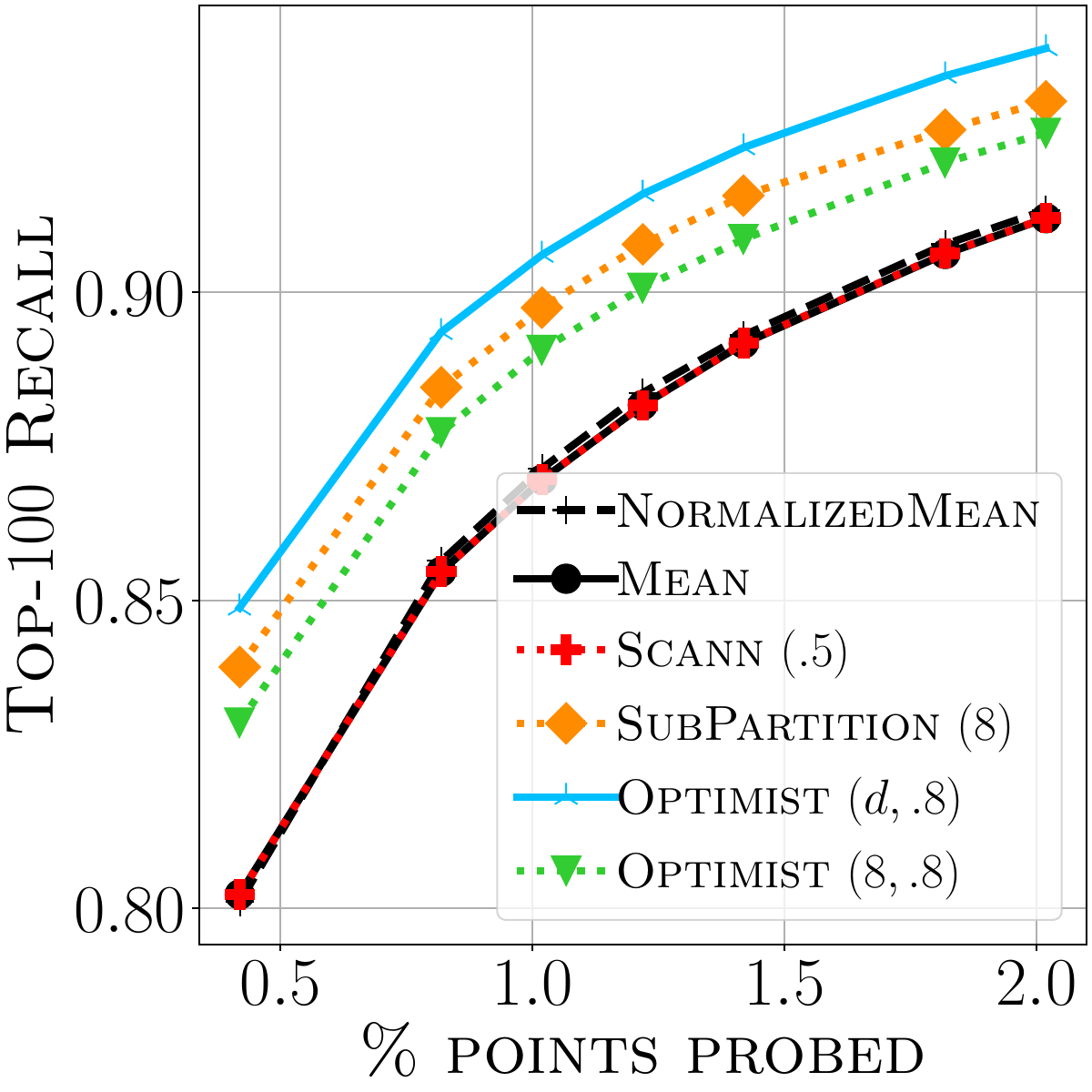}
    }
}
\caption{Top-$k$ recall vs. volume of probed data. Partitioning is with Gaussian Mixture Model. \scann has parameter $T$, \subpartition $t$ (leading to $t+2$ sub-partitions per shard), and \optimist rank $t$ and degree of optimism $\delta$. We note that, due to the dimensionality of \nq, we were unable to complete GMM clustering on this particular dataset.}
\label{figure:gmm:full}
\end{center}
\end{figure*}

\begin{figure*}[ht]
\ContinuedFloat
\begin{center}
\centerline{
    \subfloat[\deep]{
        \includegraphics[width=0.26\linewidth]{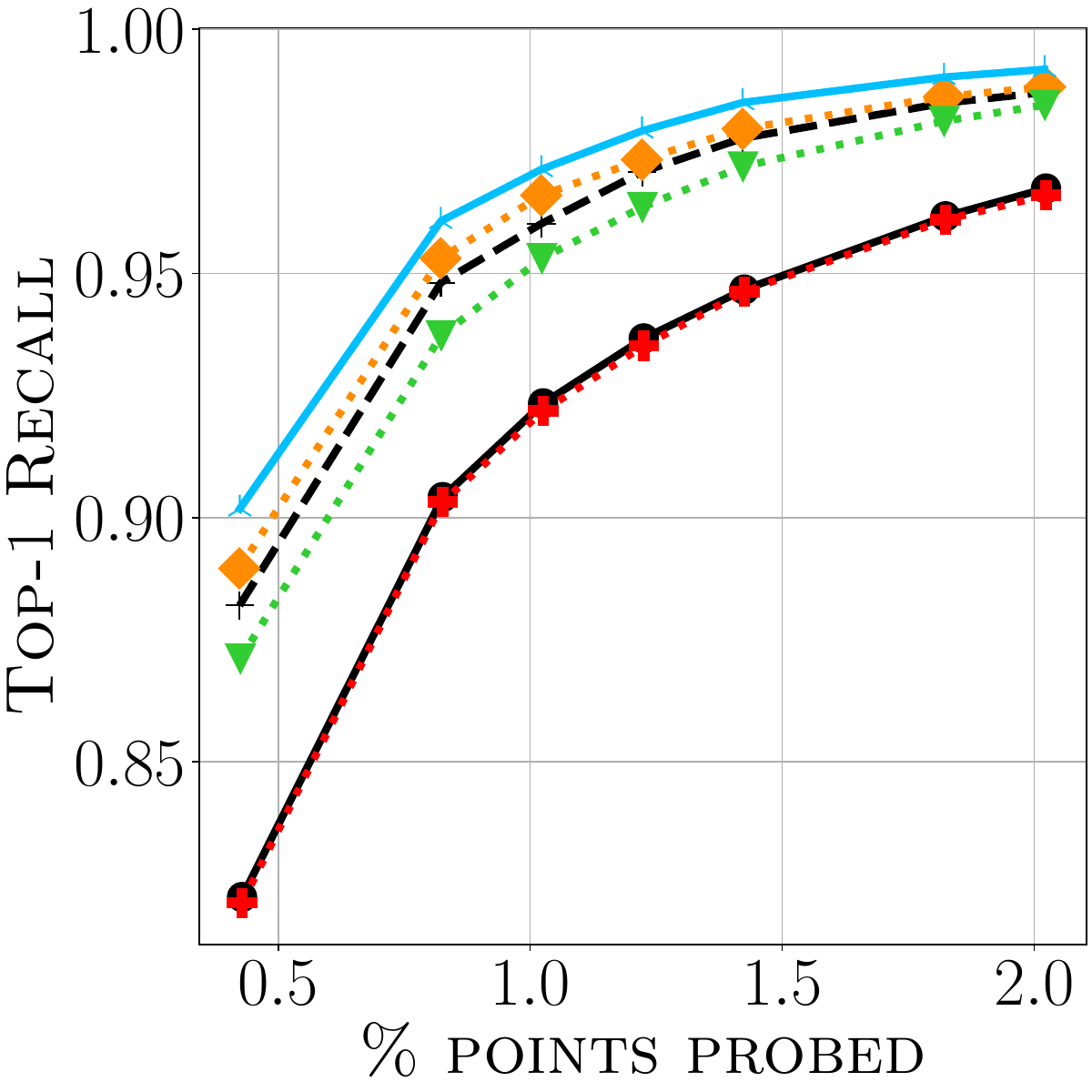}
        \includegraphics[width=0.26\linewidth]{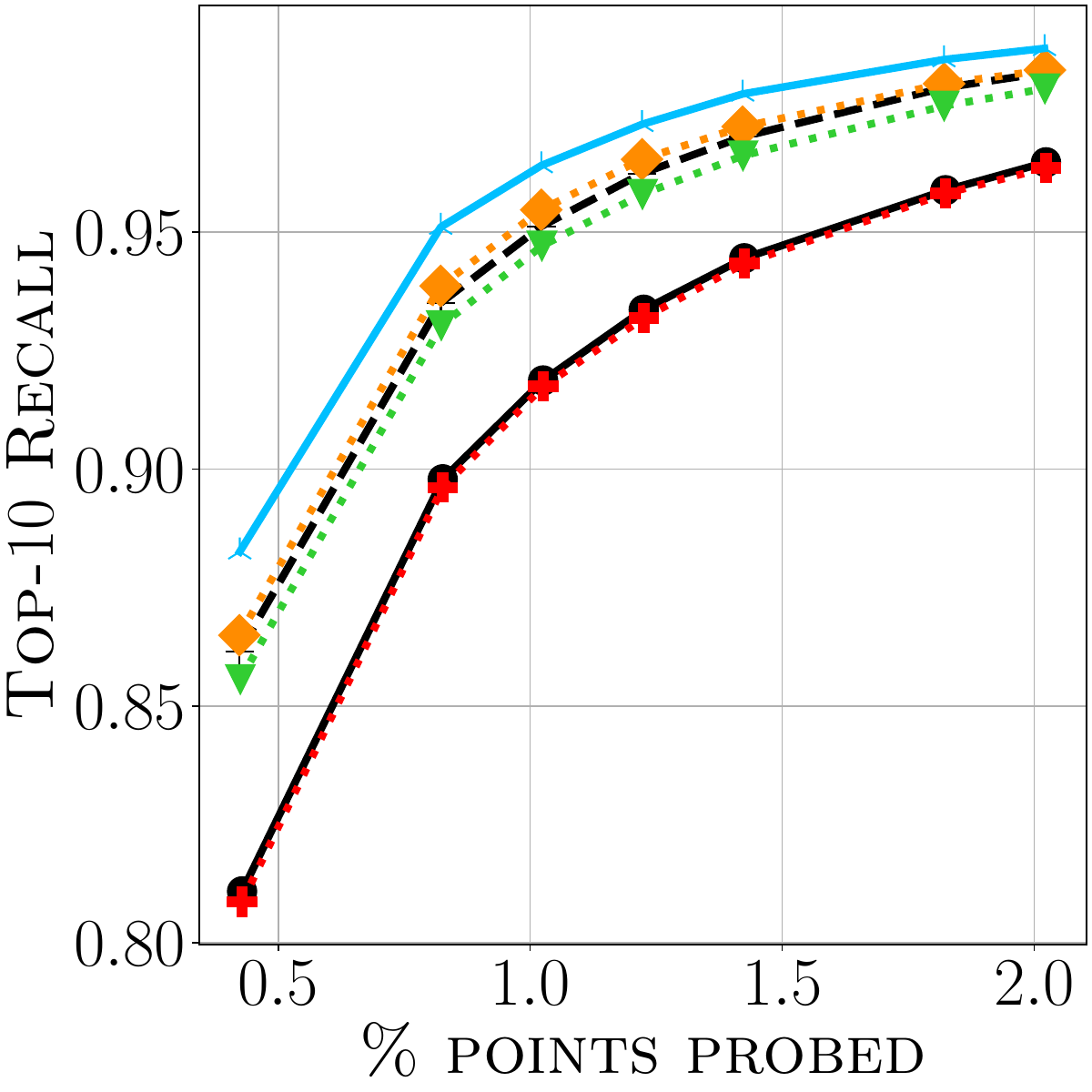}
        \includegraphics[width=0.26\linewidth]{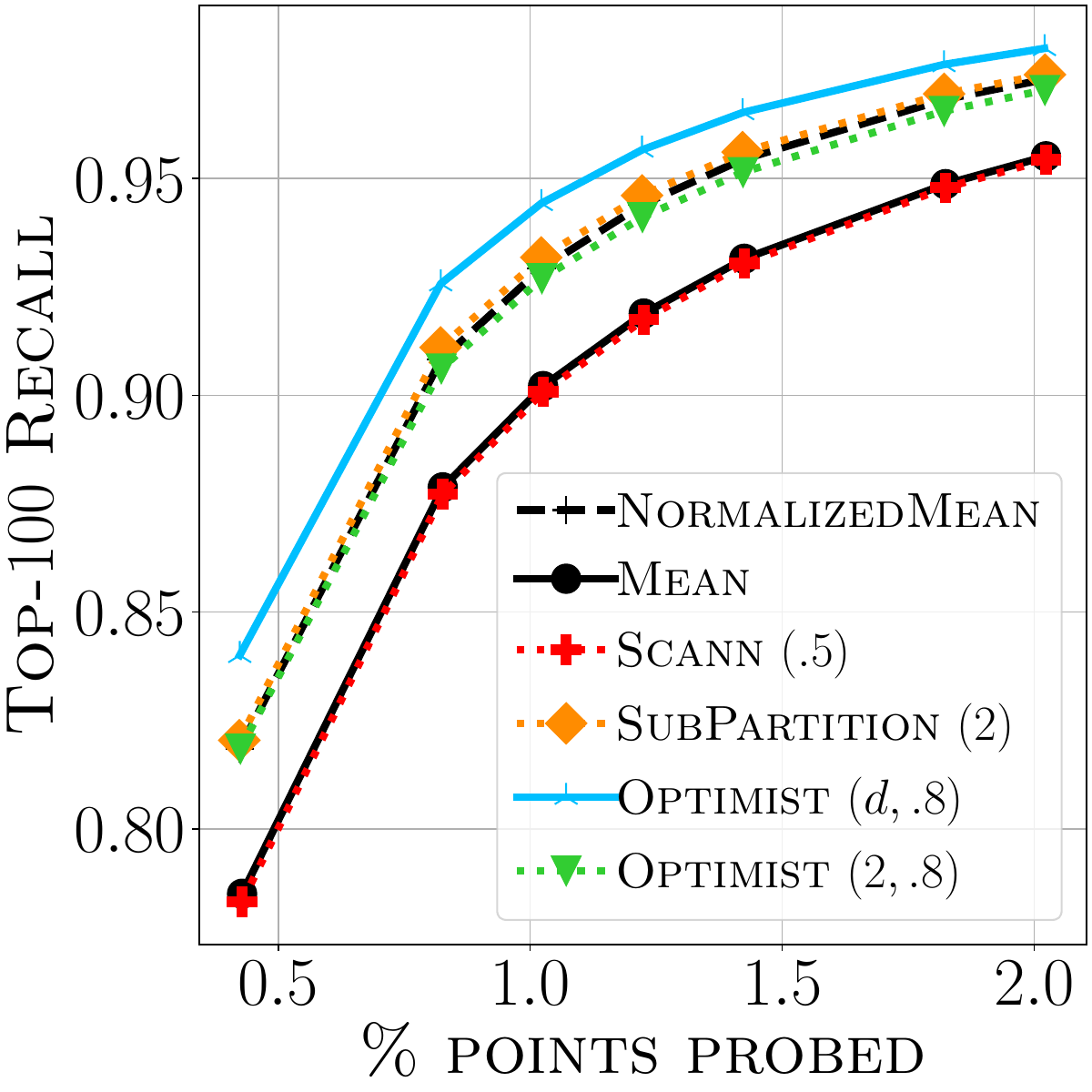}
    }
}
\caption{Top-$k$ recall vs. volume of probed data. Partitioning is with Gaussian Mixture Model. \scann has parameter $T$, \subpartition $t$ (leading to $t+2$ sub-partitions per shard), and \optimist rank $t$ and degree of optimism $\delta$. We note that, due to the dimensionality of \nq, we were unable to complete GMM clustering on this particular dataset.}
\label{figure:gmm:full}
\end{center}
\end{figure*}

\newpage
\section{Latency comparison}
\label{appendix:latency}

\begin{figure}[h]
\begin{center}
\centerline{
        \includegraphics[width=0.28\linewidth]{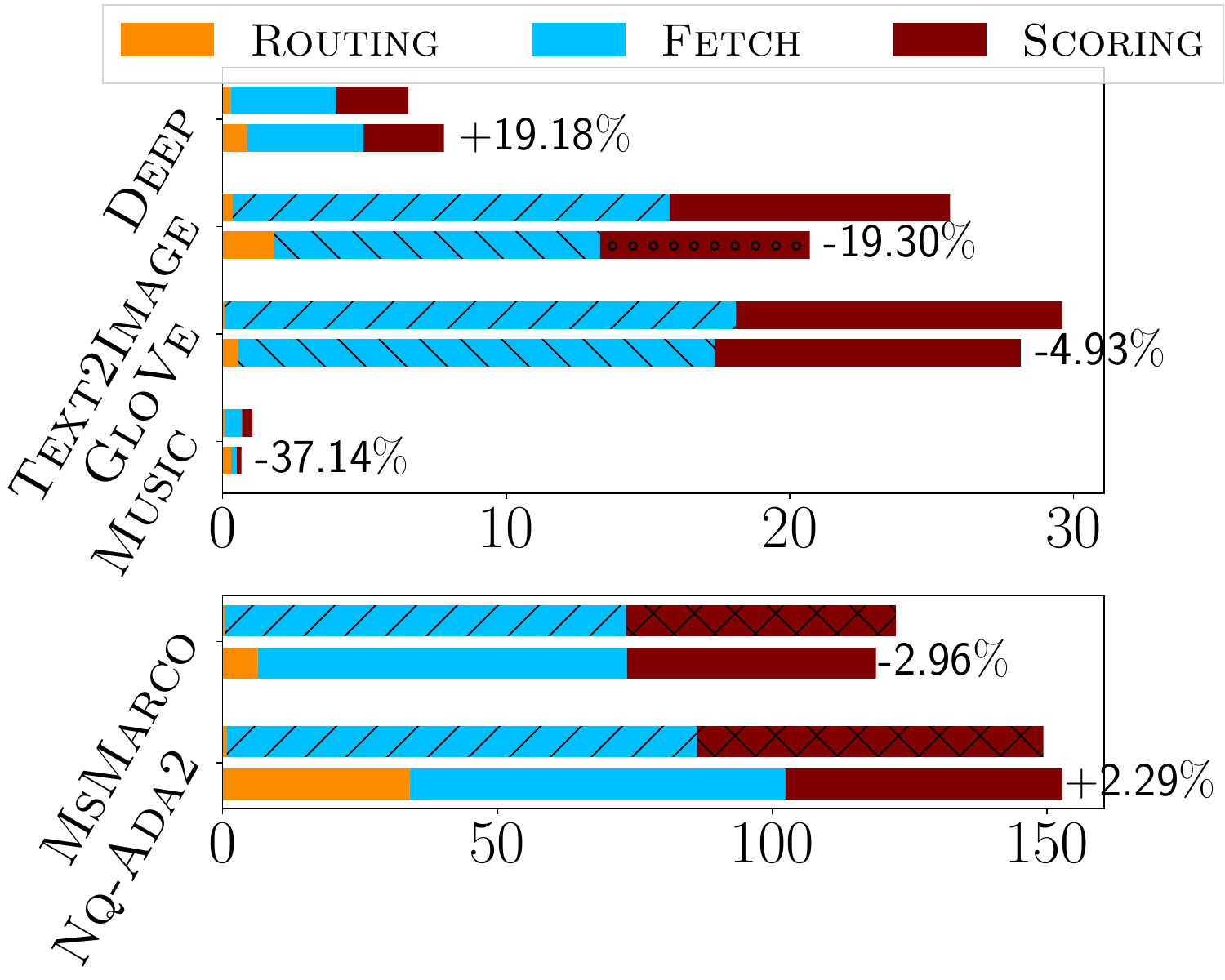}
        \includegraphics[width=0.28\linewidth]{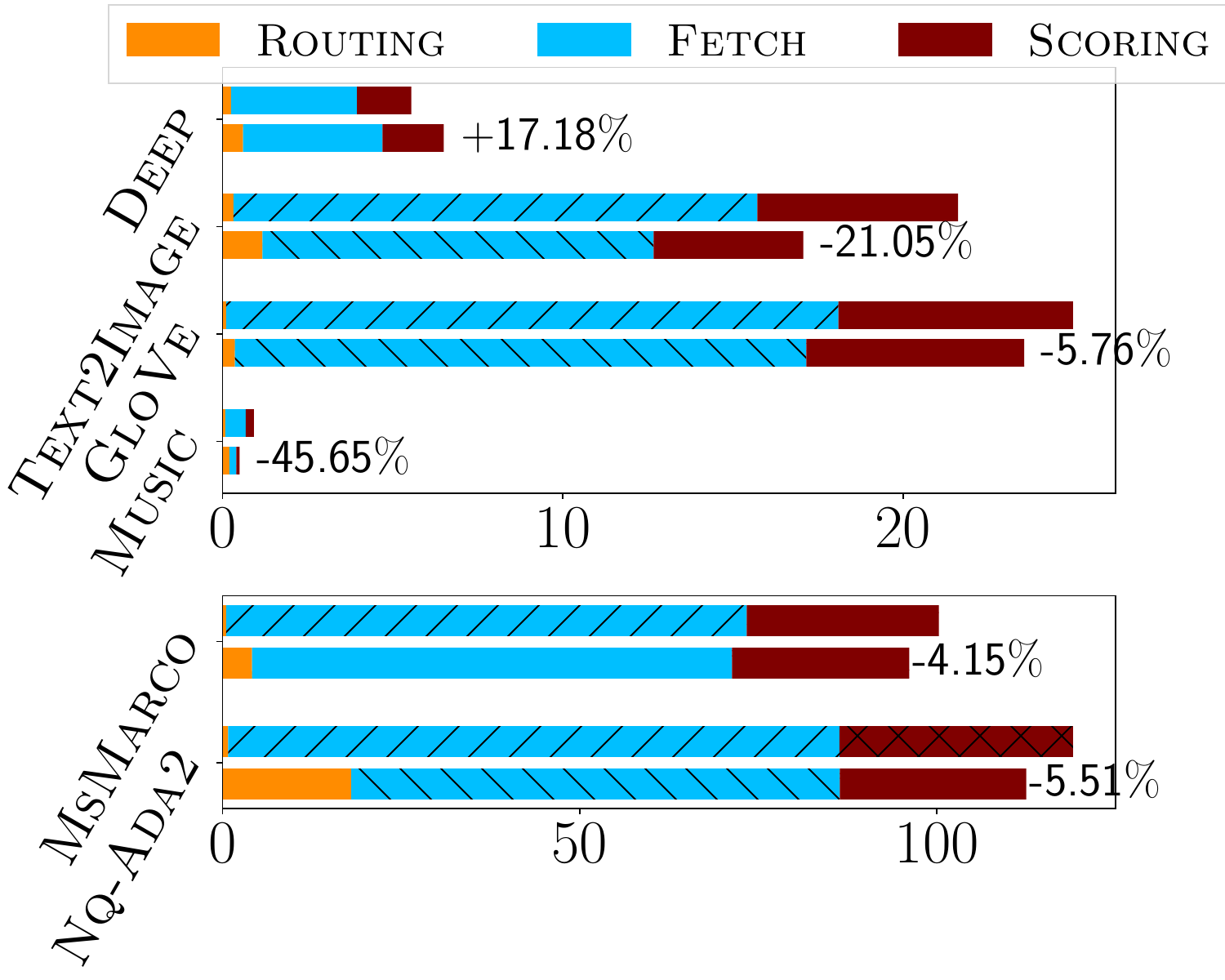}
        \includegraphics[width=0.28\linewidth]{figures/latencies/latencies_ssd_4_threads.pdf}
}
\centerline{
    \subfloat[$1$ thread]{
        \includegraphics[width=0.28\linewidth]{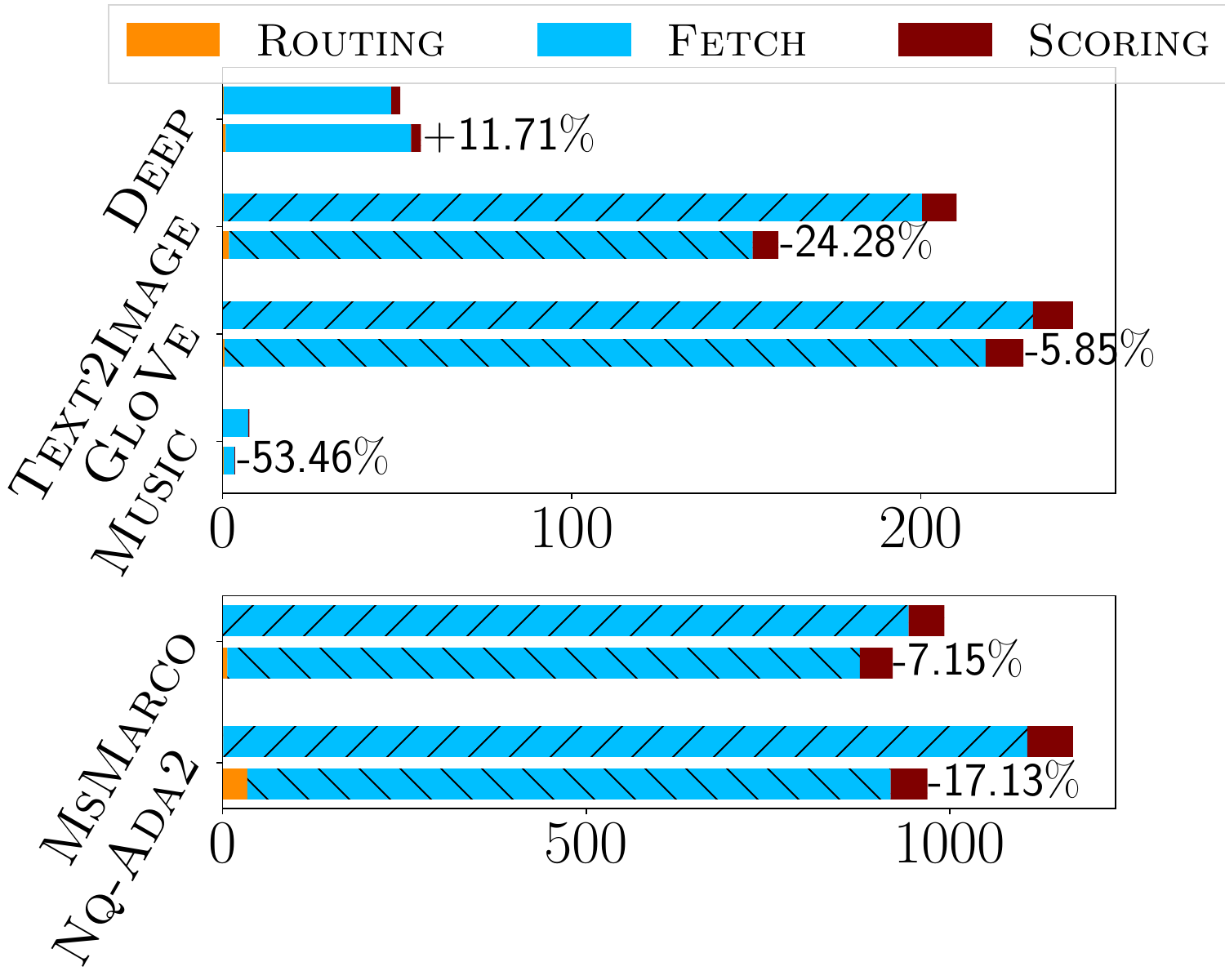}
    }
    \subfloat[$2$ threads]{
        \includegraphics[width=0.28\linewidth]{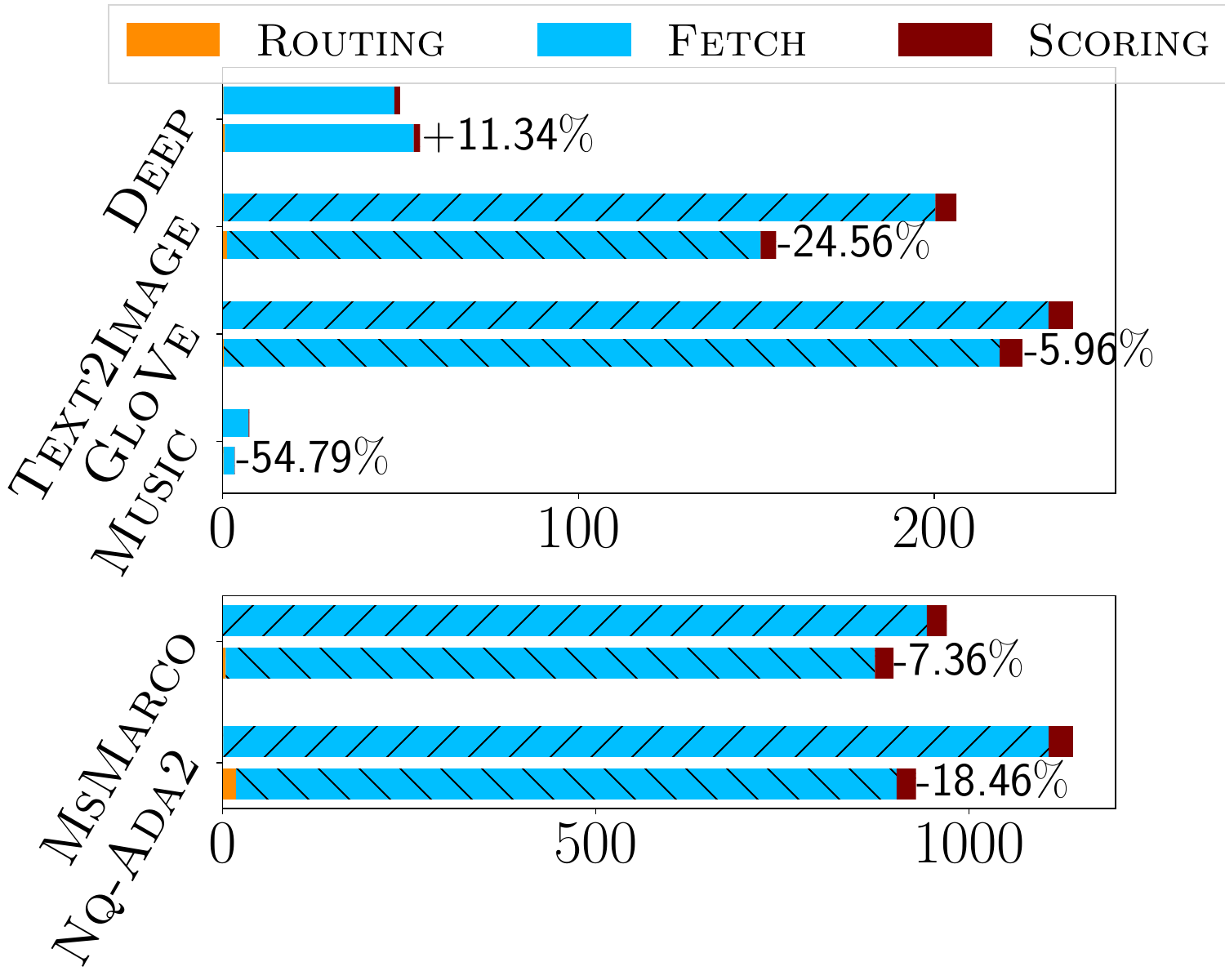}
    }
    \subfloat[$4$ threads]{
        \includegraphics[width=0.28\linewidth]{figures/latencies/latencies_blob_4_threads.pdf}
    }
}
\caption{Mean latency (ms) to reach $95\%$ recall, using different number of threads (columns), when PQ-compressed shards are on SSD (top row) and blob storage (bottom row). For each dataset, we plot the latency breakdown for \normalizedmean (top bar) and \optimist (bottom bar), and report relative gains (negative value indicates gain by \optimist).}
\label{appendix:table:latency}
\end{center}
\end{figure}

\newpage
\section{Maximum inner product prediction}
\label{appendix:max-ip-prediction}

\begin{figure}[h]
\begin{center}
\centerline{
    \subfloat[\nq]{
        \includegraphics[width=0.3\linewidth]{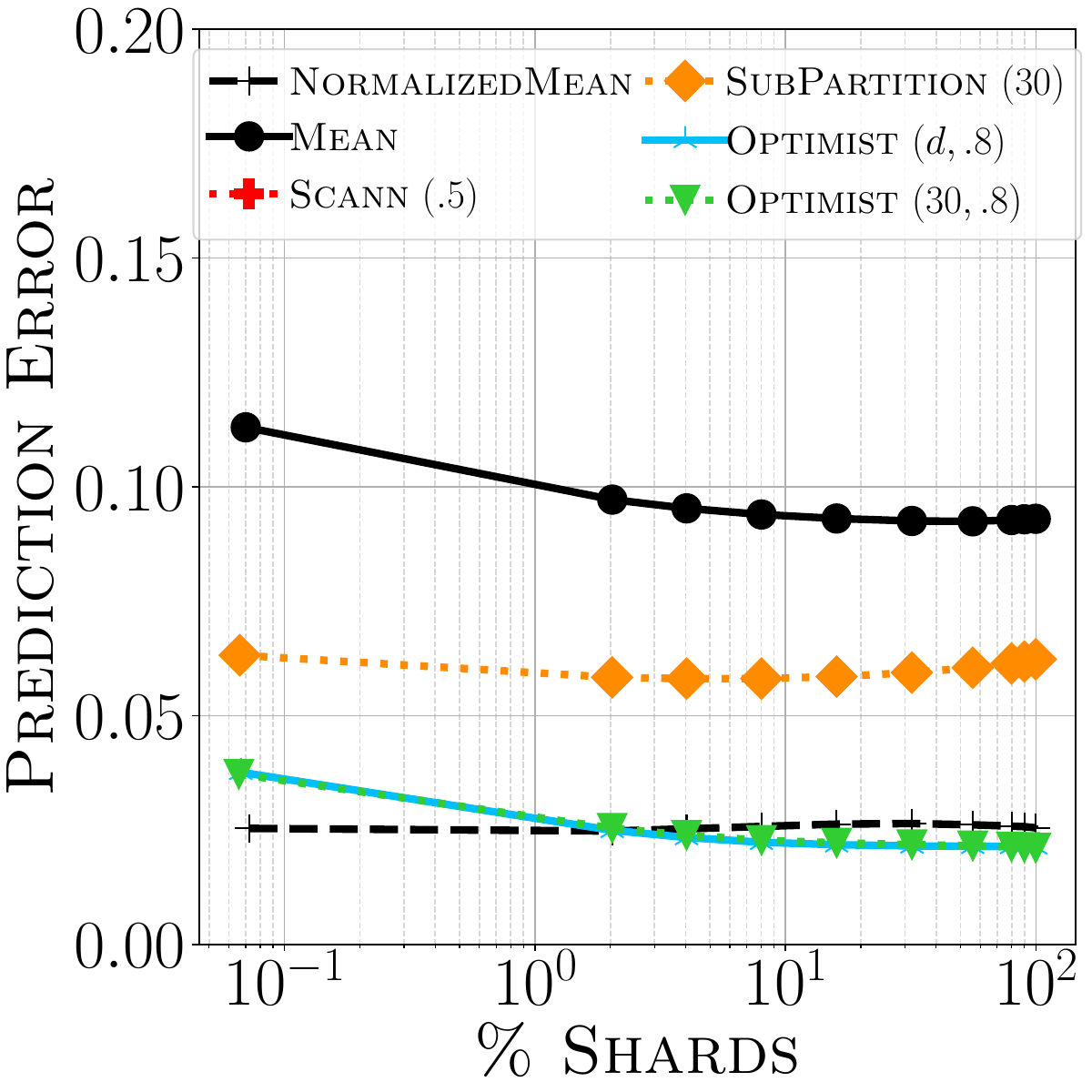}
    }
    \subfloat[\glove]{
        \includegraphics[width=0.3\linewidth]{figures/prediction_error/glove_200.pdf}
    }
    \subfloat[\msmarco]{
        \includegraphics[width=0.3\linewidth]{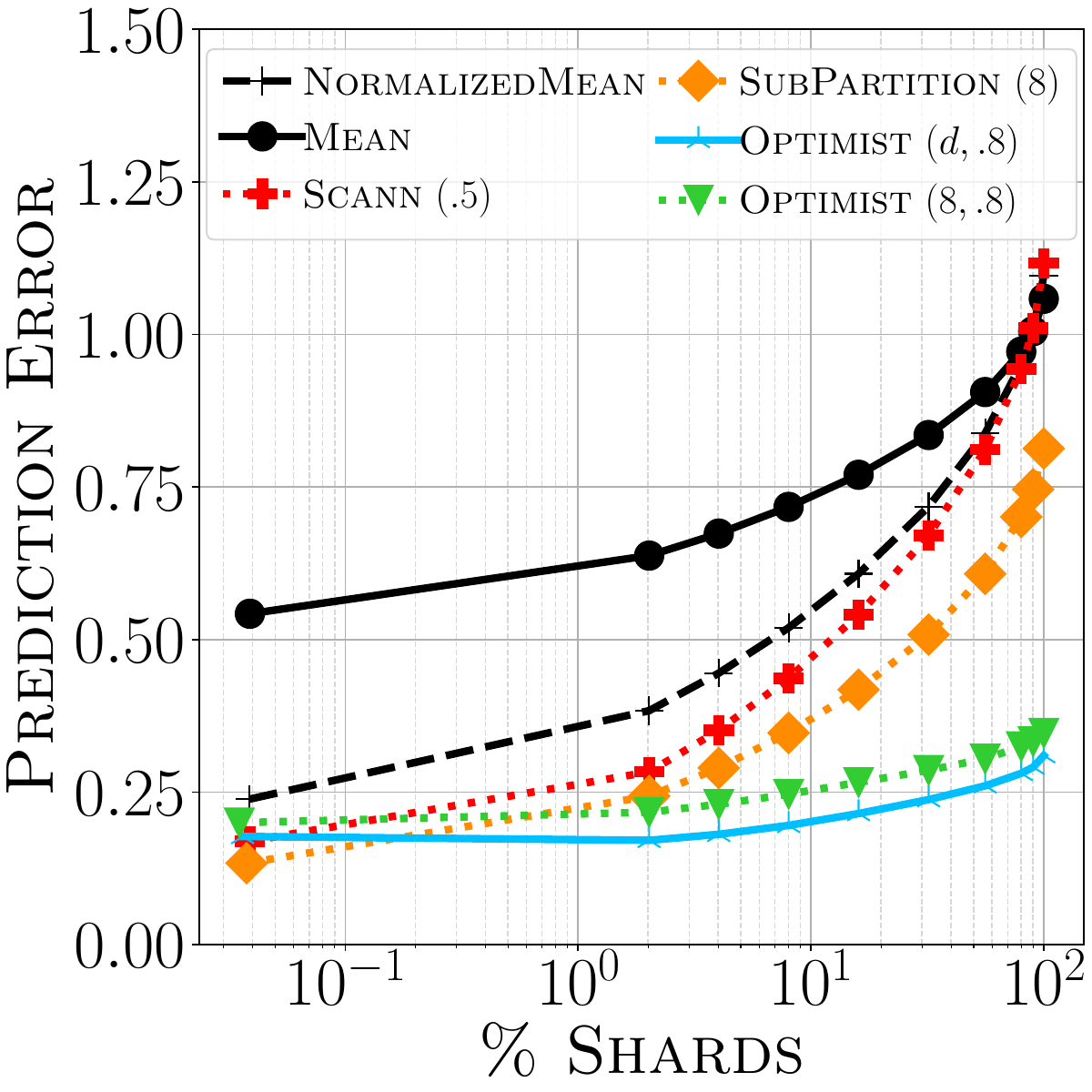}
    }
}
\centerline{
    \subfloat[\deep]{
        \includegraphics[width=0.3\linewidth]{figures/prediction_error/deep1m.pdf}
    }
    \subfloat[\music]{
        \includegraphics[width=0.3\linewidth]{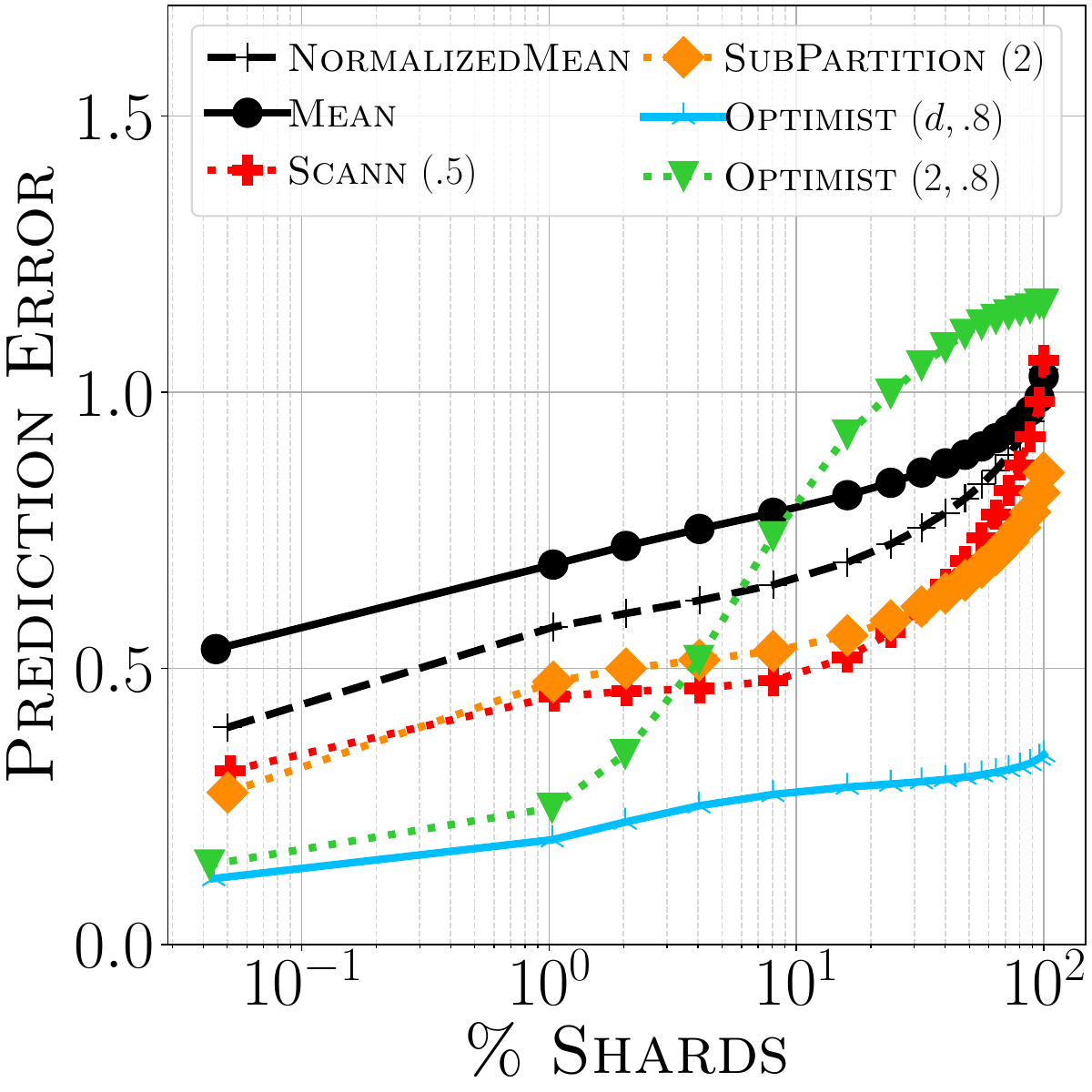}
    }
    \subfloat[\textimage]{
        \includegraphics[width=0.3\linewidth]{figures/prediction_error/text2image.pdf}
    }
}
\caption{Mean prediction error $\mathcal{E}_\ell(\tau, \cdot)$,  defined in Equation~(\ref{equation:general-prediction-error}), versus $\ell$ (expressed as percent of total number of shards), for various routers and datasets.}
\label{appendix:figure:prediction-error-breakdown}
\end{center}
\end{figure}

\end{document}